\pdfoutput=1
\documentclass[12pt]{article}
\usepackage{amsmath}
\usepackage{graphicx}
\usepackage{enumerate}
\usepackage[square,sort,comma,numbers]{natbib}
\usepackage{url} 
\usepackage{subfiles}
\usepackage{mathtools}
\usepackage{dsfont}
\usepackage[ruled,vlined]{algorithm2e}
\usepackage{amsmath,amsthm,amssymb}
\usepackage{bbm}

\usepackage{booktabs}
\usepackage{titletoc}
\usepackage{titlesec}

\usepackage{paralist}
\usepackage{mathtools}
\usepackage{dsfont}
\usepackage{hyperref}
\usepackage{url}
\usepackage{xcolor}
\usepackage{breqn}
\usepackage{footnote}
\usepackage{graphicx}
\usepackage{tipa}
\usepackage{subfig}
\usepackage{bm}
\usepackage{paralist}
\usepackage{calrsfs}
\usepackage{subfiles}
\usepackage{dutchcal}
\usepackage{upgreek}

\usepackage{amsmath,amsfonts,bm}

\def\eqref#1{equation~\ref{#1}}

\def\1{\bm{1}}

\DeclareMathAlphabet{\mathsfit}{\encodingdefault}{\sfdefault}{m}{sl}
\SetMathAlphabet{\mathsfit}{bold}{\encodingdefault}{\sfdefault}{bx}{n}

\newcommand{\R}{\mathbb{R}}

\newtheorem{theorem}{Theorem}[section]
\newtheorem{corollary}{Corollary}[theorem]
\newtheorem{lemma}{Lemma}[section]
\newtheorem{proposition}{Proposition}
\newtheorem{assumption}{Assumption}

\newtheorem{definition}{Definition}
\newtheorem{remark}{Remark}

\numberwithin{mytheorem}{section}

\numberwithin{mylemma}{section}
\newtheorem{innercustomgeneric}{\customgenericname}
\providecommand{\customgenericname}{}

\newcommand{\bb}[1]{\mathbb{#1}}

\newcommand{\wh}[1]{\widehat{#1}}

\newcommand{\bfa}[1]{\boldsymbol{#1}}

\DeclareMathAlphabet{\pazocal}{OMS}{zplm}{m}{n}
\newcommand{\ca}[1]{\pazocal{#1}}

\newcommand{\nnb}{\nonumber}

\newcommand{\msf}[1]{\mathsf{#1}}

\newcommand{\wha}[1]{\wh{\bfa #1}}

\newcommand{\lef}{\left}
\newcommand{\rig}{\right}

\newcommand{\tda}[1]{\tilde{\bfa{#1}}}

\newcommand{\pta}{\partial}

\newcommand{\mbbm}[1]{\mathbbm{#1}}

\newcommand{\tde}{\tilde}

\newcommand{\la}{\langle}
\newcommand{\ra}{\rangle}
\newcommand{\diag}{\text{diag}}

\newcommand{\ub}[1]{\underbrace{#1}}

\newcommand{\bl}{{\bigg (}}
\newcommand{\br}{{\bigg )}}

\newcommand{\sm}[1]{\small{#1}\normalsize}
\newcommand{\tny}[1]{\footnotesize{#1}\normalsize}

\DeclareMathOperator*{\argmax}{arg\,max}

\DeclareMathOperator*{\argmin}{arg\,min}

\DeclareMathOperator{\sign}{sign}

\usepackage{hyperref}
\usepackage{url}
\newcommand{\blind}{1}

\begin{document}

\def\spacingset#1{\renewcommand{\baselinestretch}%
{#1}\small\normalsize} \spacingset{1}


\if1\blind
{
  \title{\bf Learning Spectral Methods by Transformers}
   \author{\hspace{.2cm} Yihan He\thanks{Equal Contribution.}~\thanks{Princeton University, \url{{yihan.he, jqfan}@princeton.edu}
    },\quad
    Yuan Cao\footnotemark[1]~\thanks{The University of Hong Kong, \url{yuancao@hku.hk}} ,\quad Hong-Yu Chen\thanks{Northwestern University, \url{{hong-yuchen2029, yu-hsuanwu2024, hanliu}@northwestern.edu} },\quad Dennis Wu\footnotemark[4],\\\hspace{.2cm}
    Jianqing Fan\footnotemark[2],\quad and\quad Han Liu\footnotemark[4]} 
    \date{\today}
  \maketitle
} \fi

\if0\blind
{
  \bigskip
  \bigskip
  \bigskip
  \begin{center}
    {\LARGE\bf Learning Spectral Methods by Transformers}
\end{center}
  \medskip
} \fi

\bigskip
\begin{abstract}
Transformers demonstrate significant advantages as the building block of modern LLMs. In this work, we study the capacities of Transformers in performing unsupervised learning. We show that multi-layered Transformers, given a sufficiently large set of pre-training instances, can learn the algorithms themselves and perform statistical estimation tasks given new instances. This learning paradigm is distinct from the in-context learning setup and is similar to the learning procedure of human brains where skills are learned through past experience. Theoretically, we prove that pre-trained Transformers can learn the spectral methods and use the classification of the bi-class Gaussian mixture model as an example. Our proof is constructive using algorithmic design techniques. Our results are built upon the similarities of multi-layered Transformer architecture with the iterative recovery algorithms used in practice. Empirically, we verify the strong capacity of the multi-layered (pre-trained) Transformer on unsupervised learning through the lens of both the PCA and the Clustering tasks performed on the synthetic and real-world datasets.
\end{abstract}

\noindent%
\vfill

\newpage
\spacingset{1.9} 
\section{Introduction}
Large Language Models (LLMs) demonstrate significant success in learning and performing inference on real-world high dimensional datasets. Most modern LLMs use Transformers \citep{vaswani2017attention} as their backbones, which demonstrate significant advantages over many existing neural network models. Transformers achieve many state-of-the-art performances in learning tasks including natural language processing \citep{wolf2020transformers} and computer vision \citep{khan2022transformers}. However, the underlying mechanism for the success of Transformers remains largely a mystery to theoretical researchers.

It has been discussed in a line of recent works \citep{akyurek2022learning,zhang2023trained,bai2023transformers,huang2023context} that, instead of learning simple prediction rules (such as a linear model) Transformers are capable of learning to \textit{perform learning algorithms} that can automatically generate new prediction rules. For instance, when a new dataset is organized as the input of a Transformer, the model can automatically perform linear regression on this new dataset to produce a newly fitted linear model and make predictions accordingly. This idea of treating Transformers as ``algorithm approximators'' has provided insights into the power of large language models. 

However, these existing works only provide guarantees for the in-context supervised learning capacities of Transformers. It remains unclear whether Transformers are capable of handling unsupervised tasks as well. This intrigues our interest in providing theoretical foundations for such a model in the \textit{unsupervised learning} setup. Specifically, this work studies two unsupervised learning problems: \textit{Principal Component Analysis (PCA)} and \textit{Clustering Mixture of Gaussians}. PCA is the most fundamental data analysis method which finds the optimal low dimensional subspace that explains the most variance of the high dimensional data. PCA further gives rise to the class of spectral methods used in statistical inference, with the Clustering Mixture of Gaussians being one of the most significant applications.

\paragraph{Contributions.}
We summarize the contributions of this work as follows.
\begin{enumerate}
    \item This work provides formal theoretical guarantees for two unsupervised learning problems including the PCA and Clustering Mixture of Gaussians. Our results reveal that Transformers can learn from past experience and generalize toward unseen problems through the learning of algorithms. To the best of our knowledge, no existing works provide guarantees on un-supervised learning for pretrained Transformers, which is an important task in data analysis.
    
    \item The idea of our proof draws connections between statistical guarantees with algorithmic designs. In particular, we design algorithms specifically tailored to prove the approximation guarantees of Transformers. For the PCA problem, we consider drawing connections between the Power Method and the forward propagation of multi-layered Transformers. For the Clustering Mixture of Gaussians problem, we design a spectral method that can be approximated by Transformers. Moreover, our results also highlight that complicated algorithms can be dissected into multiple atomic algorithms. Then, the complete network can be decomposed into multiple sub-networks that implement individual atomic algorithms respectively.

    \item We perform extensive simulations to demonstrate that pre-trained Transformers can perform unsupervised learning tasks even when the assumptions leading to the theoretical results fail to hold. Our empirical evaluations are performed for both the synthetic data and real-world datasets. Our experiments also bridge the gap between assumptions in theory and practice.
\end{enumerate}

\subsection{Related Works}
We discuss the related literature in this section.
\paragraph{In-Context Learning of Transformers.} 
Some recent works studied the in-context learning (ICL) capacities of Transformers \citep{garg2022can, bai2024transformers}. In the in-context learning framework, the input is given by a sequence of token-label pairs $\{(\bfa X_i,y_i)\}_{i\in[N]}$ where there is an implicit function $f$ such that $y_i = f\lef(\bfa X_{[1:i]}\rig)$. The goal is to predict $y_{N+1}$ given the input of the complete episode. This learning setup resembles the language-generating mechanism of the LLMs and differs significantly from the standard supervised learning setup. \citep{bai2024transformers} considered the approximation and generalization properties of Transformers on the ICL tasks, including many linear regression and logistic regression setups. 

In contrast to the in-context learning setup, the problem considered in this work follows from unsupervised learning, where the predictor is constructed out of the complete episode of observations. Moreover, the unsupervised learning setup differs from the in-context learning through the lack of individual labels in the input matrix. For the proof machine, this work adopts similar ideas with \citep{akyurek2022learning, von2023transformers} who consider the approximation of transformers on gradient descent when performing ICL. However, the algorithms implemented in this work are more complicated and technical in their design.

Other related works on the more practical side of ICL can be found in \citep{dong2022survey} and reference therein.
\paragraph{Other Theoretical Works on Transformers.}
This paragraph reviews other theoretical works on the approximation of Transformers. \citep{yun2019transformers} studied the universal approximation properties of Transformers on sequence-to-sequence functions. \citep{perez2021attention,bhattamishra2020computational,liu2022transformers} studied the computational power of Transformers. \citep{hron2020infinite} studied the limit of infinite width multi/single head Attentions. \citep{yao2021self} showed that transformers can process bounded hierarchical languages and demonstrate better space complexity than recurrent neural networks. After a careful review of this line of work, we find that the unsupervised learning problem considered in this work does not fall into the general framework of our previous works.

\paragraph{Notations}
In this work, we follow the following notation conventions. The vector-valued variable is given by boldfaced characters. We denote $[n]:=\{1,\ldots,n\}$ and $[i:j]:=\{i,i+1,\ldots, j\}$ for $i<j$. The universal constants are given by $C$ and are ad hoc. For a vector $\bfa v$ we denote $\Vert\bfa v\Vert_2$ as its $L_2$ norm. For a matrix $\bfa A\in\bb R^{m\times n}$ we denote its operator norm as $\Vert\bfa A\Vert_2:=\sup_{\bfa v\in\bb S^{n-1}}\Vert\bfa A\bfa v\Vert_2$. Given two sequences $a_n$ and $b_n$, we denote $a_n\lesssim b_n$ or $a_n = O(b_n)$ if $\limsup_{n\to\infty}|\frac{a_n}{b_n}|<\infty$ and $a_n=o(b_n)$ if $\limsup_{n\to\infty}|\frac{a_n}{b_n}|=0$. 

\paragraph{Organizations} The rest of the paper is organized as follows: Section \ref{sect2} discusses the assumptions and results for the PCA problem; Section \ref{sect3.1} provides theoretical results for Clustering Mixture of Gaussians; Section \ref{sect4} provides extensive experimental details and results for both the synthetic and real-world datasets; Section \ref{sect5} discusses the limitations and future works. The detailed proofs and additional figures in experiments are delayed to the supplementary materials.

\section{Learning PCA}\label{sect2}
This section discusses how we construct a multi-layered Transformer model such that its forward propagation gives us the left principal eigenvectors of the input matrix. Our presentation is split into $4$ subsections: In section \ref{Tansformers} we review the mathematical forms of the Transformer model in this work; In section \ref{the power method} we review the classical Power Method algorithm to perform PCA and connect it with the multi-layered Transformer design; In section \ref{pretrain} we showcase how to perform supervised pre-training to achieve a model in section \ref{the power method}; In section \ref{sect3} we present the theoretical results.
\subsection{The Transformers}\label{Tansformers}
This section reviews the Transformer architecture considered in this work. We start by defining the Attention and the Fully Connected Layer (FC). Then we formally define both the architecture of the Transformers and its space of parameters. These definitions are similar to \citep{bai2024transformers}. 
\begin{definition}[Attention Layer]\label{def:attention}
    A self-Attention layer with $M$ heads is denoted as $Attn_{\bfa\theta_1}(\cdot)$ with parameters $\bfa\theta_1=\{(\bfa V_m,\bfa Q_m,\bfa K_m)\}_{m\in[M]}\subset\bb R^{D\times D}$. On an input sequence $\bfa H\in\bb R^{D\times N}$,
    \begin{align*}
       Attn_{\bfa\theta_1}(\bfa H)=
            \bfa H+\frac{1}{N}\sum_{m=1}^M(\bfa V_m\bfa H)\sigma\lef((\bfa Q_m\bfa H)^\top(\bfa K_m\bfa H)\rig),
    \end{align*}
    where $\sigma$ is the ReLU activation function.
\end{definition}
\begin{remark}
    We make the following simplification for Transformers:
   Instead of the concatenated feature given by multi-head Attention, we consider a simple average on the multi-head output; the activation function we considered is ReLU instead of Softmax which appears in most empirical works; we omit the layer-wise normalization used to stabilize the training procedure. We note that the approximation of the Softmax function is also doable with significant complication in the parameter design, following the representation idea in \citep{barron1993universal} instead of \citep{bach2017breaking}, which we left for future work. The other adjustments are mainly due to technical reasons. Recent works including \citep{shen2023study} empirically verified that ReLU Tansformers are strong alternatives to Softmax Tansformers. In section \ref{sect4} we carefully evaluate the effect of these additional features.
\end{remark}
The following defines the classical FC layers with residual connections.
\begin{definition}[FC Layer] A FC layer with hidden dimension $D^\prime$ is denoted as $FC_{\bfa\theta}(\cdot)$ with parameter $\bfa\theta_2\in(\bfa W_1,\bfa W_2)\in\bb R^{D^\prime\times D}\times\bb R^{D\times D^\prime}$. On any input sequence $\bfa H\in\bb R^{D\times N}$, we define
$$   FC_{\bfa\theta_2}(\bfa H):=\bfa H+\bfa W_2\sigma(\bfa W_1\bfa H).$$
\end{definition}
Then we use the above two definitions on the FC and the Attention layers to define the Transformers.
\begin{definition}[Transformer]
    We define a Tansformer $TF_{\bfa\theta}(\cdot)$ as a composition of self-Attention layers with FC layers. When the output dimension is $D$, an $L$-layered Transformer is defined by  $$   TF_{\bfa\theta}(\bfa H) := \tda W_0\times FC_{\bfa\theta_{2}^L}(Attn_{\bfa\theta_{1}^L}(\cdots FC_{\bfa\theta_{2}^1}(Attn_{\bfa\theta_{1}^1}(\bfa H)))\times\tda W_1,$$
    where $\tda W_0\in\bb R^{d_1\times D}$ and $\tda W_1\in\bb R^{N\times d_2}$.
\end{definition}
We use $\bfa \theta$ to denote the vectorization of all the parameters in the Transformer and the super-index $\ell$ to denote the parameter matrix corresponding to the $\ell$-th layer.
    Under these definitions, the parameter $\bfa\theta$ of the Transformer is given by 
    \begin{align*}
        \bfa \theta = \{\{(\{\bfa Q_m^\ell,\bfa K_m^\ell,\bfa V_m^\ell\}_{m\in[M]}, \bfa W_{1}^\ell,\bfa W_{2}^\ell)\}_{\ell\in[L]},\tda W_{0},\tda W_{1}\}.
    \end{align*}
\begin{remark}
    The model's Lipchitzness can be strictly governed by the following parameters: (1) The number of layers; (2) The number of heads; (3) The maximum operator norm of the parameters. These results further lead to an upper bound on the generalization error through the complexity of Lipschitz functions. The following defines the norm on the parameters of Transformers. 
    \begin{align*}
        \Vert\bfa\theta\Vert_{op}:=\max_{\ell\in[L]}\Big\{\max_{m\in[M^{\ell}]}\lef\{\Vert \bfa Q_m^\ell\Vert_2,\Vert\bfa K_m^\ell\Vert_2\rig\}+\sum_{m=1}^{M^{\ell}}\Vert\bfa V_m^\ell\Vert_2+\Vert\bfa W_1^{\ell}\Vert_2+\Vert\bfa W_2^{\ell}\Vert_2 +\Vert\tda W_0\Vert_2+\Vert\tda W_1\Vert_2\Big\}, 
    \end{align*}
    where $M^{\ell}$ is the number of heads of the $\ell$-th Attention layer.

    The two additional matrices $\tda W_0$ and $\tda W_1$ serve for the dimension adjustment purpose such that the output of $TF_{\bfa\theta}()$ will be of dimension $\bb R^{d_1\times d_2}$.
\end{remark}
\subsection{The Power Method}\label{the power method}
To show that the pre-trained Transformers are able to perform PCA for the input data, we show that a particular parameter construction of the Transformer can approximate algorithms. In particular, the approximation part of our theorem is proved by showing that multi-layered Transformers are able to approximate the classical Power Method used for Singular Value Decomposition \citep{golub2013matrix}. The formal algorithm is given by \ref{alg:almoexactrecov}.

\begin{algorithm}[ht]
 \caption{Power Method for the Left Singular Vectors}
\label{alg:almoexactrecov}
\KwData{Matrix $\bfa X\in\bb R^{d\times N}$, Number of Iterations $\tau$, Number of Eigenvectors $k$.}
Symmertize $\bfa A=\bfa X \bfa X^\top\in\bb R^{d\times d}$ \;
Let the set of eigenvectors be $\ca V=\{\}$. Initialize $\bfa A_{1}\gets \bfa A$\;
\For{$\ell\gets 1$ to $k$}{
Sample a random vector $\bfa v_{0,\ell}\in\bb S^{N-1}$. Initialize $\bfa v_{\ell}^{(0)} \gets \bfa v_{0,\ell}$\;
\For{$t\gets 1$ to $\tau$}{
Apply the procedure to obtain the principal eigenvector
$    \bfa v_\ell^{(t)}=\frac{\bfa A_{\ell}\bfa v_{\ell}^{(t-1)}}{\Vert \bfa A_\ell\bfa v_{\ell}^{(t-1)}\Vert_2}$\;
}
Let $\ca V\gets\ca V\cup\{\bfa v_{\ell}^{(\tau)}\}$\;
Compute the eigenvalue estimate $\wh\lambda_{\ell}\gets \Vert \bfa A_{\ell}\bfa v_{\ell}^{(\tau)}\Vert_2$\;
Update the matrix by $\bfa A_{\ell+1}=\bfa A_{\ell}-\wh\lambda_{\ell}\bfa v_{\ell}^{(\tau)}\bfa v_{\ell}^{(\tau),\top}$ \;
}
 \Return $\ca V$\;
\end{algorithm}
    \begin{figure}[htbp]
  \includegraphics[width=\linewidth]{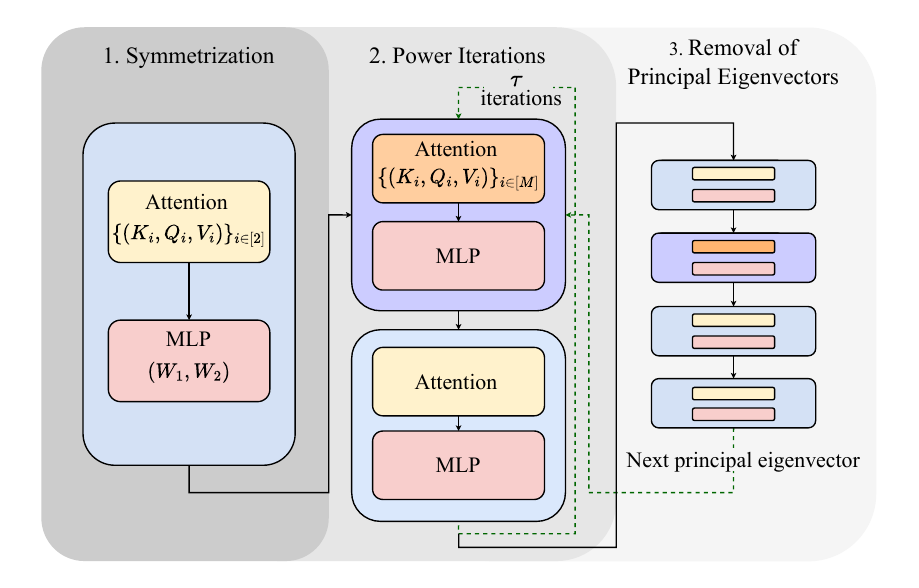}
\caption{
\textbf{The Constructive Proof for the Approximation of PCA.}
The above diagram illustrates the construction of our Transformer model in the existence proof of theorem \ref{thm3.1}. There are three important sub-networks in the design: \emph{(1) The Symmetrization }sub-network symmetrizes $\bfa X$ and stamp $\bfa X\bfa X^\top$ in the output, which corresponds to the first step in the Power Method. \emph{(2) The Power Iterations }sub-network performs in a total of $\tau$ iterations for each of the principal eigenvectors, corresponds to the iterative update step in the Power Method. \emph{(3) The Removal of principal Eigenvectors} subnetwork performs the estimate of $\wh\lambda_{\ell}$ and the update of matrix $\bfa A_{\ell}$ in the Power Method. Finally, we apply $\tda W_0$ and $\tda W_1$ to adjust the dimension of the output. The different colors in the diagram correspond to the different types of layers: \emph{{(1) Yellow Blocks}} denote the Attention layer with 2 heads. \emph{{(2) Orange Blocks}} denote the multihead Tansformers with larger $M\gg 2$.  \emph{{(3) Pink Blocks}} denote the FC layer.
}\label{fig:diagram}
\end{figure}

The algorithm first generates the symmetrized covariate matrix $\bfa A=\bfa X \bfa X^\top$. Then, for each eigenvector that we hope to recover, the Power Method generates a random vector uniformly distributed on the unit sphere. Then, we iteratively take the matrix product between the symmetric matrix $\bfa A$ and $\bfa v_{0,\ell}$, followed by normalizing it. Note that such an iterative matrix product can finally converge to the top-1 principal eigenvector.

The iterative structure is akin to the forward propagation of the Transformer architecture. Given such similarities, we show that there exists a parameter setup for Transformers such that the forward propagation performs PCA. The challenge is in constructing parameters that approximate the complete procedure of the Power Method.

In figure \ref{fig:diagram}, we provide an \emph{approximate} algorithm for the Power Method through the lens of a forward propagation on the Transformer. This algorithm dissects the approximation of the Power Method into the propagation along multiple sub-networks, each phase corresponding to a single step in the Power Method. Combining them, we show in section \ref{sect3} that Transformers achieve good approximation guarantees.

\subsection{Pretraining via Supervised Learning}\label{pretrain}
The standard PCA problem is unsupervised where no labels are given. However, the Transformers are usually used in the supervised learning setup. To make full use of Transformers in the PCA task, we need to perform supervised pre-training. In our theoretical analysis, we construct the input of the Transformer as a \emph{context-augmented matrix} given by the following
\begin{align}\label{HPequations}
    \bfa H=\begin{bmatrix}
        \bfa X\\
        \bfa P
    \end{bmatrix}\in\bb R^{D\times N},\qquad \bfa P=\begin{bmatrix}
        \tda p_{1,1},\ldots,\tda p_{1,N}\\
        \tda p_{2,1},\ldots,\tda p_{2,N}\\
        \vdots\\
        \tda p_{D-d,1},\ldots,\tda p_{D-d, N}
    \end{bmatrix}\in\bb R^{(D-d)\times N},
\end{align}
where the matrix $\bfa P$ contains contextual information, which is specified in section \ref{sect3}. The design also makes sure $\bfa P$ is unrelated to $\bfa X$. In the experiments, we show that the auxiliary matrix $\bfa P$ is not necessary for the pre-trained Transformer to perform PCA with high accuracy. For the output, our theoretical analysis gives the following matrix
\begin{align*}
    TF_{\bfa\theta}(\bfa H)=\begin{bmatrix}
        \wha v_1^\top&\ldots&\wha v_k^\top
    \end{bmatrix}^\top\in\bb R^{d\times k}
\end{align*}
which corresponds to the estimated principal eigenvectors of the matrix $\bfa X$.

\paragraph{The Learning Problem.} Consider a set of samples $\{\bfa X^{(i)}\}_{i\in[n]}$ i.i.d. sampled from some distribution $p_{\bfa X}$, we construct their oracle top-$k$ principal components as $\bfa V^{(i)}=\begin{bmatrix}
    \bfa v^{i,\top}_1&\ldots&\bfa v_k^{i,\top}
\end{bmatrix}^\top$
and the context-augmented input matrix as $\bfa H_i$ for each $\bfa X_i$. Then, the pretraining procedure is given by minimizing the following objective for some convex loss function $L(\cdot,\cdot):\bb R^{d\times k}\times \bb R^{d\times k}\to\bb R$,
\begin{align}\label{ERM}
 \wha\theta =\argmin_{\bfa\theta\in\Theta(B_{\bfa\theta}, B_M)}\sum_{i=1}^nL(TF_{\bfa\theta}(\bfa H^{(i)}), \bfa V^{(i)}).
\end{align}
Here we consider $\Theta(B_{\bfa\theta}):=\{\bfa\theta:\Vert\bfa\theta\Vert\leq B_{\bfa\theta}, \max_{\ell}M^{\ell}\leq B_M\}$ to be the space of parameters. We also consider guarantees when $L(\bfa x_1,\bfa x_2):=\Vert\bfa x_1-\bfa x_2\Vert_2$ in the theory. Despite the problem is non-convex and global minima are not computationally feasible to achieve, the ERM solution still reveals the strong capacities of Transformers in statistical inference tasks.
We further show that the local minimizers obtained through stochastic gradient descent achieve good empirical performance in section \ref{sect4}.
\subsection{Theoretical Results}\label{sect3}
This section presents our theoretical results. Our proof goes by constructing a particular instance of the transformers and shows that the forward propagation on our constructed instance approximates the power method. For this construction, we also carefully design the auxiliary matrix $\bfa P$ explained as follows.

\paragraph{The Design of Auxillary Matrix.} Our design of the matrix $\bfa P$ consists of three parts:
\begin{enumerate}
    \item \emph{Place Holder.} For $\ell\in\{1\}\cup[4:k+3]$ and $i\in[N]$, we let $\tda p_{\ell, i}=\bfa 0\in\bb R^{d\times 1}$. The placeholders in $\bfa P$ record the intermediate results in the forward propagation. Recall that $k$ is the number of eigenvectors we hope to recover.
    \item \emph{Identity Matrix.} We let $ \begin{bmatrix}
        \tda p_{2,1}&\ldots&\tda p_{2,N}
    \end{bmatrix}=\begin{bmatrix}
        \bfa I_d&\bfa 0_{d\times(N-d)}
    \end{bmatrix}$. The identity matrix in $\bfa P$ helps us screen out all the covariates $\bfa X$ in the forward propagation.
    \item \emph{Random Samples on the Hypersphere.} We let $\tda p_{3,1},\ldots\tda p_{3,k}$ be the i.i.d. samples uniformly distributed on $\bb S^{d-1}$. The random samples on the sphere correspond to the initial vectors $\bfa v_{0,\ell}$ for $\ell\in[k]$ in algorithm \ref{alg:almoexactrecov}.
\end{enumerate}
The auxiliary matrix is designed for purely technical reasons. Moreover, our experiments suggest that such an auxiliary matrix is not necessary for the task, which is detailed in section \ref{sect4}.
Given the above construction on the auxiliary matrix $\bfa P$, we are ready to state the approximation theorem, given as follows.

\begin{theorem}[Transformer Approximation of the Power Method]\label{thm3.1} Assume that the eigenvalues of $\bfa X\bfa X^\top$ to be $\lambda_1>\lambda_2>\ldots>\lambda_k>\ldots$. Let $\Delta:=\min_{1\leq i<j\leq k}|\lambda_i-\lambda_j|$. Assume that the initialized vectors $\tda p_{3,1},\ldots\tda p_{3,N}$ 
    satisfy  $\tda p_{3,i}^\top\bfa v_{i}\geq\delta$ for all $i\in[k]$ and make the rest of the vectors $\bfa 0$. 
    Then, there exists a Transformer model with the number of layers $L=2\tau+4k+1$ and the number of heads $B_M\leq \lambda_1^d\frac{C}{\epsilon^2}$ with $\tau\leq\frac{\log(1/\epsilon_0\delta)}{\epsilon_0}$ such that for all $\epsilon_0,\epsilon>0$,
    the final output $\begin{bmatrix}
        \wha v_1,\ldots,\wha v_{k}
    \end{bmatrix}$ given by the Transformer model achieves
    \begin{align*}
        \lef\Vert\wha v_{\eta+1}-\bfa v_{\eta+1}\rig\Vert_2\leq C\tau \epsilon\lambda_1^2+\frac{C\lambda_1\sqrt{\epsilon_0}}{\Delta}\prod_{i=1}^{k}\frac{5\lambda_{i+1}}{\Delta}.
    \end{align*} 
    Moreover, consider the accuracy of multiple $\bfa v$s as a whole. There exists $\bfa\theta\in\Theta(B_{\lambda_1}, B_M)$ that satisfies
    \begin{align*}
        L\lef(TF_{\bfa\theta}(\bfa H),\bfa V\rig)\leq C\tau\epsilon k\lambda_1^2+C\bl\frac{\epsilon_0\lambda_1^2}{\Delta^2}\sum_{\eta=1}^{k-1}\prod_{i=1}^{\eta}\frac{25\lambda_{i+1}^2}{\Delta^2}\br^{1/2}.
    \end{align*}
\end{theorem}
\begin{remark}
\label{remark3}
    The approximation error consists of two terms. The first term comes from the approximation of the Power Method iterations by Transformers. The second term comes from the error caused by finite iteration $\tau$ in the power method. To get an intuition of the error terms and its order of magnitude, we consider a special case where the eigenvalues $\lambda_1\asymp\lambda_2\asymp\ldots\asymp\lambda_k\asymp\Delta$. Then our results boil down to 
    \begin{align*}
        \lef\Vert TF_{\bfa\theta}(\bfa H)-\begin{bmatrix}
            \bfa v_1^\top,\bfa v_2^\top,\ldots,\bfa v_k^\top
        \end{bmatrix}^\top\rig\Vert_2\lesssim \tau\epsilon k \lambda_1^2+ \frac{\lambda_1}{\Delta}\sqrt{k\epsilon_0}.
    \end{align*}
    These results hide dimension $d$ in the universal constant. We note that the dimension significantly affects the approximation bound of Transformers. This is mainly due to the limitations given by approximating high dimensional functions by ReLU neural networks.
\end{remark}
In the above theorem, our results rely on the random initialization of $\bfa P$. We show that the conditions on $\tda p_{3,1},\ldots,\tda p_{3,N}$ can be achieved through sampling from isotropic Gaussians, given by the following lemma.
    \begin{lemma}\label{lm3.1.1}\label{lm3.1}
        Consider $\bfa x$ to be a random vector sampled uniformly at random on $\bb S^{d-1}$. Let $\bfa v$ be any unit length vector, then we have for all $\delta<\frac{1}{2}d^{-1}$,
           $ \bb P\lef(|\bfa v^\top\bfa x|\leq\delta\rig)\leq  \frac{1}{\sqrt{\pi}}\sqrt\delta+\exp\lef(-C\delta^{-\frac{1}{2}}\rig)$.
        Therefore, for all $\delta<\frac{1}{2}d^{-1}$, the event in theorem \ref{thm3.1} is achieved with
        \begin{align*}
             \bb P\bl\exists i\in[k]\text{ such that }&\bfa x_i^\top\bfa v_i\leq\frac{\delta}{\sqrt d}\br\leq \frac{k\sqrt{\delta}}{\sqrt{\pi}}+k\exp(-C\delta^{-1}).
        \end{align*}
    \end{lemma}
Given the approximation error provided by theorem \ref{thm3.1}, we further provide the generalization error bound for the ERM defined by \eqref{ERM}. This requires us to consider the following regularity conditions on the underlying distribution of $\bfa X\bfa X^\top$ (which also translates to the distribution for the pre-training instances $\bfa X$).
\begin{assumption}[Problems for Pre-training]\label{assump1}
    The distribution of $\bfa X\bfa X^\top$ supports on the following space $$\bb X:=\lef\{\bfa A: \bfa A\in\bfa S^d_{++},  B_X\geq\lambda_1(\bfa A)>\lambda_2(\bfa A)>\ldots>\lambda_k(\bfa A),\inf_{1\leq i<j\leq k}\lambda_i(\bfa A)-\lambda_j(\bfa A)\geq\Delta\rig\}.$$
\end{assumption}
\begin{remark}
    The above assumption can be generalized to a distribution that supports $\bb X$ with high probability. Examples of such distribution include the Wishart distribution under the Gaussian design, which is elaborated in the Gaussian mixture model example in section \ref{sect3.1}.
\end{remark}
Given the above assumption, we are ready to state the generalization bound.
\begin{proposition}\label{genbound}
    Under assumption \ref{assump1} and  using the notations given by theorem  \ref{thm3.1}, with probability at least $1-\xi$, the ERM solution $\wha\theta$ satisfies
    \begin{align*}
        \bb E\lef[L\lef(TF_{\wha\theta}(\bfa H),\bfa V\rig)\Big|\wha\theta\rig]&\leq\inf_{\bfa\theta\in\Theta(B_{\bfa\theta},B_M)}\bb E\lef[L\lef(TF_{\bfa\theta}(\bfa H),\bfa V\rig)\rig]\\
        &+ C \sqrt{\frac{k^3LB_Md^2\log(B_X+k)+\log(1/\xi)}{n}},
    \end{align*}
    where the expectation is taken over the new sample $\bfa X$.
\end{proposition}
    \begin{corollary}
        Under assumption \ref{assump1}, with probability at least $1-\xi-\frac{k\sqrt\delta}{\sqrt\pi}-k\exp\lef(-C\delta^{-1/2}\rig)$ for all $\delta<d^{-1}$ we have for all $\epsilon,\epsilon_0>0$,
        \begin{align*}
           L_{PCA}(\wha\theta,\bfa P):&=\bb E\lef[L\lef(TF_{\wha\theta}(\bfa H),\bfa V\rig)\Big|\wha\theta,\bfa P\rig]\lesssim \tau\epsilon k\lambda_1^2+\bl\frac{\epsilon_0\lambda_1^2}{\Delta^2}\sum_{\eta=1}^{k-1}\prod_{i=1}^{\eta}\frac{25\lambda_{i+1}^2}{\Delta^2}\br^{1/2}\\
            &+\sqrt{\frac{k^3\log(\delta/
            \epsilon_0)B_X^d d^2\log(B_X+k)+\log(1/\xi)}{n\epsilon_0\epsilon^2}}.
        \end{align*}
    \end{corollary}
    \begin{remark}
        If we consider optimizing the bound w.r.t. $\epsilon_0$ and $\epsilon$, we obtain that  $L_{PCA}(\wha\theta,\bfa P)\lesssim n^{-1/5}$ with high probability, given that the parameters and the dimension $d$ are of constant scales.
    \end{remark}

\section{Clustering a Mixture of Gaussians}\label{sect3.1}
Based on the results obtained in section \ref{sect2}, this section provides theoretical results for clustering a mixture of Gaussians, a problem that can be solved by spectral algorithms \citep{kannan2009spectral} using PCA as a sub-procedure. However, existing literature always requires additional algorithms to accomplish this task, which are not easily implementable by Transformers. To address this limitation, we design a simple spectral algorithm that is simple to implement by the Transformer infrastructure. Our results in this section demonstrate that Transformers can memorize a much larger class of inference algorithms with PCA used as sub-procedures. In section \ref{mixture} we review the problem setup. In section \ref{gmm:theoretical} we present the algorithms and theoretical results for Transformer learning.

\subsection{Problem Setup}\label{mixture}
In the problem of clustering two class Gaussian mixture model, we are given a total of $N$ i.i.d. samples $\{\bfa X_i\}_{i\in[N]}$ from a spherical Gaussian mixture model with two centers $\bfa \mu_0$ and $\bfa\mu_1$. We consider the data generating procedure of $\bfa X_i$ as first flipping a fair coin and if a tail appears we assign $\bfa X_i$'s cluster to be $0$ and generate it from $N(\bfa\mu_0, I_d)$. Else, we assign $\bfa X_i$'s cluster to be $1$ and generate it from $N(\bfa\mu_1, I_d)$. Denote the cluster membership as $z\in\{0,1\}^N$. The conditional p.d.f. of $\bfa X_i$ is given as follows:
\begin{align*}
    \bb P(\bfa X_i=\bfa x|z_i=\alpha)=\frac{1}{(2\pi)^{\frac{d}{2}}}\exp\Big(-\frac{1}{2}\Vert\bfa x-\bfa\mu_{\alpha}\Vert_2^2\Big),\quad\alpha\in\{0,1\}.
\end{align*}
For the task of clustering mixtures of 2 GMMs, we use the following loss function
\begin{align}\label{gmmloss}
    L_{GMM}(\wh z, z) = \min_{\tde z\in\{z,\mbbm 1-z\}}\frac{1}{N}\sum_{i=1}^N \Vert\wh z_i-\tde z_i\Vert_1.
\end{align}
We are interested in the statistical guarantees for the ERM solution of the above problem given in a total of $n$ pretraining instances $\{\bfa X^{(i)}\}_{i\in[n]}$ and $\{z^{(i)}\}_{i\in[n]}$. We denote the estimator given by the Transformer as $TF_{\bfa\theta}(\bfa H^{(i)})$ for the $i$-th pretraining instance. Then the ERM estimator is given by
\begin{align}\label{gmmerm}
    \wha\theta_{GMM} :=\argmin_{\bfa\theta}\sum_{i=1}^nL_{GMM}\lef(TF_{\bfa\theta}(\bfa H^{(i)}),z^{(i)}\rig).
\end{align}

\subsection{Theoretical Results}\label{gmm:theoretical}
We develop algorithm \ref{alg:gmm} to prove the theoretical guarantees for ERM solution given by \eqref{gmmerm}. In comparison to the algorithms proposed in the literature \citep{kannan2009spectral}, our algorithm does not require extra procedures. Instead, algorithm \ref{alg:gmm} approximates the Bayes estimator in this problem, whose decision boundary is given by the hyperplane orthogonal to $\bfa\mu_1-\bfa\mu_0$ and passes through the point $\frac{1}{2}(\bfa\mu_1+\bfa\mu_0)$. To avoid the difficulties of analyzing the performance of the post-selection estimator, we first split the data into the first $N_1$ columns and the rest $N-N_1$ columns. We denote the mean vector $\bar{\boldsymbol{X}}:=\frac{1}{N_1}\sum_{i=1}^{N_1}\bfa X_i$. Then we utilize the empirical estimate of the population covariance matrix given by the first $N_1$ columns as 
\begin{align}\label{empiriclcov}
    \wh\Sigma = \frac{1}{N_1}\sum_{i=1}^{N_1}\lef( \bfa X_i-\bar{\boldsymbol{X}}\rig)\lef( \bfa X_i-\bar{\boldsymbol{X}}\rig)^\top.
\end{align}
It is not hard to note that the principal direction of the population matrix is exactly $\bfa\mu_0-\bfa\mu_1$. We therefore project all the rest of the samples to the direction given by the first principal component $\wha v_1$ of $\wh\Sigma$ and compare it with the midpoint $\bfa v_1^\top\lef(\frac{1}{N_1}\sum_{i=1}^{N_1}\bfa X_i\rig)=\bfa v_1^\top\bar{\boldsymbol{X}}$.

\begin{algorithm}[ht]
 \caption{Clustering Bi-class GMM}
\label{alg:gmm}
\KwData{Matrix $ \bfa X\in\bb R^{d\times N}$, Number of Iterations $\tau$}
Consider the estimated sample covariance matrix given by \eqref{empiriclcov}\;

Apply algorithm \ref{alg:almoexactrecov} on $\wh\Sigma$ and obtain its first principal component $\wha v_1$\;
Project the $N$ samples on to the direction $\wha v_1$ by $\tde z_i = \bfa X_i^\top\wha v$ and cluster based on the sign for $i\in[N]$\;
\end{algorithm}

Before we provide formal guarantees for the Transformer solution, we provide the following theorem for algorithm \ref{alg:gmm}.
\begin{theorem}\label{thm411}
    Let $N_1\asymp d^{\frac{1}{3}}N^{\frac{2}{3}}(\Vert\bfa\mu_1-\bfa\mu_0\Vert_2+\log N)^{\frac{2}{3}}$. Assume that $\Vert\bfa\mu_1-\bfa\mu_0\Vert_2\gtrsim \sqrt{\log N/d}$. Then algorithm \ref{alg:gmm} achieves the following upper bound
    \begin{align*}
        \bb E\Big[L_{GMM}\lef(\wh z,z\rig)\Big]\lesssim \bl\frac{d\log^2N}{N}\br^{\frac{1}{3}}.
    \end{align*}
\end{theorem}
\paragraph{Transformer Approximation.}
We construct weights for a multi-layered  Transformer whose forward propagation approximates algorithm \ref{alg:gmm}. Our first result demonstrates the approximation error of the Transformer model to that of algorithm \ref{alg:gmm} as follows.
\begin{lemma}\label{lm:4.1}
    Consider input with the auxiliary matrix defined by \eqref{auxillarymat}. There exists a Transformer with number of layers $L=2\tau+7$, number of heads $B_M\leq \lambda_1^d\frac{C}{\epsilon^2}$ and $\tau\leq \frac{\log(1/\epsilon_0\delta)}{\epsilon_0}$ such that with probability at least $1-\frac{\sqrt{\delta}}{\sqrt{\pi}}-\exp(-C\delta^{-1})$ there exist parameters $\bfa\theta$ such that the output of this Transformer satisfies
    \begin{align*}
        TF_{\bfa\theta}(\bfa H) = \begin{bmatrix}
               \tanh\lef(\beta\wha v_1^{TF,\top}(\bfa X_1-\bar{\boldsymbol{X}})\rig)&\ldots&\tanh\lef(\beta\wha v_1^{TF,\top}(\bfa X_N - \bar{\boldsymbol{X}})\rig)
            \end{bmatrix},
    \end{align*}
    where $\Vert\wha v_1^{TF}-\wha v_1\Vert_2\lesssim\frac{\log(1/\epsilon_0\delta)}{\epsilon_0}\epsilon\lambda_1^2+\frac{\lambda_1\sqrt{\epsilon_0}}{\Delta}+\epsilon_0$. 
    \end{lemma}

For technical purposes, in our proof we consider auxiliary matrix $\bfa P$ appearing in \eqref{HPequations} to be designed as
\begin{align}\label{auxillarymat}
   \bfa P^{GMM}:= \begin{bmatrix}
        &\bfa 0_{d\times N}&\\
        \tda p_{1,1}&\ldots &\tda p_{1,N}\\
        \tda p_{2,1}&\ldots &\tda p_{2,N}\\
        &\vdots&\\
        \tda p_{\ell,1}&\ldots&\tda p_{\ell, N}
    \end{bmatrix}.
\end{align}
where instead of using $\tda p_{1,i}\in\bb R^d$ and $\tda p_{1,i,j}:=\mbbm 1_{i=j, j\leq d}$ in the proof of theorem \ref{thm3.1} we consider using $\tda p_{1,i}\in\bb R^{N_1}$ and $\tda p_{1,i,j}:=\mbbm 1_{i=j,j\leq N_1}$. Then we consider the following assumption characterizing the space.
\begin{assumption}[Problems for Pre-training]\label{assump2}
    We assume that the pre-training instances $\{\bfa\mu_1^{(i)},\bfa\mu_0^{(i)}, N^{(i)}\}_{i\in[n]}$ are sampled from an arbitrary distribution supported on the following space
    \begin{align*}
       \Theta(B_{\bfa\mu}, B_N):=\lef\{\bfa\mu_1,\bfa\mu_0,N:\sqrt{\log(N/d)}\lesssim\Vert\bfa\mu_1-\bfa\mu_0\Vert_2\leq B_{\bfa\mu}, N\asymp B_N\rig\}.
    \end{align*}
\end{assumption}
Then, the following theorem provides a theoretical guarantee for the expected error of $\wha\theta_{GMM}$. 
\begin{theorem}\label{thm3.2}
    Under assumption \ref{assump2} and use $n$ i.i.d. instances $\{(\bfa X^{(i)} z^{(i)})\}_{i\in[n]}$ to pretrain the transformer. Then we take $\epsilon_0 = \frac{1}{\sqrt{\epsilon
        }}$ and $\epsilon=\lef(n^{-\frac{1}{2}}B_{\bfa\mu}^{d-2}d(\log B_{\bfa\mu})^{\frac{1}{2}}\rig)^{\frac{4}{7}}$. The ERM solution $\wha\theta_{GMM}$ satisfies
    \begin{align*}
            \bb E[L_{GMM}(&TF_{\wha\theta_{GMM}}(\bfa H),z)]\lesssim \Big(\frac{d\log^2N}{N}\Big)^{\frac{1}{3}} + B_{\bfa\mu}^{\frac{2d + 8}{7}} d^{\frac{2}{7}}n^{-\frac{1}{7}}(\log B_{\bfa\mu})^{\frac{1}{7}}\beta^{\frac{4}{7}}.
        \end{align*}
\end{theorem}
\begin{remark}
    Our theorem suggests that the final learning error of Transformers in clustering the mixture of Gaussians can be characterized by two terms: (1) The first term being oracle error given by theorem \ref{thm411}; (2) The second term being the error incurred by the pre-training procedure. It is also noted that for a sufficiently large pre-training set where $n\gtrsim \lef(\frac{d\log^2N}{N}\rig)^{\frac{7}{3}}B_{\bfa\mu}^{{2d + 8}} d^{{2}}(\log B_{\bfa\mu})\beta^{{4}}$ we have the final learning bound matching that of theorem \ref{thm411}. We further discuss the improvement of the rates in section \ref{sect5}.
\end{remark}

\section{Simulations}\label{sect4}

\begin{figure}[t]
    \centering
    \minipage{0.33\textwidth}
        \includegraphics[width=\linewidth]{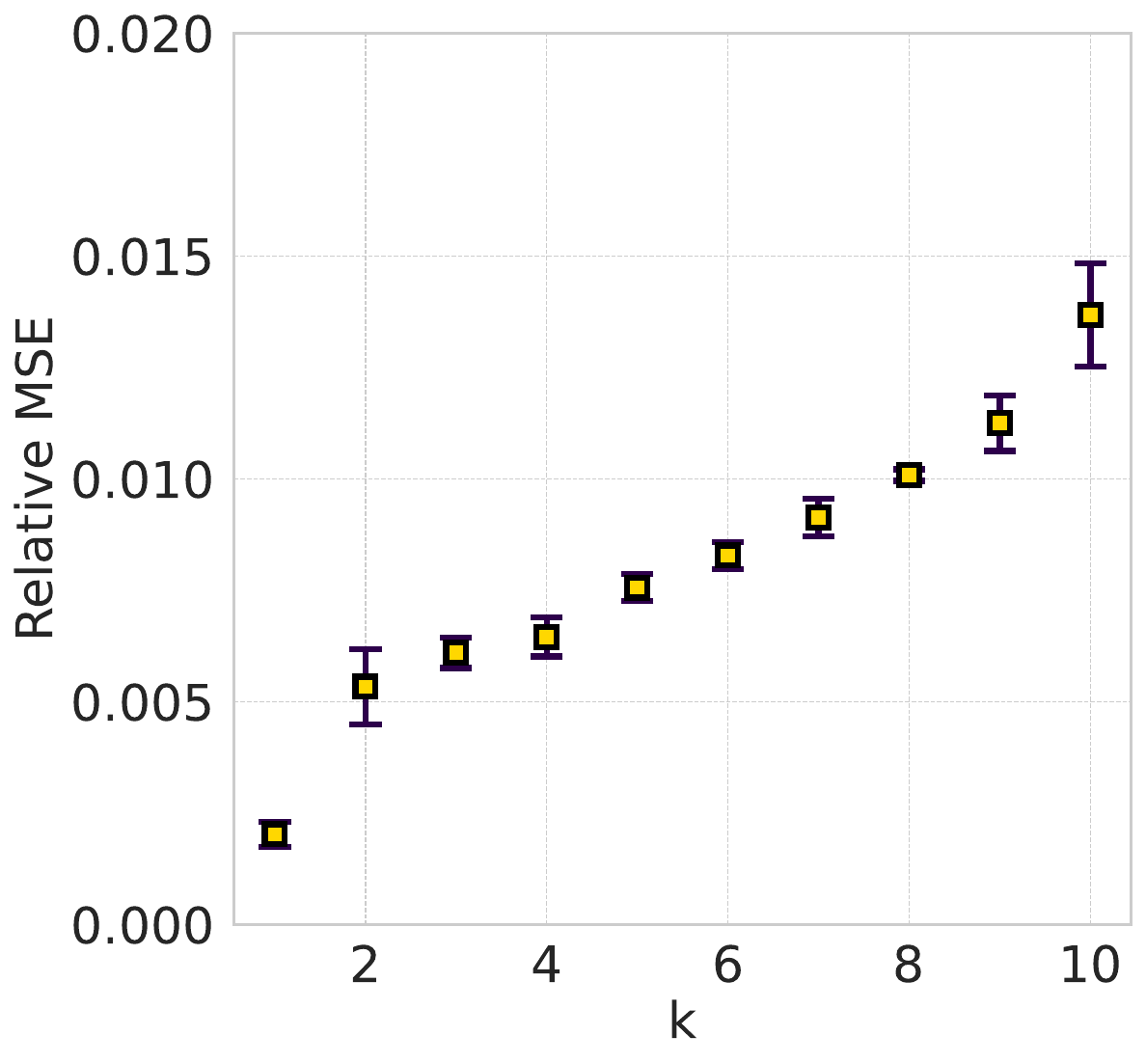}
    \endminipage\hfill
    \minipage{0.33\textwidth}
        \includegraphics[width=\linewidth]{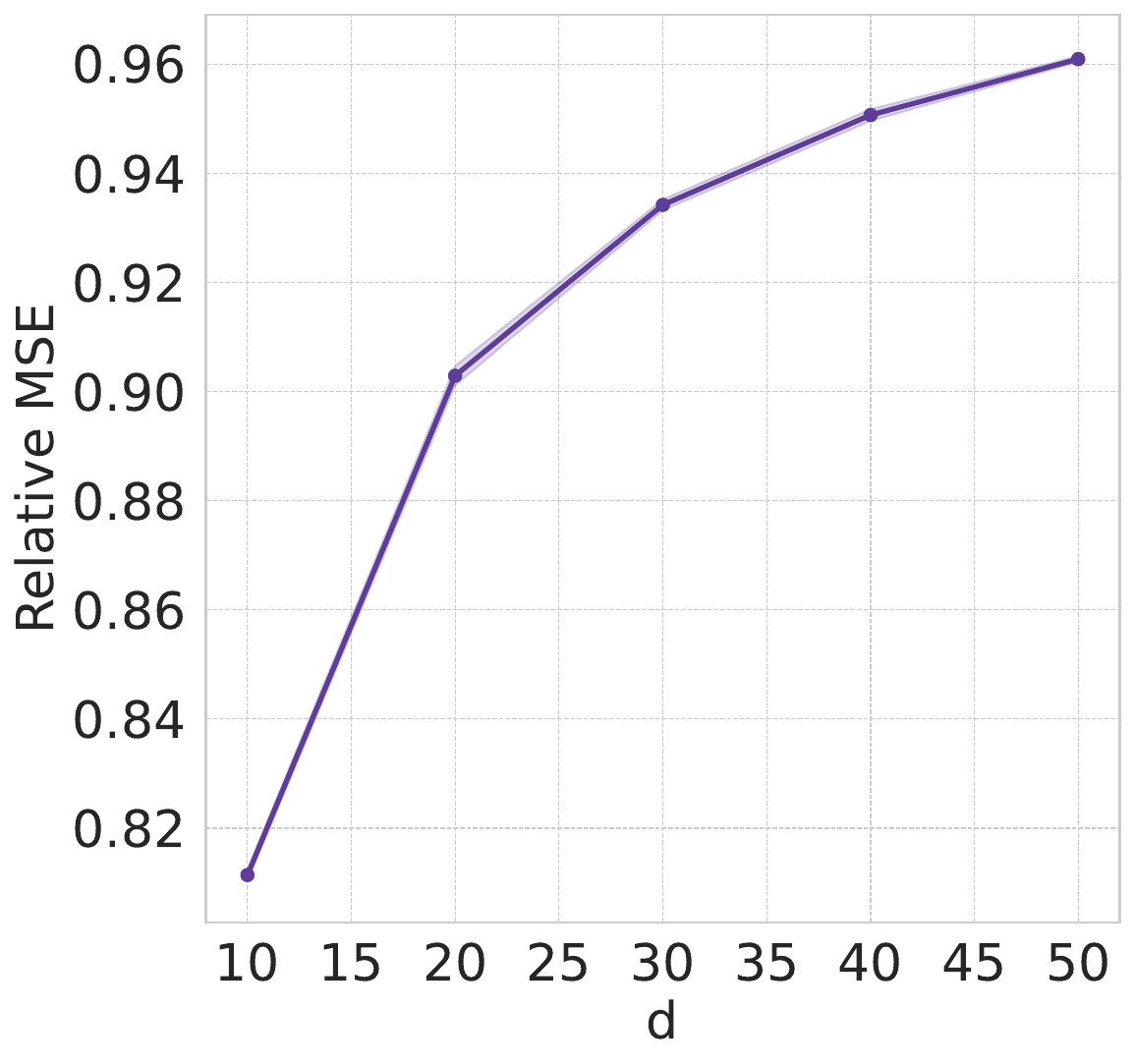}
    \endminipage\hfill
    \minipage{0.33\textwidth}
        \includegraphics[width=\linewidth]{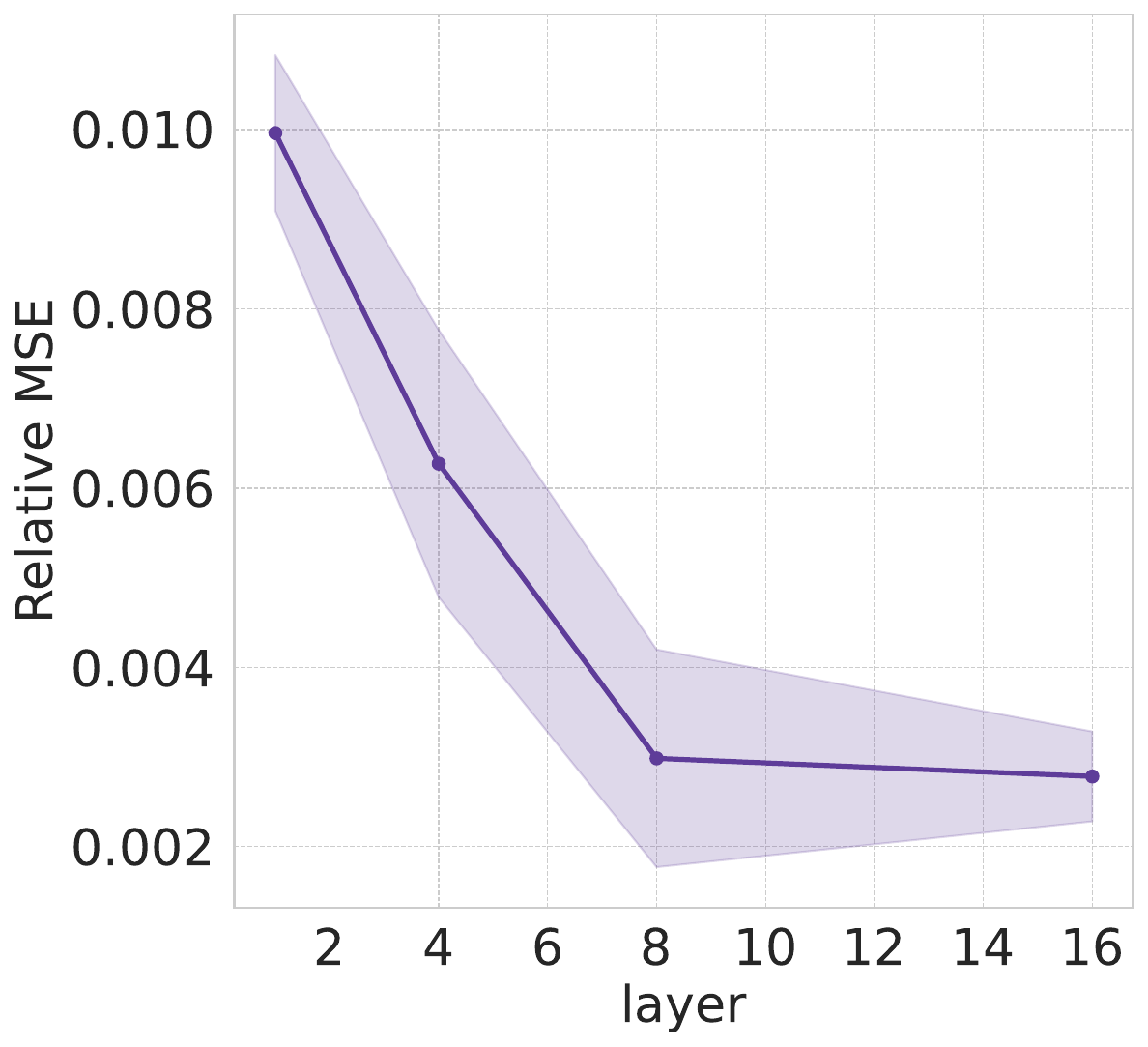}
    \endminipage
    \vspace{-1em}
    \caption{{\textbf{Eigenvalue Prediction on Synthetic Data.} 
    \emph{(1) Left: Evidence of Transformer's Ability to Predict Multiple Eigenvalues.} 
    We use a small Transformer (layer $=3$, head $=2$, embedding $= 64$) to predict top $10$ eigenvalues with $d=20$ and $N=50$.
    \emph{(2) Middle: Predictions of Eigenvalues with Different Input Dimension $d$.} We use a small Transformer and use $N=10$ in this experiment. 
    \emph{(3) Right: Predictions of Eigenvalues with Different Number of Layers.} We use the same input as the previous multiple eigenvalues predictions experiment, and use a small Transformer to predict top-$3$ eigenvalues.} 
    }
    \label{fig:encoder_eigenvalues}
\end{figure}

\begin{figure}[h]
    \centering
    \minipage{0.33\textwidth}
        \includegraphics[width=\linewidth]{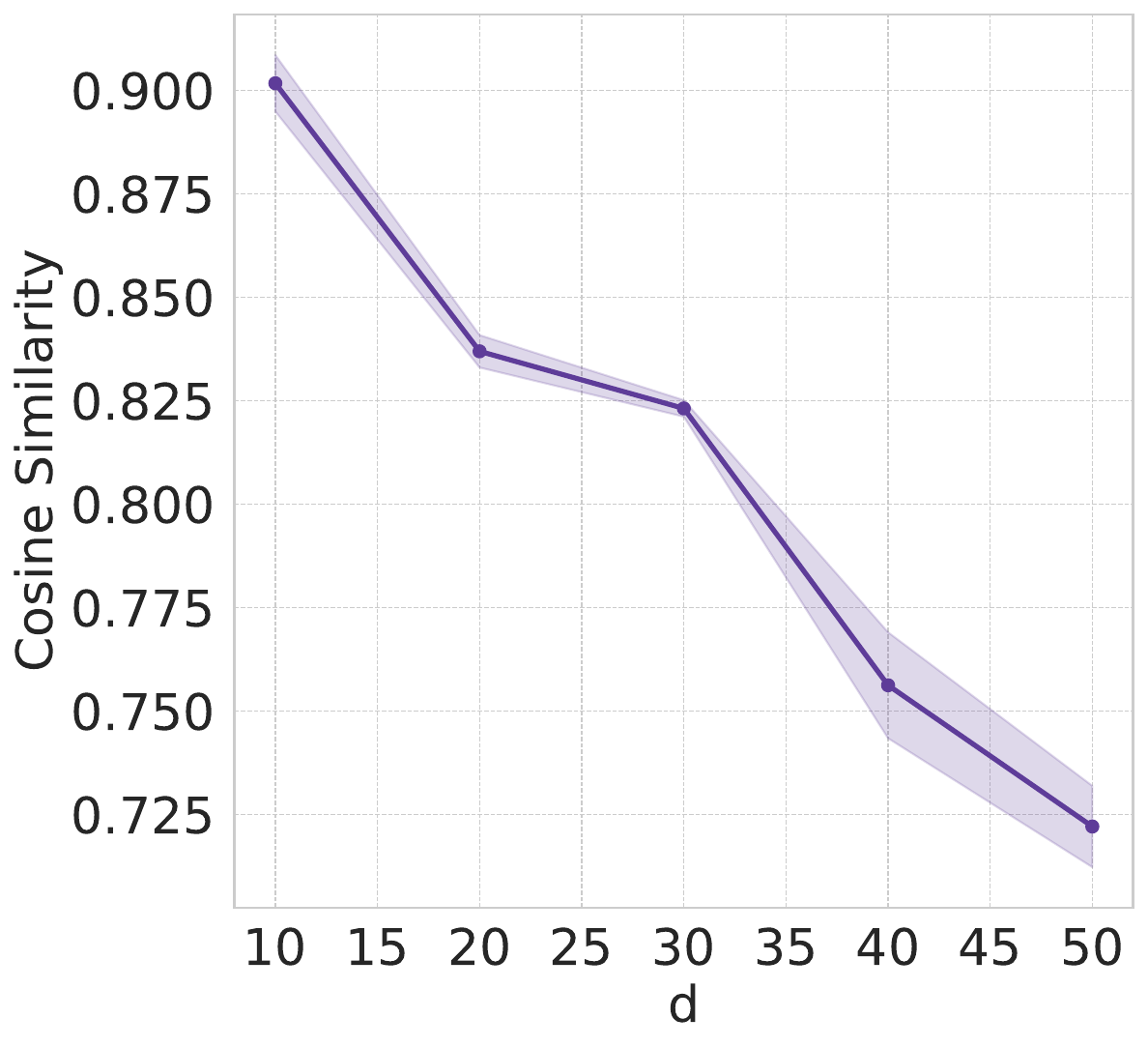}
    \endminipage\hfill
    \minipage{0.33\textwidth}
        \includegraphics[width=\linewidth]{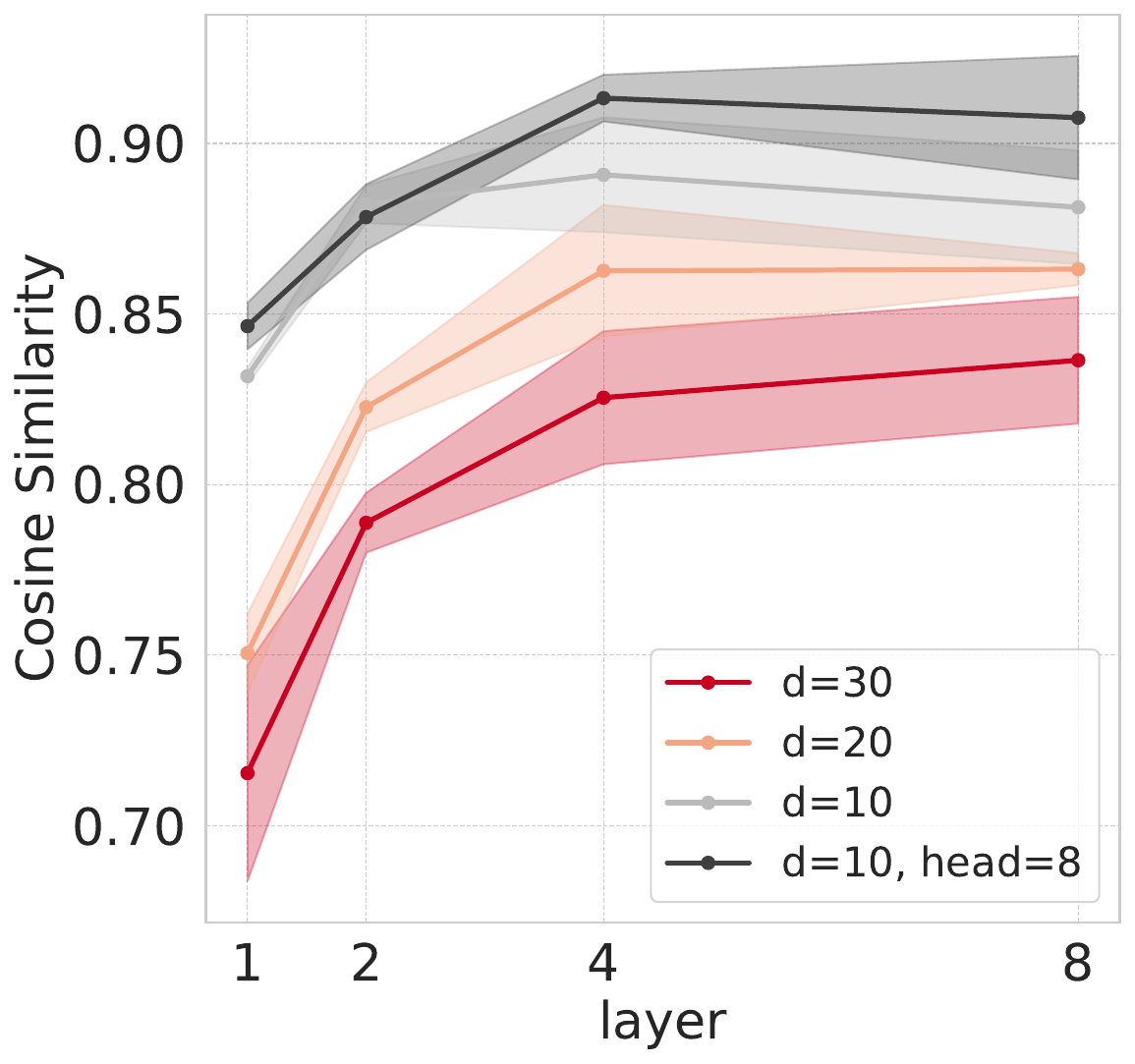}
    \endminipage
    \minipage{0.33\textwidth}
        \includegraphics[width=\linewidth]{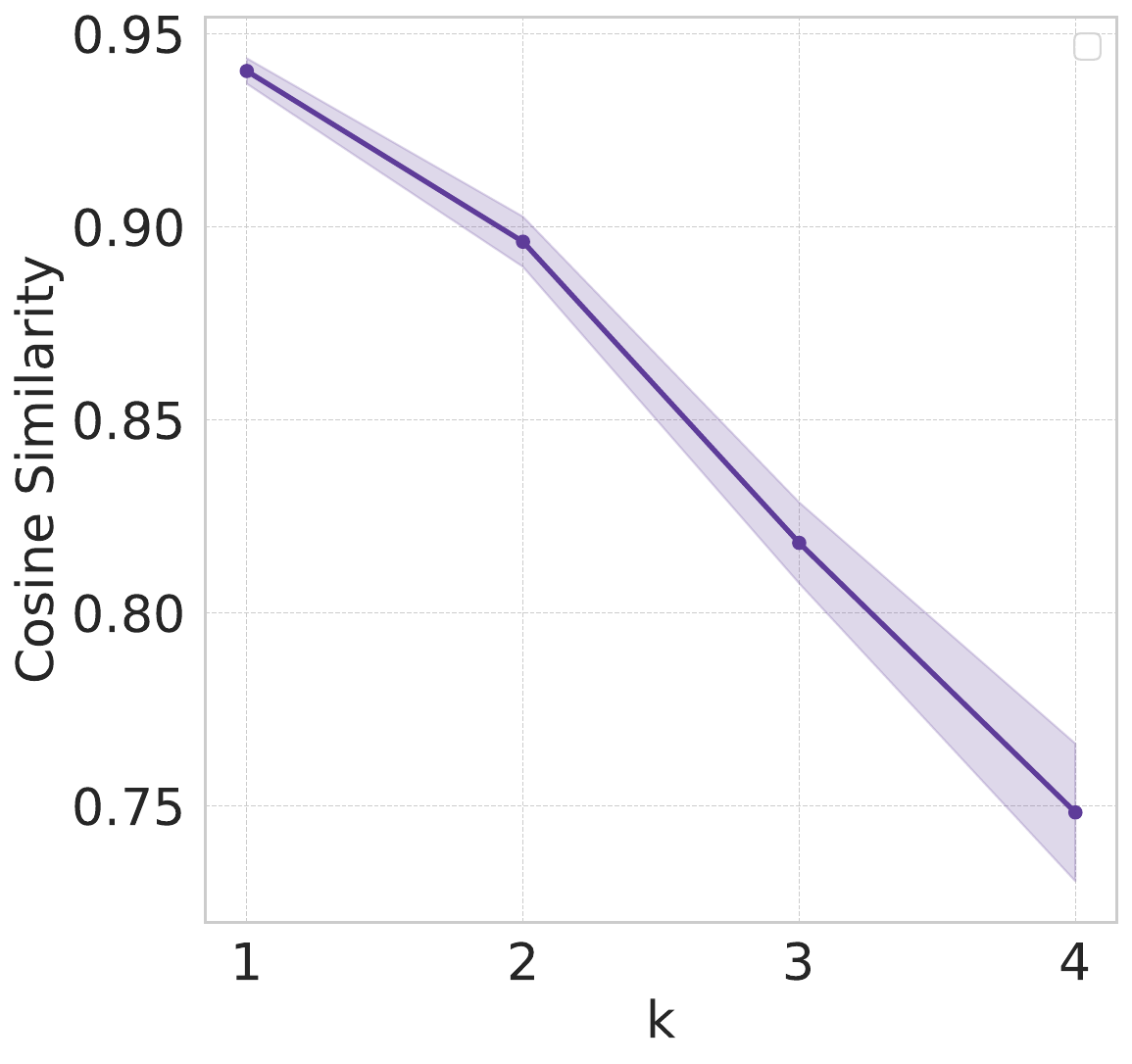}
    \endminipage\hfill
    \vspace{-1em}
    \caption{\textbf{Eigenvector Prediction on Synthetic Data.}
    \emph{(1) Left: Prediction of top-$1$ eigenvector with different input dimension $d$.} We use a small Transformer and $N=10$.
    \emph{(2) Middle: Prediction of top-$1$ eigenvector with varying number of layers.} 
    We start from a small Transformer and use $N=10$ and $d=10$. 
    \emph{(3) 
    Right: Predictions of eigenvectors with different numbers of $k$.} We use $N=10$ and $d=10$ in this experiment. The decrease in performance when the number of eigenvectors $k$ gets larger might be due to the number of layers being small.
    }
    \label{fig:encoder_eigenvector}
\end{figure}

\begin{figure}[t]
    \centering
    \minipage{0.45\textwidth}
        \includegraphics[width=\linewidth]{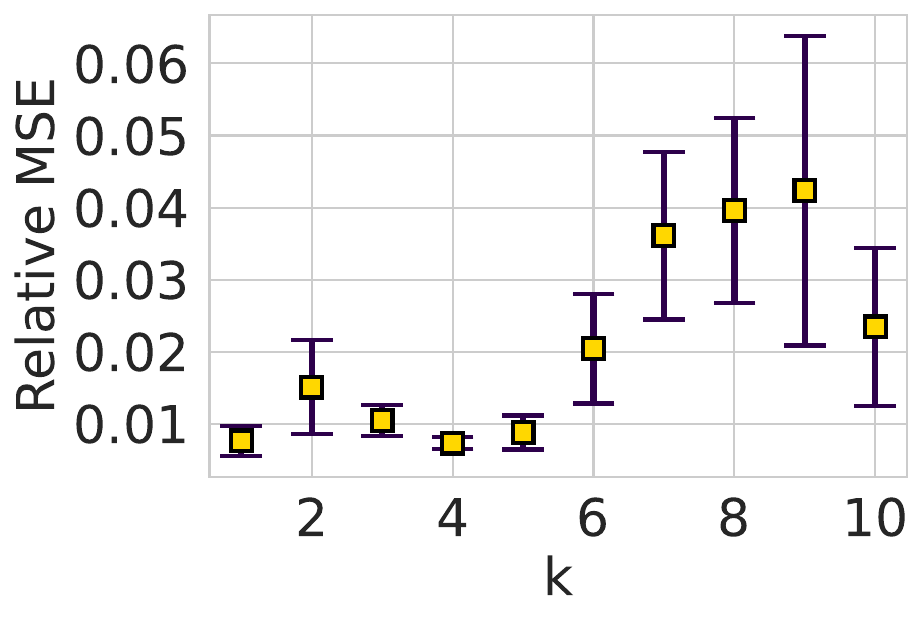}
    \endminipage\hfill
    \minipage{0.45\textwidth}
        \includegraphics[width=\linewidth]{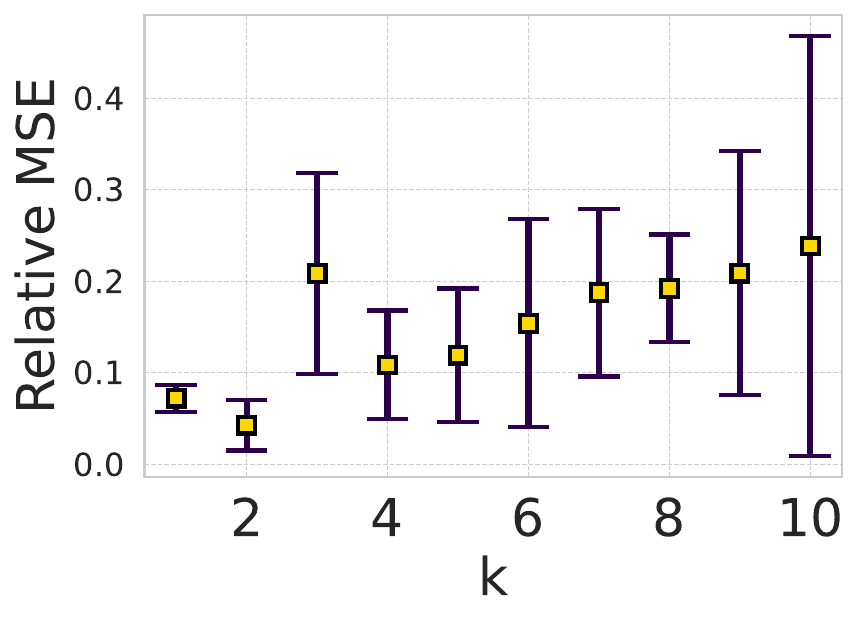}
    \endminipage\hfill
    \vspace{-1em}
    \caption{\textbf{Eigenvalues Prediction on Real World Data.} 
    \emph{(1) Left: Predicting Top-$10$ Eigenvalues on the MNIST Dataset.}  
    \emph{(2) Right: Predicting Top-$10$ Eigenvalues on the FMNIST Dataset.} Our results indicate that the pre-trained Transformer model can automatically extract principal components from the complicated, high-dimensional images.
    }
    \label{fig:encoder_eigenvalues-realworld}
\end{figure}

\begin{figure}[t]
    \centering
    \minipage{0.45\textwidth}
        \includegraphics[width=\linewidth]{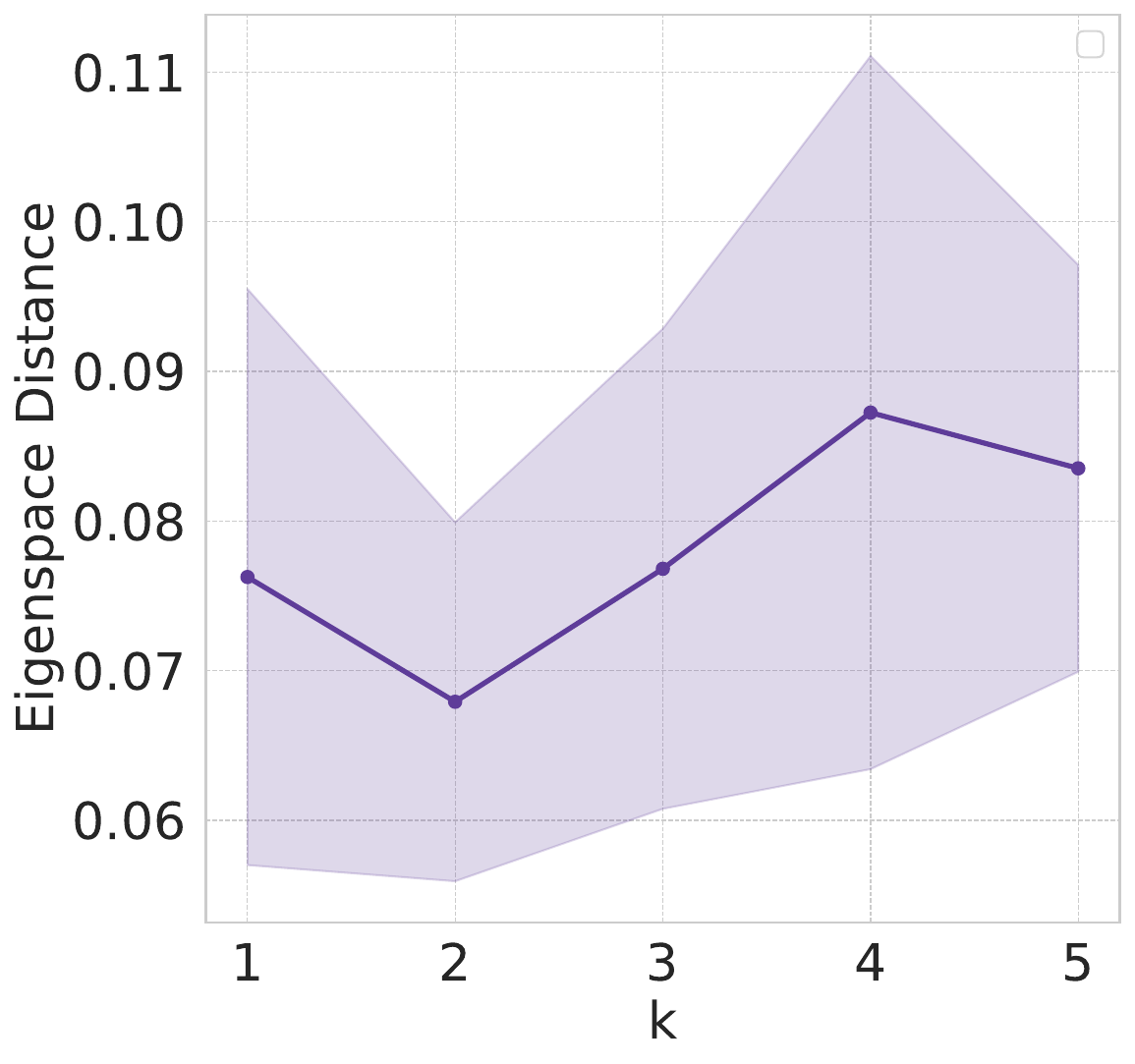}
    \endminipage\hfill
    \minipage{0.45\textwidth}
        \includegraphics[width=\linewidth]{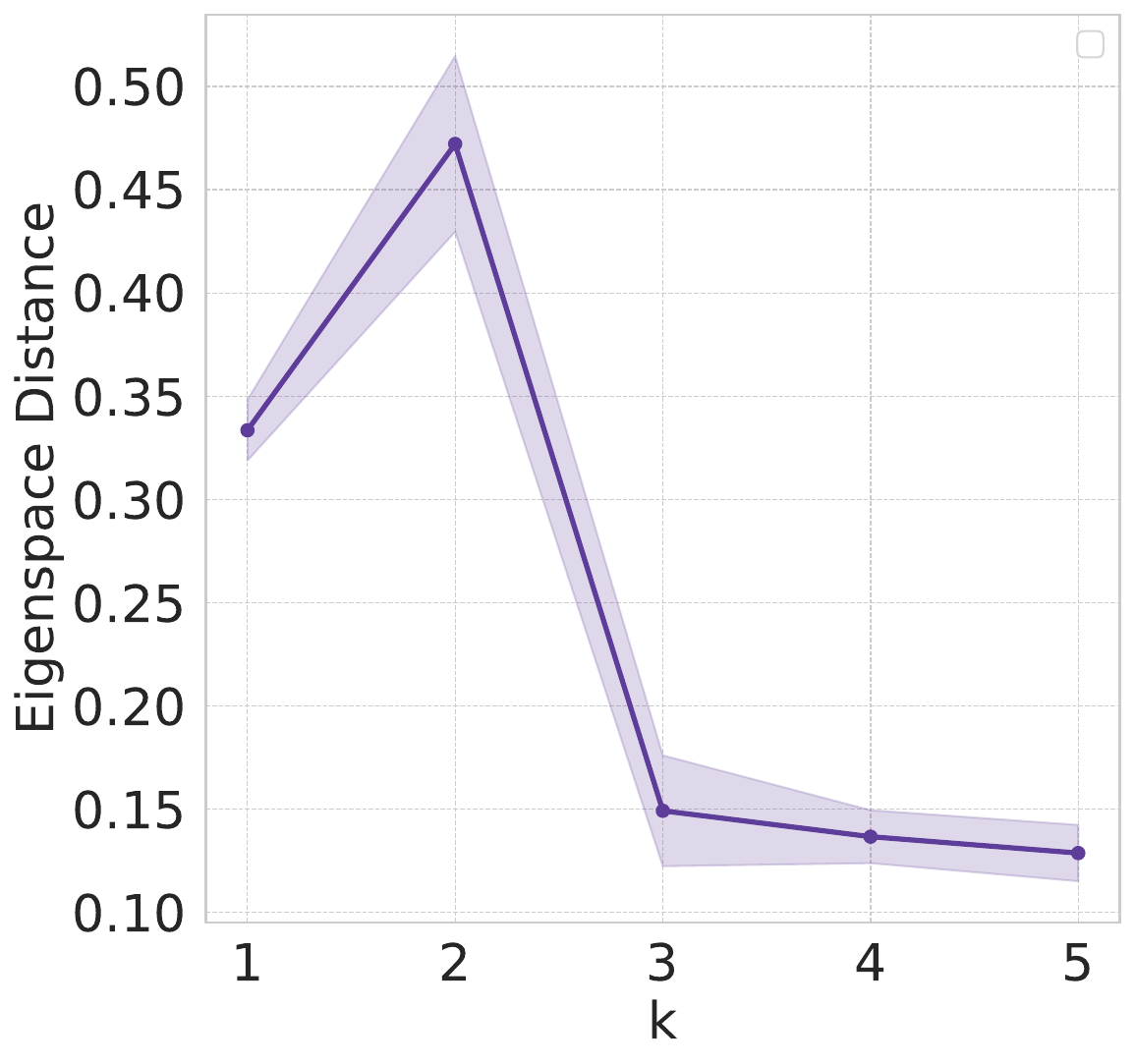}
    \endminipage\hfill
    \vspace{-1em}
    \caption{{\textbf{Principal Space Prediction.} 
    We train a large Transformer to predict multiple principal component, but replace cosine similarity loss with eigenspace distance loss.
    \emph{(1) Left: Eigenspace Prediction on the Synthetic Data.}  \emph{(2) Right: Eigenspace Prediction on the MNIST dataset.}
    Each point in the two figures is obtained by training the large Transformer for $100k$ steps and evaluated on $2048$ testing data, and average over 10 runs. We note that the performance degradation at $k=1,2$ in the MNIST dataset might be caused by the spectral gap for the first few principal components being too large.}}
    \label{fig:encoder_eigenvalues-realworld}
\end{figure}


\begin{table}[t]
\centering
\caption{{\textbf{Multiple Principal Components Prediction.}}
{We train a large Transformer ($L=12, Embed-Dim = 256, M = 8$) on synthetic dataset to predict top-$4$ principal components, and use cosine similarity as metric. Since the task is much harder, we train the model for $150k$ steps compared to $20k$ steps for other synthetic tasks in section \ref{sect401}.
For MNIST, we train the model for $500k$ steps.
}}
\tny{\begin{tabular}{lcccc}
\toprule
\textbf{k-th eigenvec.} & \textbf{k=1} & \textbf{k=2} & \textbf{k=3} & \textbf{k=4} \\ \hline
${\text{Transformer (Synthetic)}}$    & $0.9525 (0.0065)$ & $0.8648 (0.0092)$ & $0.7286 (0.0142)$ & $0.4471 (0.0722)$ \\
${\text{the Power Method (Synthetic)}}$ & $0.6807 (0.0022)$ & $0.4297 (0.0074)$ & $0.2954 (0.0019)$ & $0.2050 (0.0015)$ \\
${\text{Transformer (MNIST)}}$   & $0.9044 (0.0057)$ & $0.6425 (0.0091)$ & 0.4851 (0.0111) & 0.2904 (0.0117) \\
\bottomrule
\end{tabular}}
\label{tab:mul_k_vector_encoder}
\end{table}

This section provides simulation details and results on synthetic and real-world datasets. Our simulations are designed for three problems: (1) Prediction of eigenvalues and principal components are given by section \ref{sect401}; (2) Clustering Mixture of Gaussians is given by section \ref{sect42}.
\subsection{The Prediction of Pincipal Components/Eigenvalues}\label{sect401}
In this section, we first provide a brief overview of the experimental setup. Then, we provide details on the data preparations for both the synthetic and real-world datasets, the evaluation metrics, and discussions. The figures for this experiments are given by \ref{fig:encoder_eigenvalues}, \ref{fig:encoder_eigenvector}, and \ref{fig:encoder_eigenvalues-realworld}.
\paragraph{Experimental Setup.}\label{sect411}
Our experiments for the prediction of principal components and eigenvalues goes by first constructing the training set. 
Using this set, we train a Transformer model by modifying the GPT-2 architecture \citep{radford2019language}into an encoder-based model with a ReLU activation function. Then we evaluate the model on the evaluation dataset each with 1,280 data points. Each experiment repeats this procedure with three different random seeds. We record the average performance of Transformers in the prediction tasks of both top-k eigenvalues and principal eigenvectors.  

All experiments run on RTX 2080 Ti GPUs.
We construct a training pipeline using PyTorch library \citep{paszke2019pytorch} and generate data using the Scikit-learn library \citep{pedregosa2011scikit}.
Training with 20k steps takes approximately 0.5 hours for a small Transformer.
Most tasks use 20k training steps with learning rates $0.001$ with the exception being the multiple top-$k$ eigenvector prediction task where we train the model for 150k steps with a learning rate of $0.0005$.

\paragraph{Metrics.}
For eigenvalues, we use relative mean squared error (RMSE) as loss function ${L}_{\text{RMSE}}$ and evaluation metric.
For the loss of predicting top-$k$ eigenvalue, the loss function is defined as follows
\begin{align*}
    {L}_{\text{RMSE}}( \Lambda, \hat{\Lambda} )
    \coloneqq
    \frac{1}{k} \sum_{i=1}^{k} \frac{\lambda_{i} - \hat{\lambda}_{i}}{\lambda_{i} + \epsilon},\quad\Lambda =\diag(\lambda_i,i\in[k]),\quad\hat\Lambda=\diag(\hat\lambda_i,i\in[k]).
\end{align*}
The $\epsilon$ in the denominator ensures numerical stability by preventing division by zero.
For eigenvectors, we use cosine similarity as the loss function and evaluation metric.
{For predicting top-$k$ eigenvectors, denote $\bfa V=\begin{bmatrix}
    \bfa v_1&\ldots&\bfa v_k
\end{bmatrix} \in \R^{d \times k}$ be the matrix of ground truth eigenvectors and $\hat{\bfa V}$ the predicted eigenvector matrix.}
When we are predicting $k$ eigenvectors, the loss function is defined as
\begin{align*}
    {L}_{\text{cos}}({\bfa V, \hat{\bfa V}})
    \coloneqq
    \frac{1}{k} \sum_{i=1}^{k} 1 - \frac{\bfa v_{i} \cdot \wha{v}_i}{\max (\Vert\bfa  v_{i}\Vert_2 \cdot \Vert \wha{v}_{i}\Vert_2, \epsilon)},
\end{align*}
where $\bfa v_{i}$ represent the $i$-th eigenvector. It is not hard to check the relationship between these loss functions with the classical $L_2$ losses. 
{In addition to cosine similarity, we also use the eigenspace distance as another training objective. The eigenspace distance is defined as the Frobenius norm of the difference between the two projection matrices:
\begin{align*}
L_{\text{eig}}(\bfa V, \hat{\bfa V}) = \frac{1}{2} \Vert \bfa V \bfa V^\top - \hat{\bfa V} \hat{\bfa V}^\top \Vert_{\text{F}}^2,
\end{align*}
which measures how well the subspaces spanned by the eigenvectors are preserved in the prediction.}

\paragraph{Preparation for the Synthetic Data.}
For synthetic data $\bfa X \in \R^{{d} \times N}$, we generate each column with a randomly initialized multivariate Gaussian distribution $\ca N(\mu, \Sigma)$ where $\Sigma$ is a random matrix.
Specifically, for each $\bfa X_i \in \R^d$, we sample $\bfa Z_i \sim N(0, I) \in \R^d$.
We form $\bfa Z = [\bfa Z_1, \cdots, \bfa Z_N]$ and transform it using matrix $\bfa L \sim N(0, I_{d^2}) \in \R ^{d \times d}$, yielding the desired training sample $\bfa X$.
We then generate the labels as the top-$k$ eigenvalues $\lambda$ and eigenvectors $\bfa V$ of the empirical covariance matrix $\bfa X^\top \bfa X/N$ via \texttt{numpy.linalg.eigh}.

\paragraph{Data Preparation for Real World Data.}
We choose MNIST \citep{lecun1998mnist} and Fashion-MNIST (FMNIST) \citep{xiao2017fashion} as our real-world datasets. The training set is constructed by applying the SVD on the samples and using them as the labels for the input. And in the testing phase, we input unseen images and evaluate the result according to the different metrics for the eigenvalues and principal components.

\subsection{Clustering a Mixture of Gaussians}\label{sect42}
\begin{figure}[htbp]
    \centering
    \minipage{0.33\textwidth}
        \includegraphics[width=\linewidth]{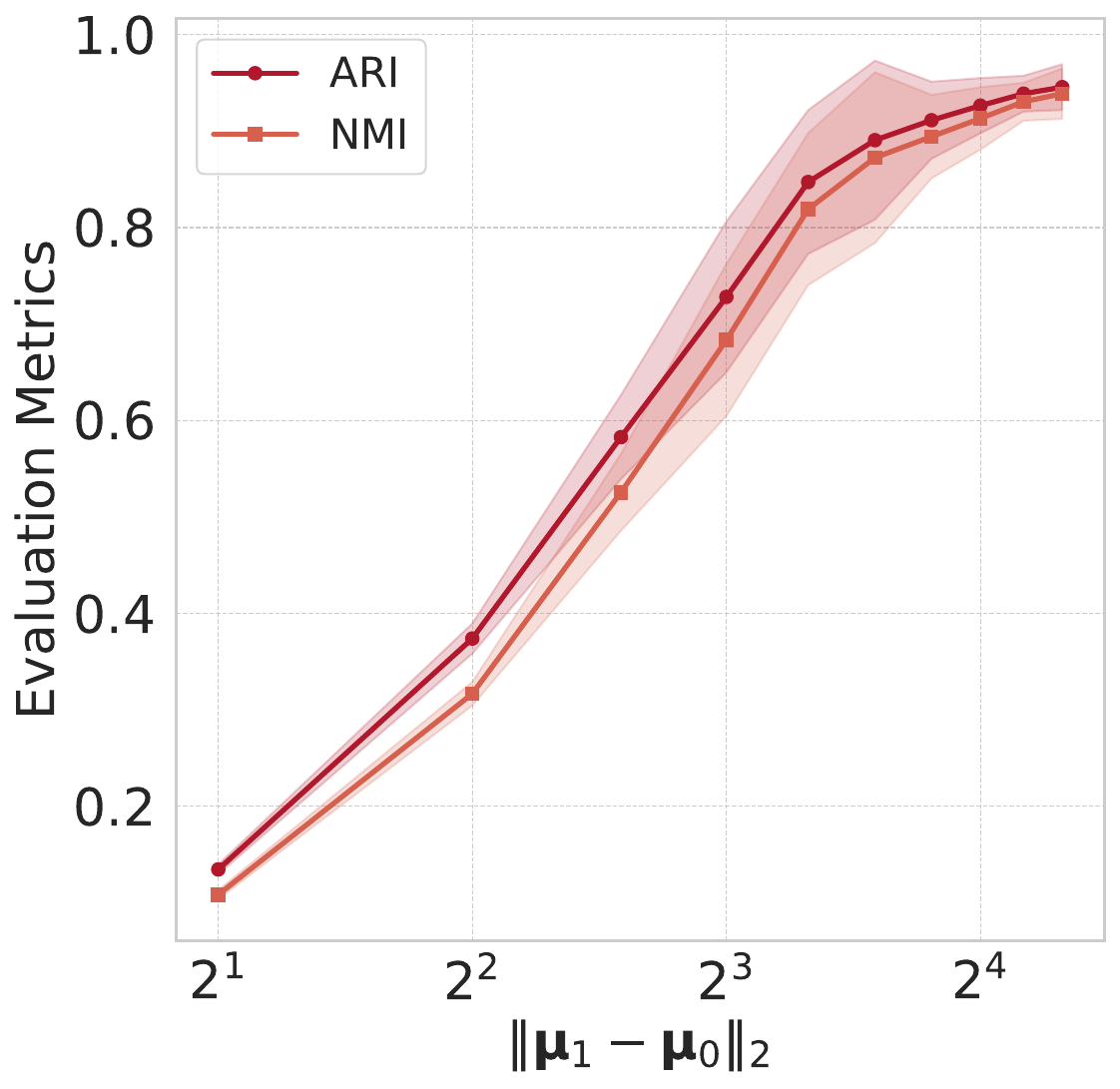}
    \endminipage\hfill
    \minipage{0.33\textwidth}
        \includegraphics[width=\linewidth]{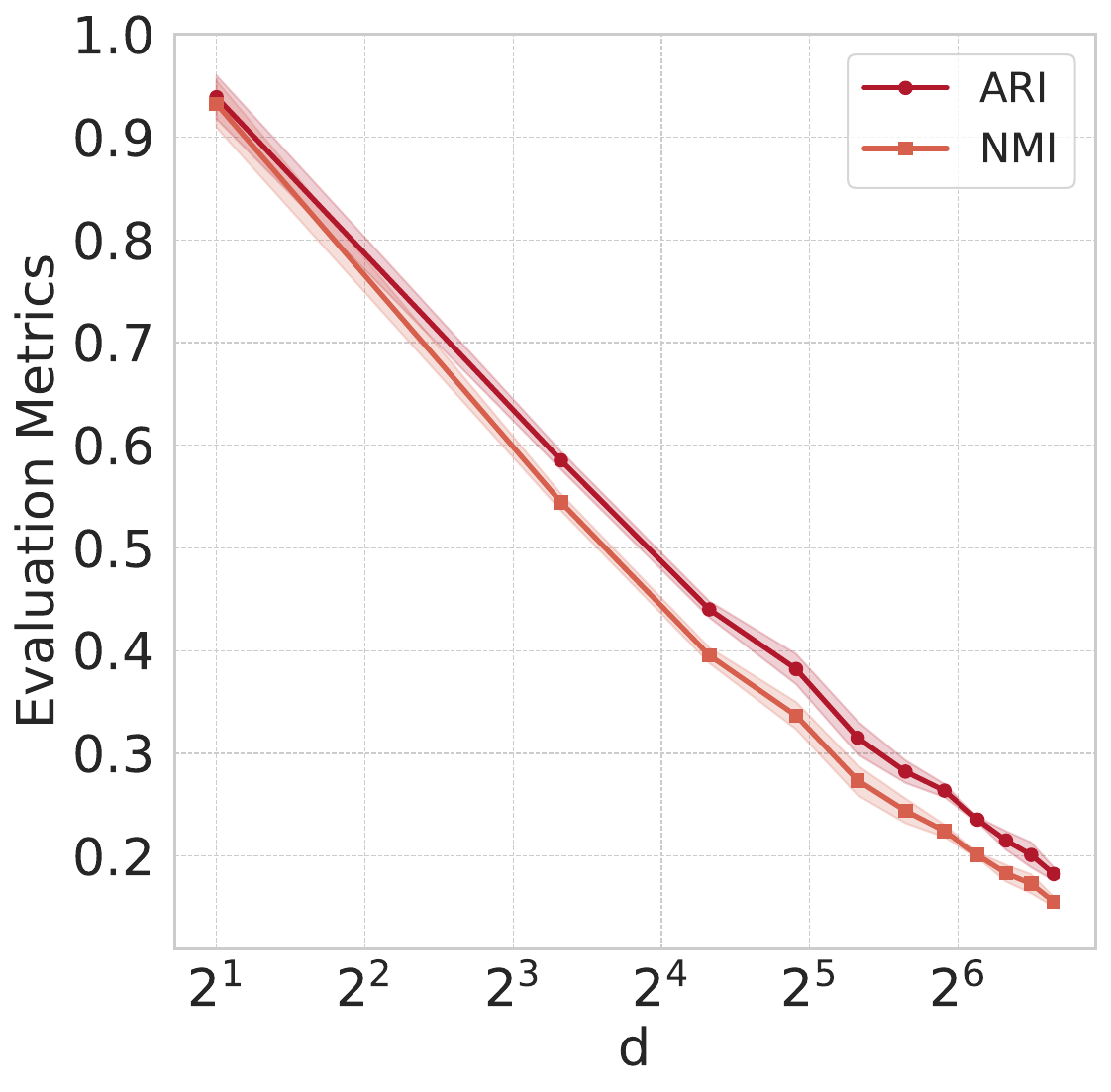}
    \endminipage\hfill
    \minipage{0.33\textwidth}
        \includegraphics[width=\linewidth]{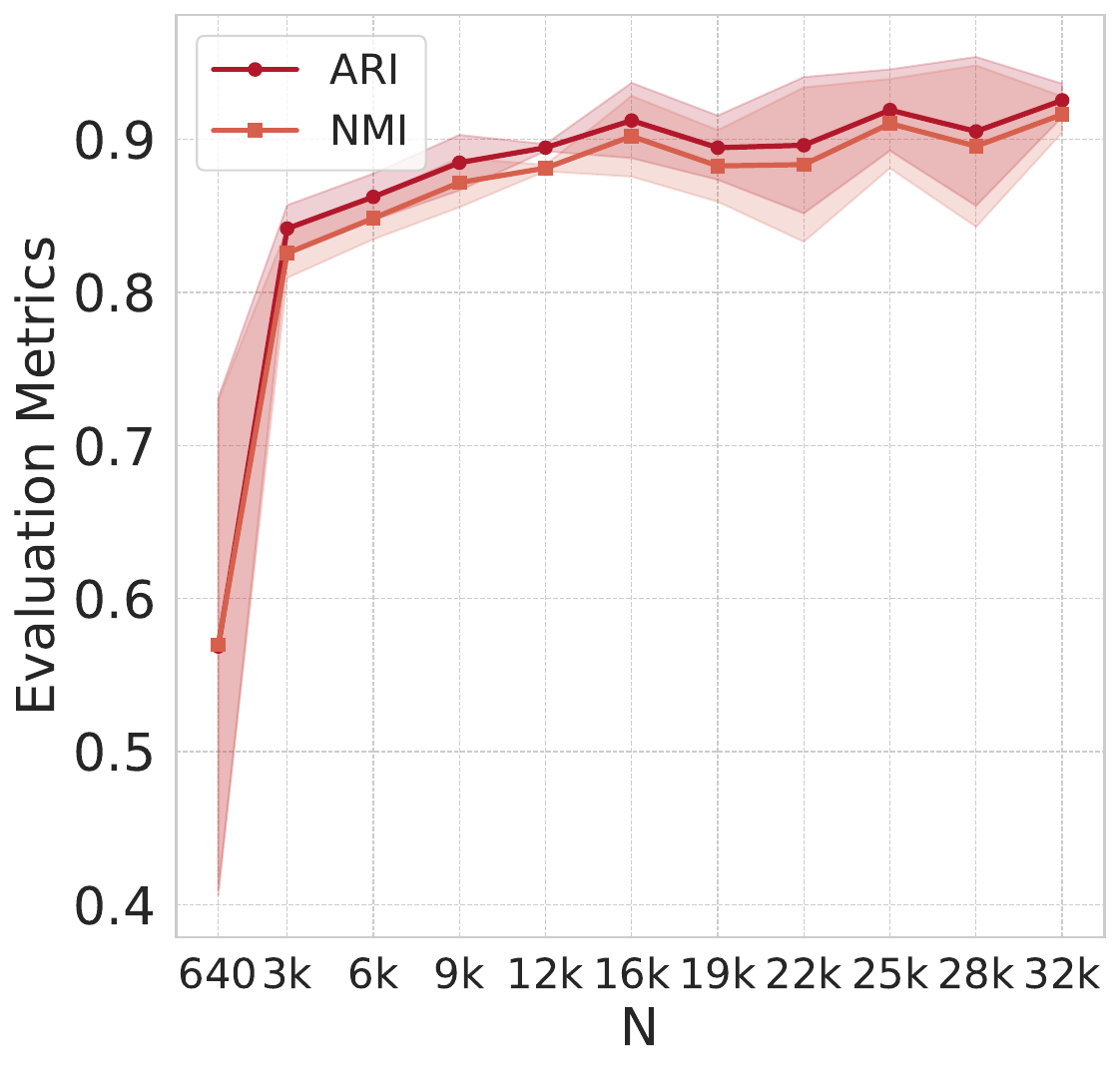}
    \endminipage\vspace{1em}
    \minipage{0.33\textwidth}
        \includegraphics[width=\linewidth]{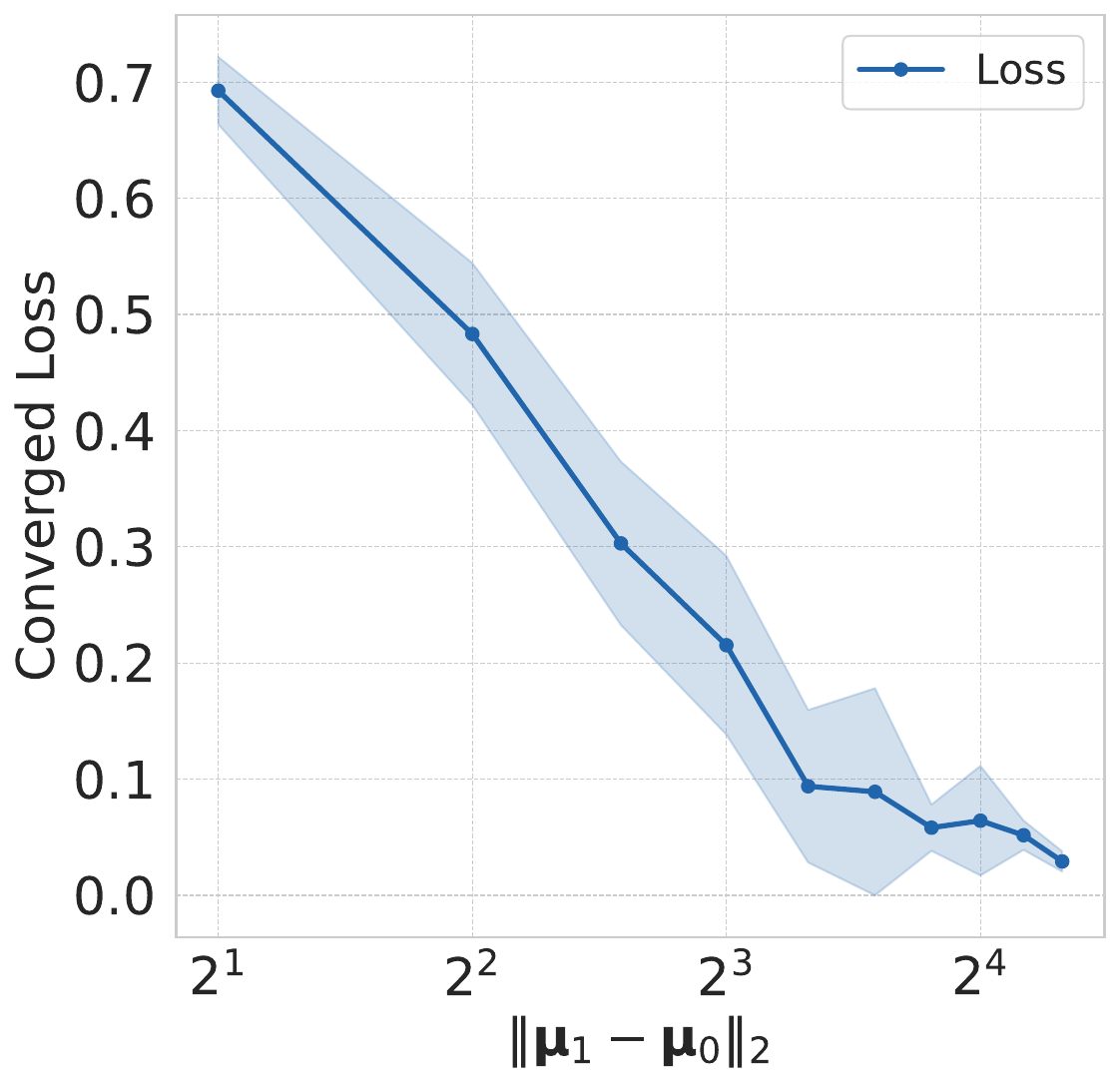}
    \endminipage\hfill
    \minipage{0.33\textwidth}
        \includegraphics[width=\linewidth]{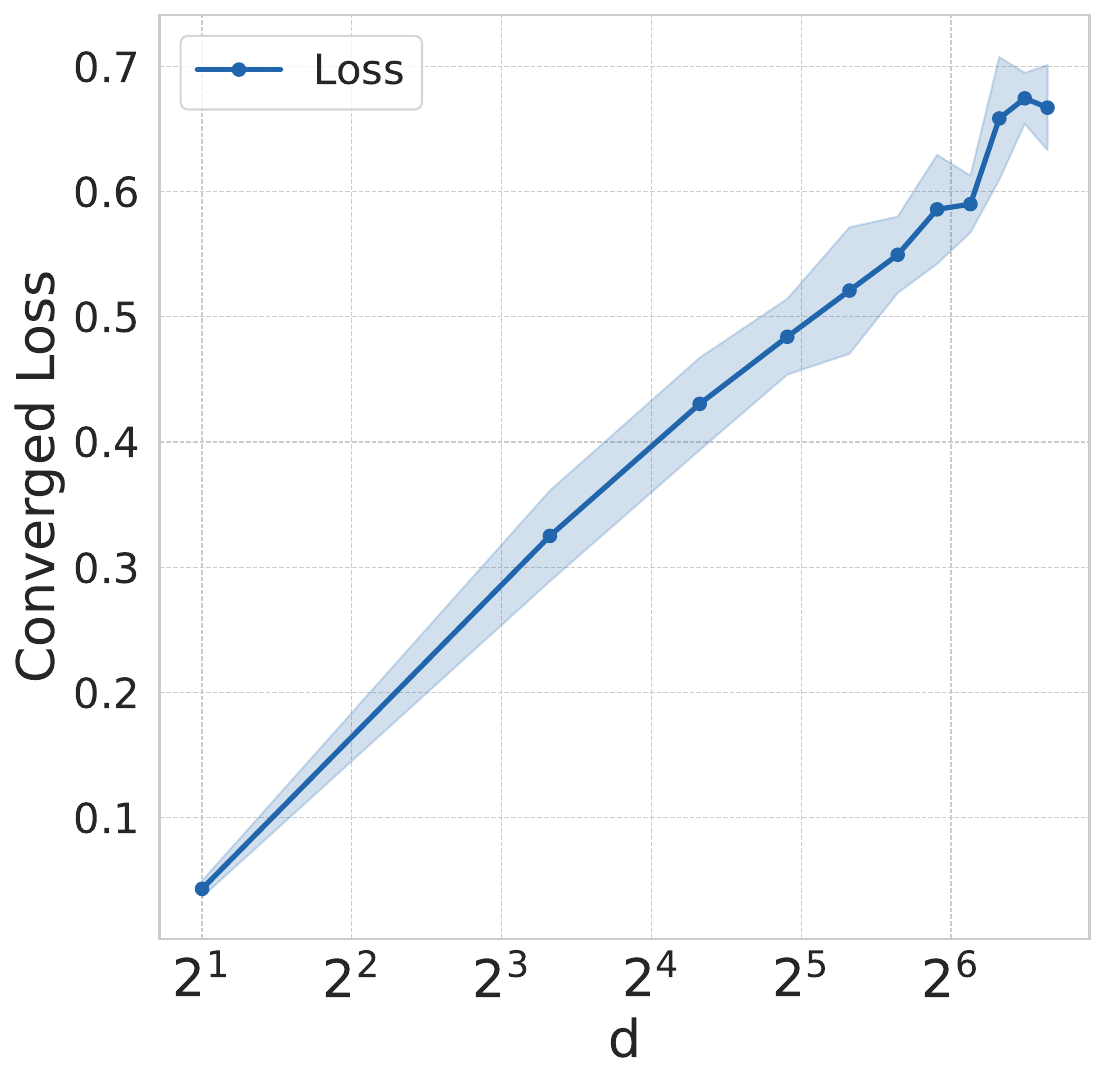}
    \endminipage\hfill
    \minipage{0.33\textwidth}
        \includegraphics[width=\linewidth]{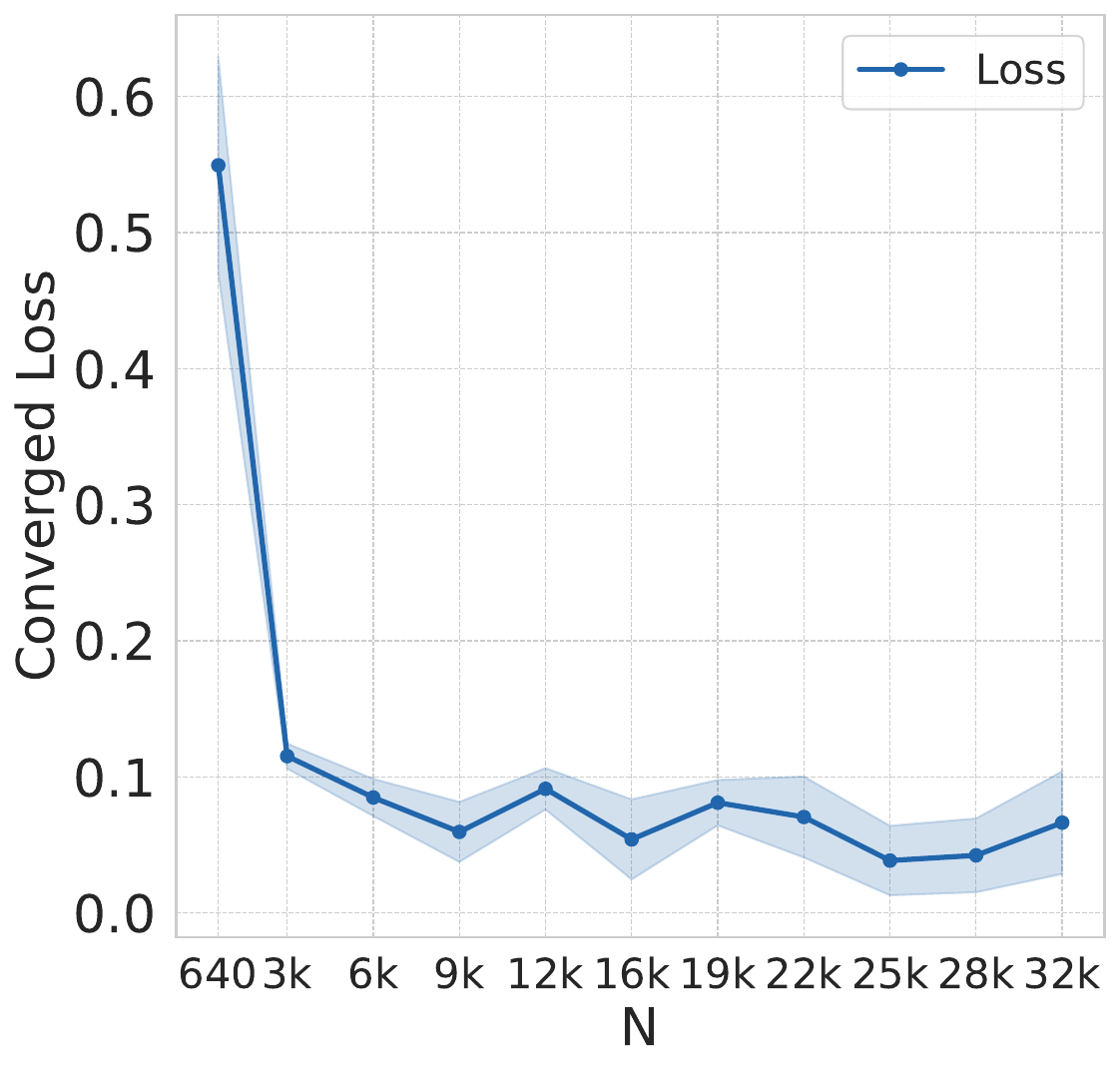}
    \endminipage
    \caption{{\textbf{2-Class GMM Clustering on Synthetic Dataset.}
    For each input $X$ we set $N=200$ with balanced counts for each cluster. We consider three factors and their effects on the testing errors.
    The upper row displays the clustering results evaluated on ARI and NMI metrics, and the bottom row displays the loss on the testing set at convergence. 
    }}
    \label{fig:gmm_syn}
\end{figure}

\begin{figure}[htbp]
    \centering
    \minipage{0.45\textwidth}
        \includegraphics[width=\linewidth]{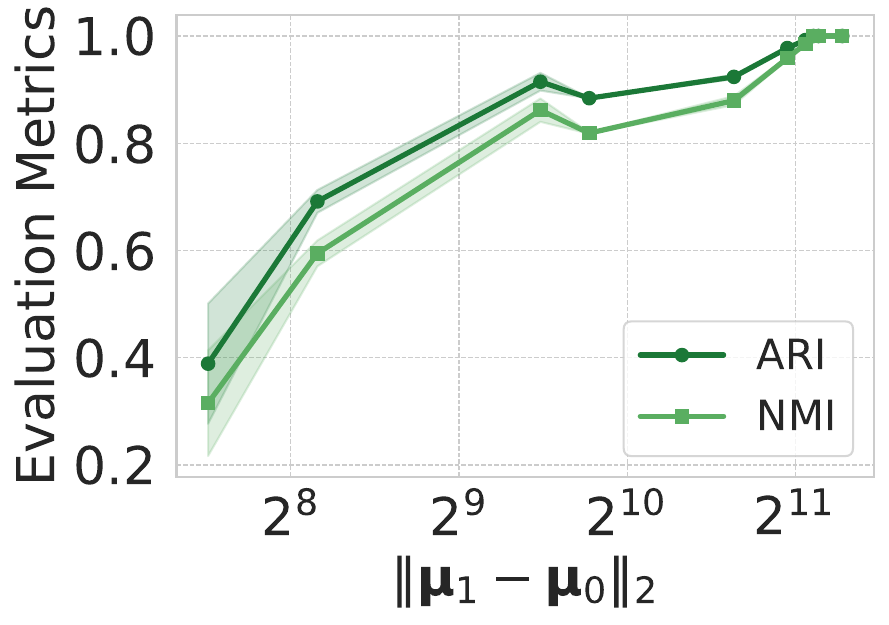}
    \endminipage\hfill
    \minipage{0.45\textwidth}
        \includegraphics[width=\linewidth]{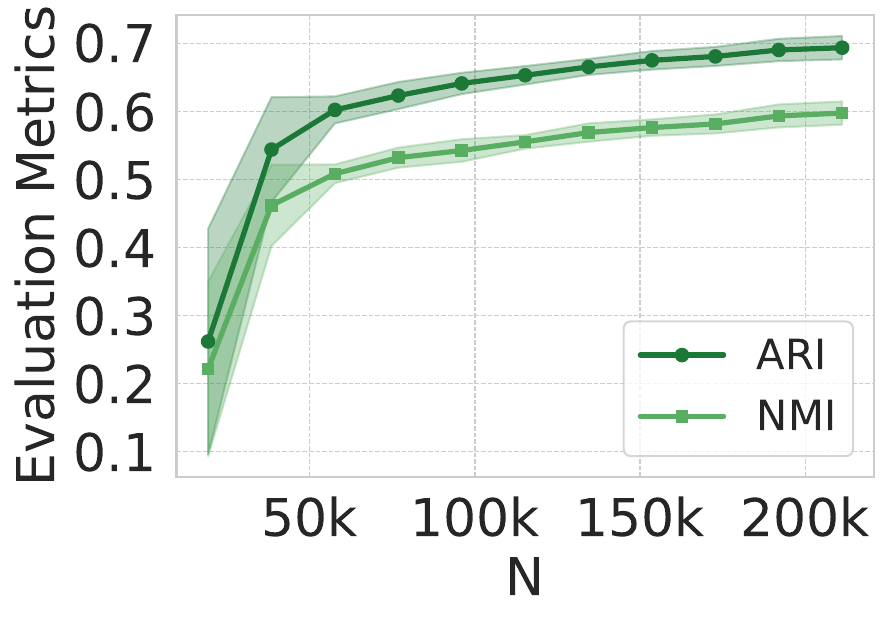}
    \endminipage\vspace{-0.1em}
    \minipage{0.45\textwidth}
        \includegraphics[width=\linewidth]{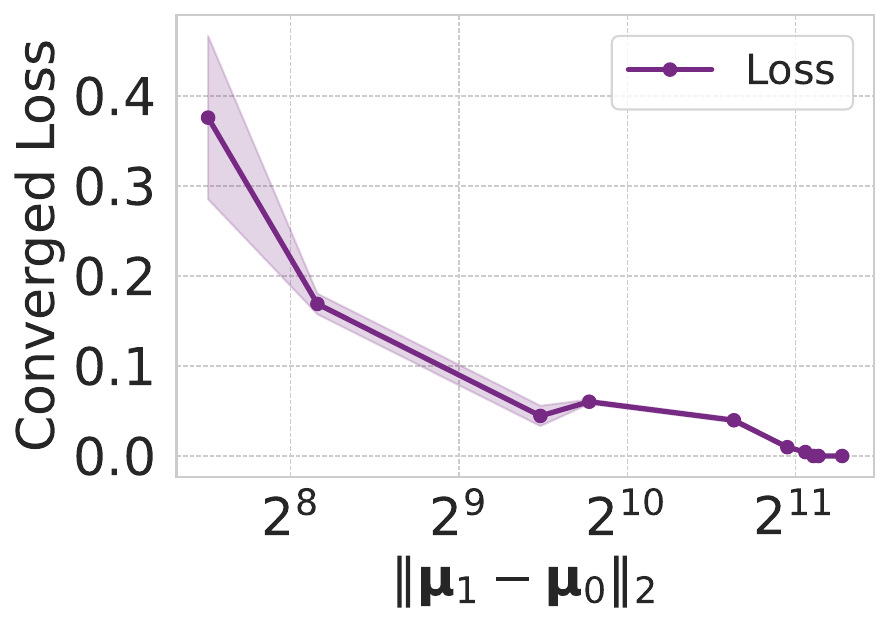}
    \endminipage\hfill
    \minipage{0.45\textwidth}
        \includegraphics[width=\linewidth]{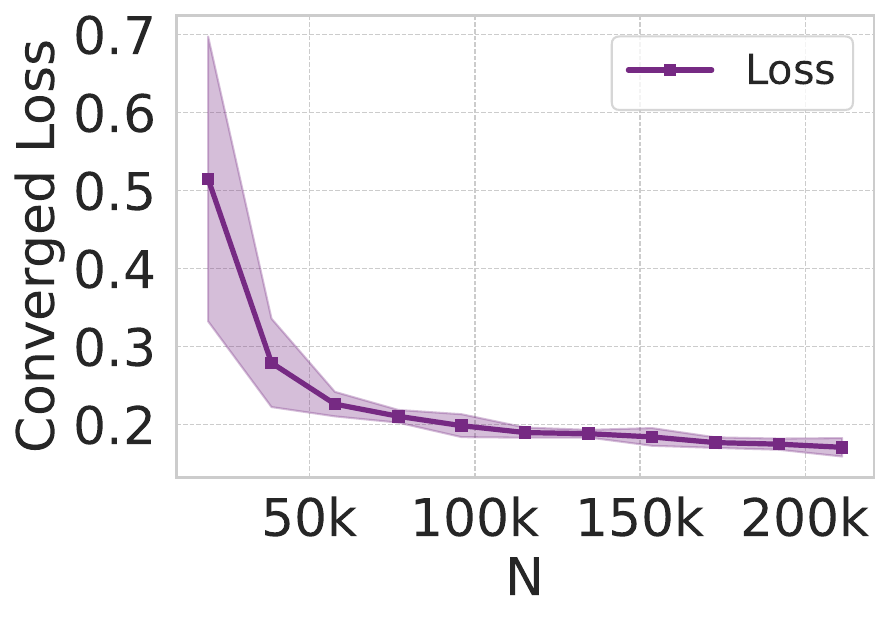}
    \endminipage
    \vspace{-1em}
    \caption{\textbf{2-Class GMM Clustering on Real World Dataset (Covertype)}.
    We test the Transformer's ability to perform 2-class clustering on the Covertype \citep{blackard1999comparative} dataset with $581,012$ observations, $d=54$, and $7$ different class.
    \emph{(1) Left Column: Distance between Two Cluster $\Vert\bfa\mu_1-\bfa\mu_0\Vert_2$;}
    \emph{(2) Right Column: Number of Training Samples $N$.}
    The upper row displays the clustering results evaluated on ARI and NMI metrics, and the bottom row displays the converged loss.
    As the separation or number of samples increases, the clustering performance improves (higher ARI/NMI) and the converged loss decreases. 
    }
    \label{fig:gmm_real}
\end{figure}

We present here the simulation results for the problem of Clustering Mixture of Gaussians. Our results are given by plots \ref{fig:gmm_syn} and \ref{fig:gmm_real}. The rest of the section is organized similarly to \ref{sect401}.

\paragraph{Experimental Overview.}
We verify the theoretical results by exploring the relationship between three factors versus the clustering performance: distance between two clusters $\Vert\bfa\mu_1-\bfa\mu_0\Vert_2$, data dimension $d$, and the number of training samples $N$. We use the same model architecture as in the previous section with layer $L=3$, head $M=2$, and embedding dimension $64$. We train the model with {3} different random seeds and report both the mean and standard deviation. The experiments in this section run on NVIDIA A100 80G GPUs. We train the model for $300$ iterations on synthetic data and train $3500$ iterations on the real-world data using stochastic gradient descent with the Pytorch package. 
Every point in the figure is evaluated on $512$ testing data points.
{\paragraph{Metrics.}}
 The metric that we evaluate the performance of our estimated cluster assignment $\wh z$ is given 
\begin{align*}
{L}^k_{GMM}(\wh z, z) \coloneqq \min_{\pi \in S_k} \frac{1}{N} \sum_{i=1}^N \lef\Vert \pi(z)-\wh z\rig\Vert_1,
\end{align*}
where $S_k$ denotes the set of all $k!$ permutations. For a special case, the loss function of two-class clustering is written out as equation \ref{gmmloss}. Additionally, we use two commonly used permutation-invariant metrics to evaluate the clustering problem: Adjusted Rand Index (ARI) and Normalized Mutual Information (NMI) \citep{ma2019learning, huang2020partially, monnier2020deep,sun2024lsenet,li2024image}.
\paragraph{Preparation for the Synthetic Data.}
{We generate our synthetic data as follows: For each input $\bfa X \in \R^{d \times N}$, we sample equal counts of data coming from the two clusters. We set $N=200$ in both synthetic and real-world data experiments. Each sample is generated according to isotropic Gaussian with covariances $\sigma^2 \bfa I$ where $\sigma^2 \sim Uniform[0.5,5.0]$. To generate mean vectors $\bfa\mu_0$ and $\bfa\mu_1$ in $\R^d$ with a fixed distance $\delta = \Vert\bfa\mu_1-\bfa\mu_0\Vert_2$, a random center point $\bfa c \in \R^d$ is first sampled. Then, a random vector $\bfa v$ is drawn from $N(0, 1)^d$ and normalized to a unit vector, $\bfa u = \bfa v / \Vert\bfa v\Vert$. The centroids are then given by $\bfa\mu_0 = \bfa c - \delta / 2 \cdot \bfa u$ and $\bfa\mu_1 = \bfa c + \delta / 2 \cdot \bfa u$.}
\paragraph{Preparation for the Real-World Data.}
For the real-world data evaluation, {we use the Covertype \citep{blackard1999comparative} dataset, which consists of $581,012$ observations with $d=54$ and $7$ different classes. The task is to predict the forest cover type with cartographic variables. For the experiment involving varying $\Vert\bfa\mu_1-\bfa\mu_0\Vert_2$, we select $10$ different combinations of two classes. For fixed $\Vert\bfa\mu_1-\bfa\mu_0\Vert_2$, each training sample $\bfa X \in \R^{d \times N}$ is formed by randomly sampling $N/2$ points from each of the selected classes.} 

\subsection{Additional Simulations}

\begin{figure}[!h]
    \centering
    \begin{minipage}{0.33
    \textwidth}
        \centering
        \includegraphics[width=\linewidth]{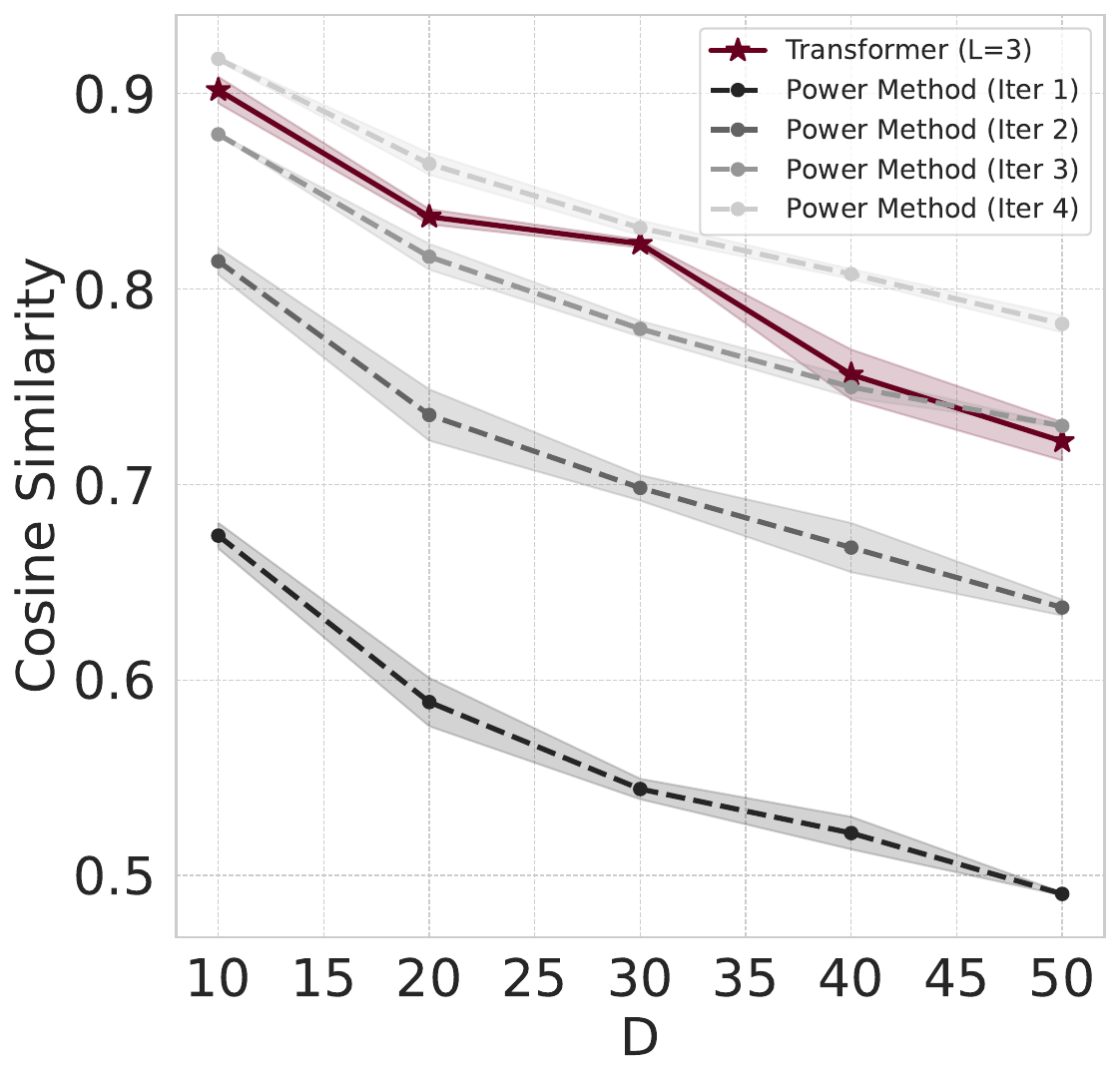}
    \end{minipage}\hfill 
    \begin{minipage}{0.33\textwidth}
        \centering
        \includegraphics[width=\linewidth]{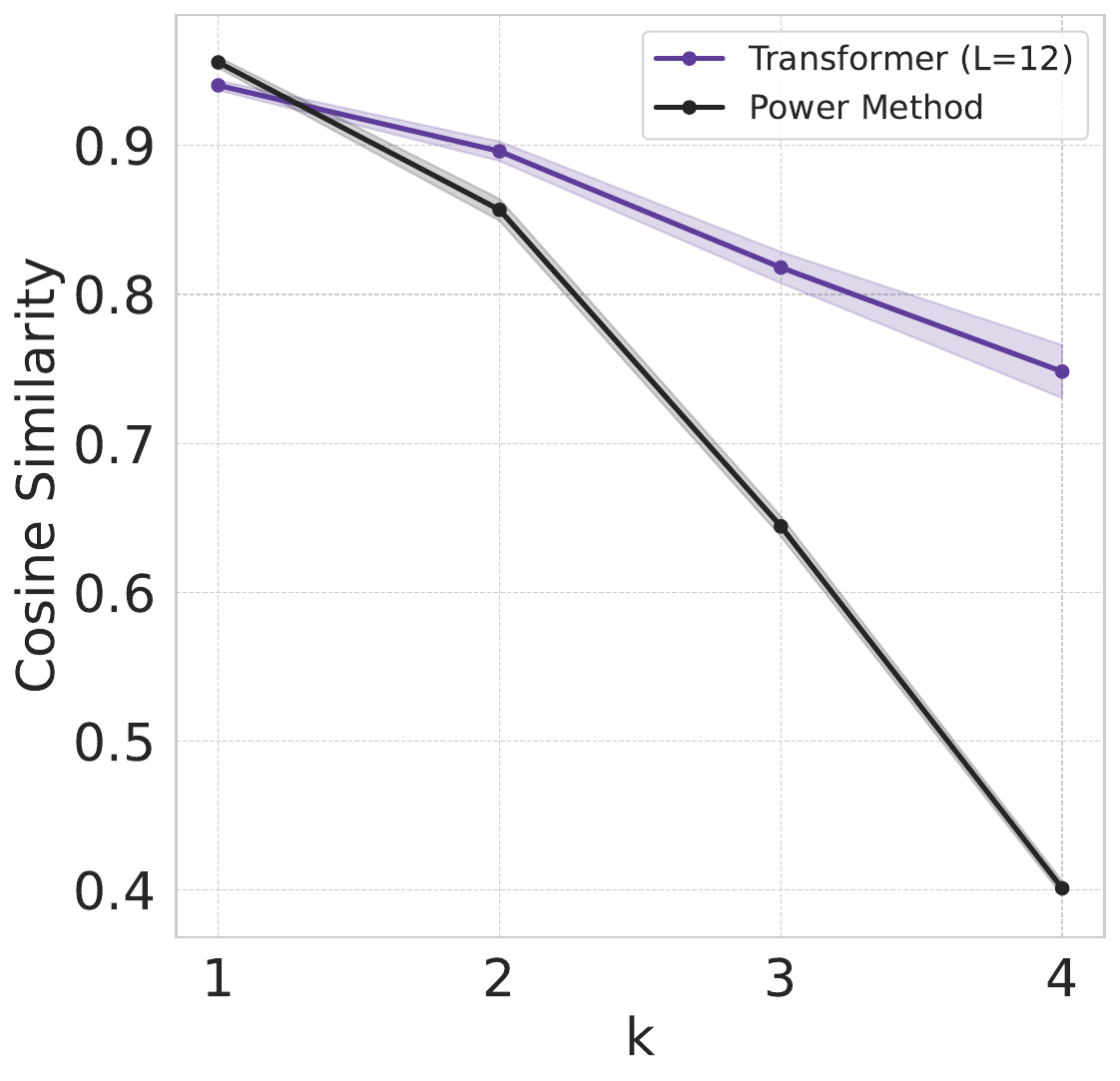}
    \end{minipage}\hfill
    \begin{minipage}{0.33\textwidth}
        \centering
        \includegraphics[width=\linewidth]{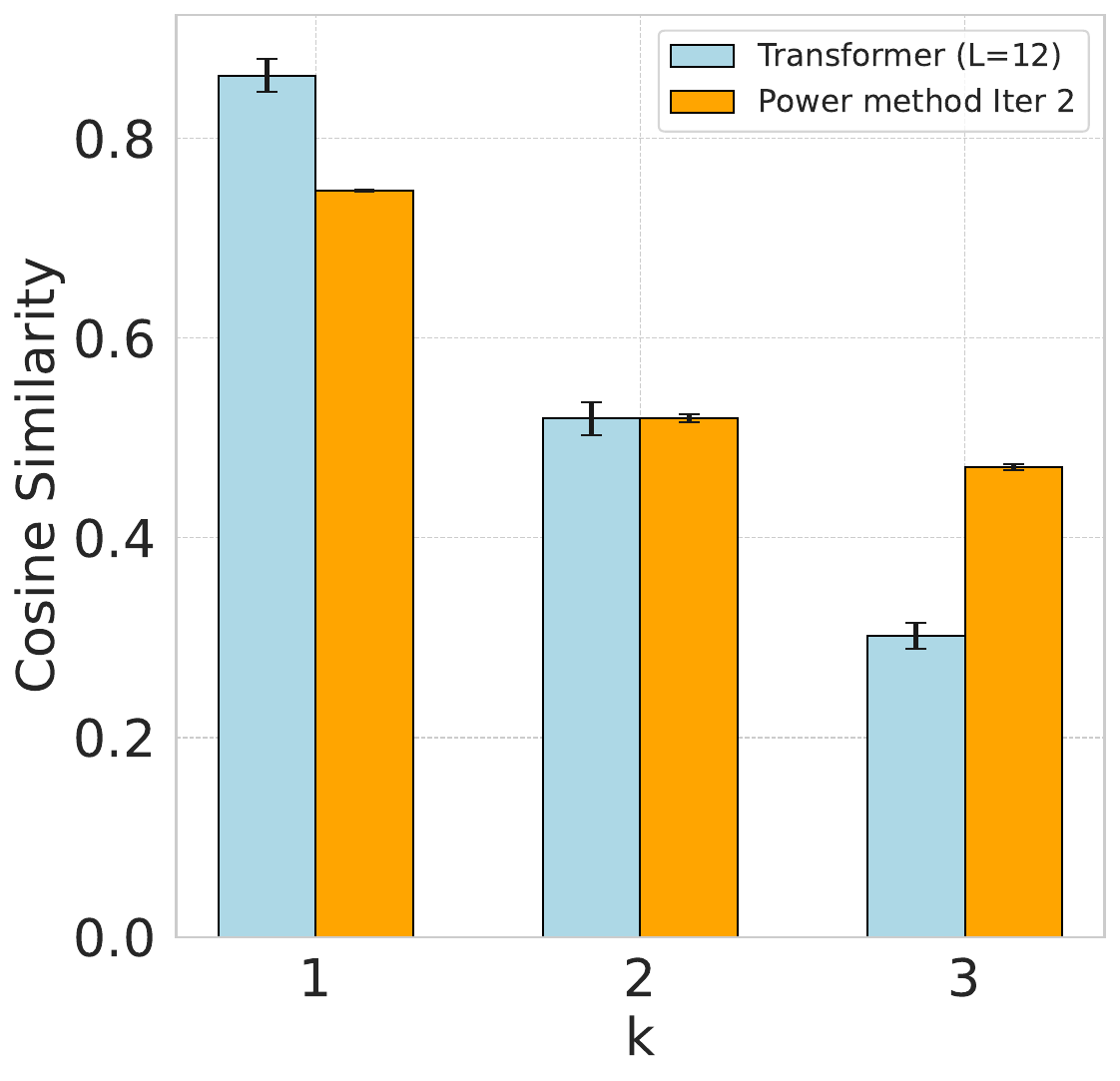}
    \end{minipage}
    \caption{{\textbf{Eigenvector Prediction Comparision of Transformer and the Power Method Baseline on the Synthetic and Real-World Datasets.}
    \emph{Left: Prediction of Top-$1$ Eigenvector with Different Input Dimensions. }
    We compare a $3$-layer small Transformer trained for $20k$ steps with the Power Method with different iterations $\tau$ on the top-$1$ principal component prediction tasks across different input dimensions.
    \emph{Middle: Synthetic Multiple Eigenvectors Prediction with different $k$.}
    This plot compares the Power Method baseline and the Transformer on the synthetic dataset.
    \emph{Right: MNIST Multiple Eigenvectors Prediction with different $k$.}
    This plot compares between the Power Method baseline and the Transformer on the MNIST dataset.
    }
    }
\label{fig:mnist_power_method_eigenvector}
\end{figure}

\begin{figure}[htbp]
    \centering
    \minipage{0.45\textwidth}
        \includegraphics[width=\linewidth]{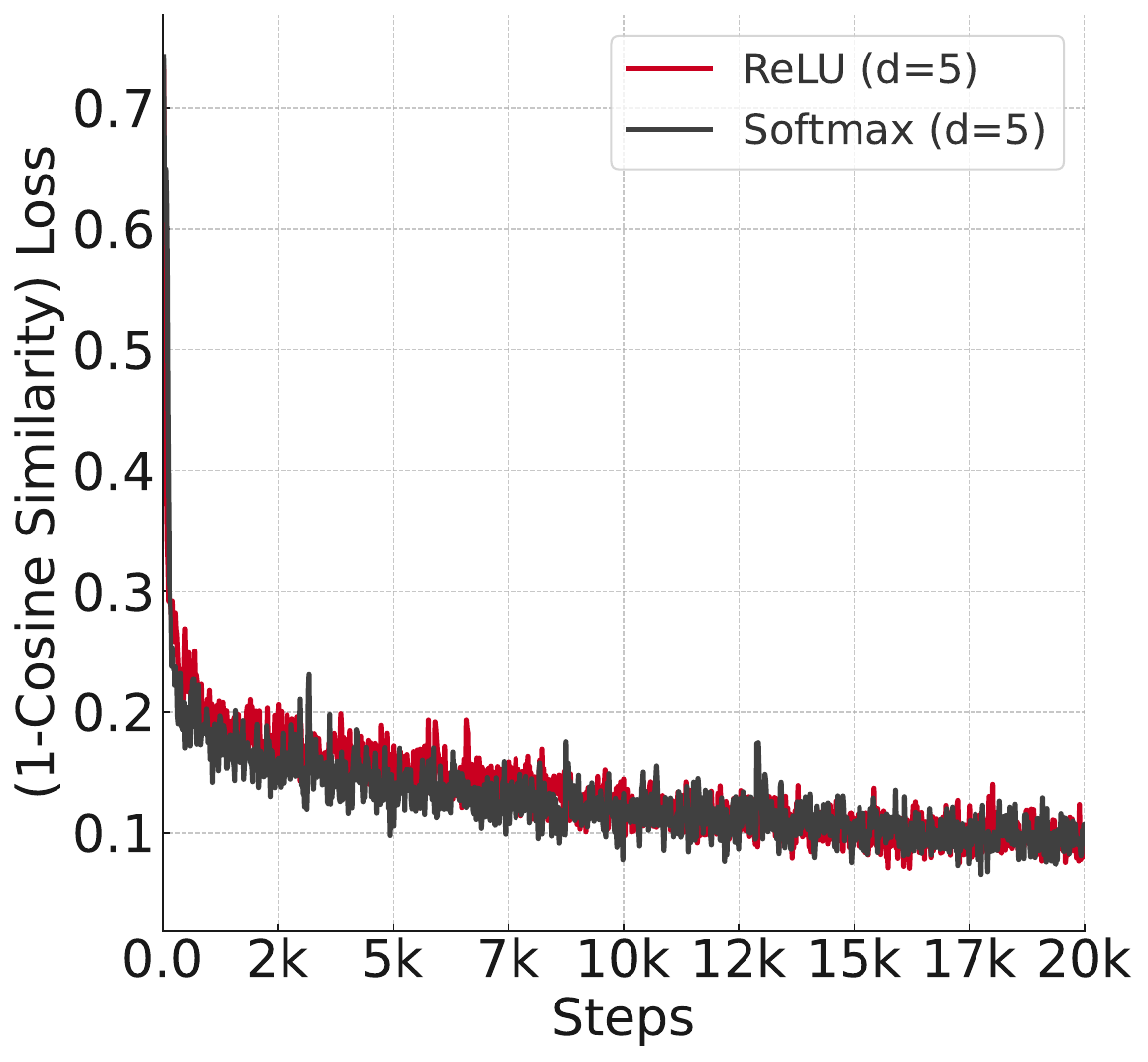}
    \endminipage
    \minipage{0.45\textwidth}
        \includegraphics[width=\linewidth]{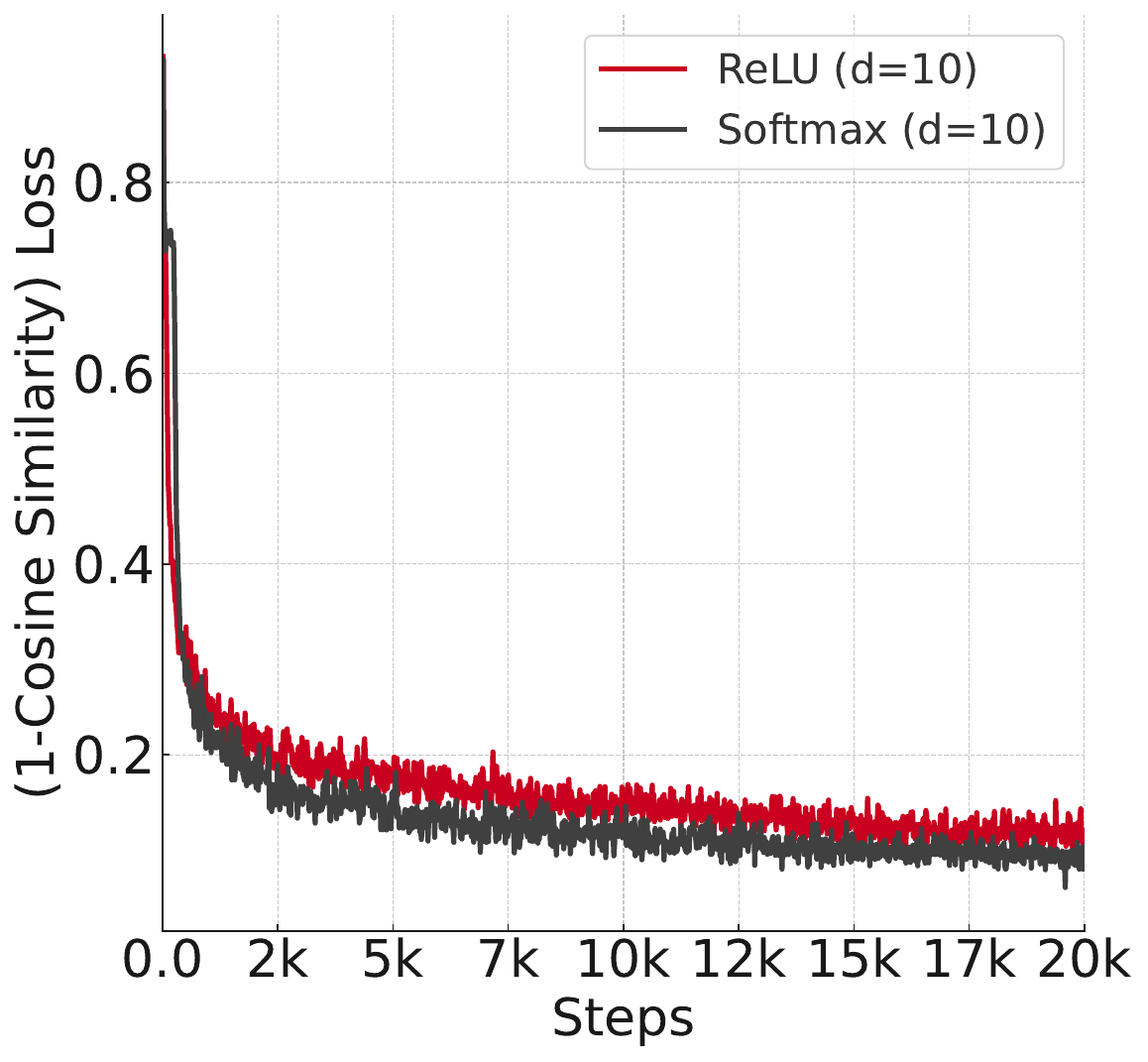}
    \endminipage\vspace{-0.1em}
    \minipage{0.45\textwidth}
        \includegraphics[width=\linewidth]{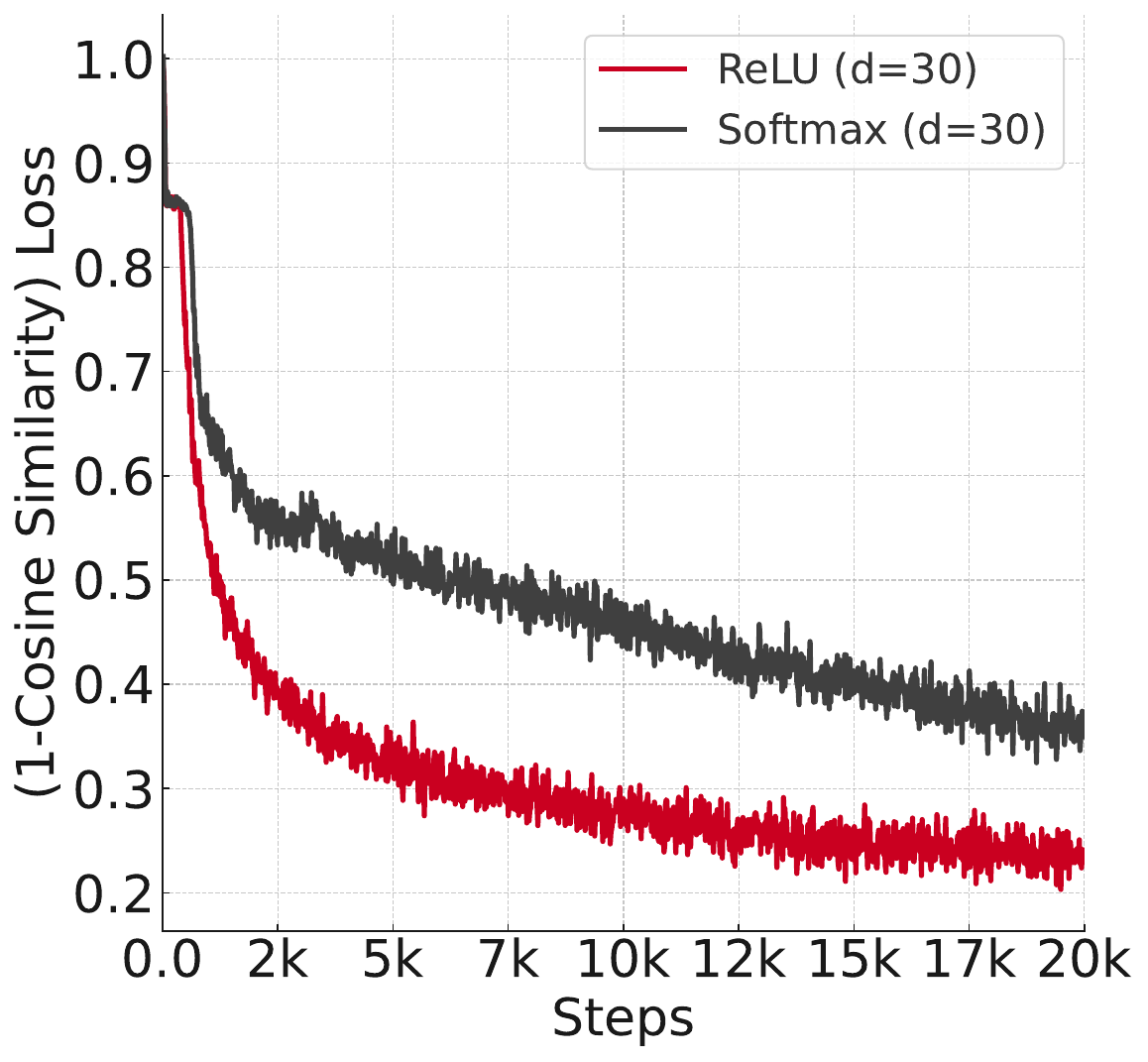}
    \endminipage
    \minipage{0.45\textwidth}
        \includegraphics[width=\linewidth]{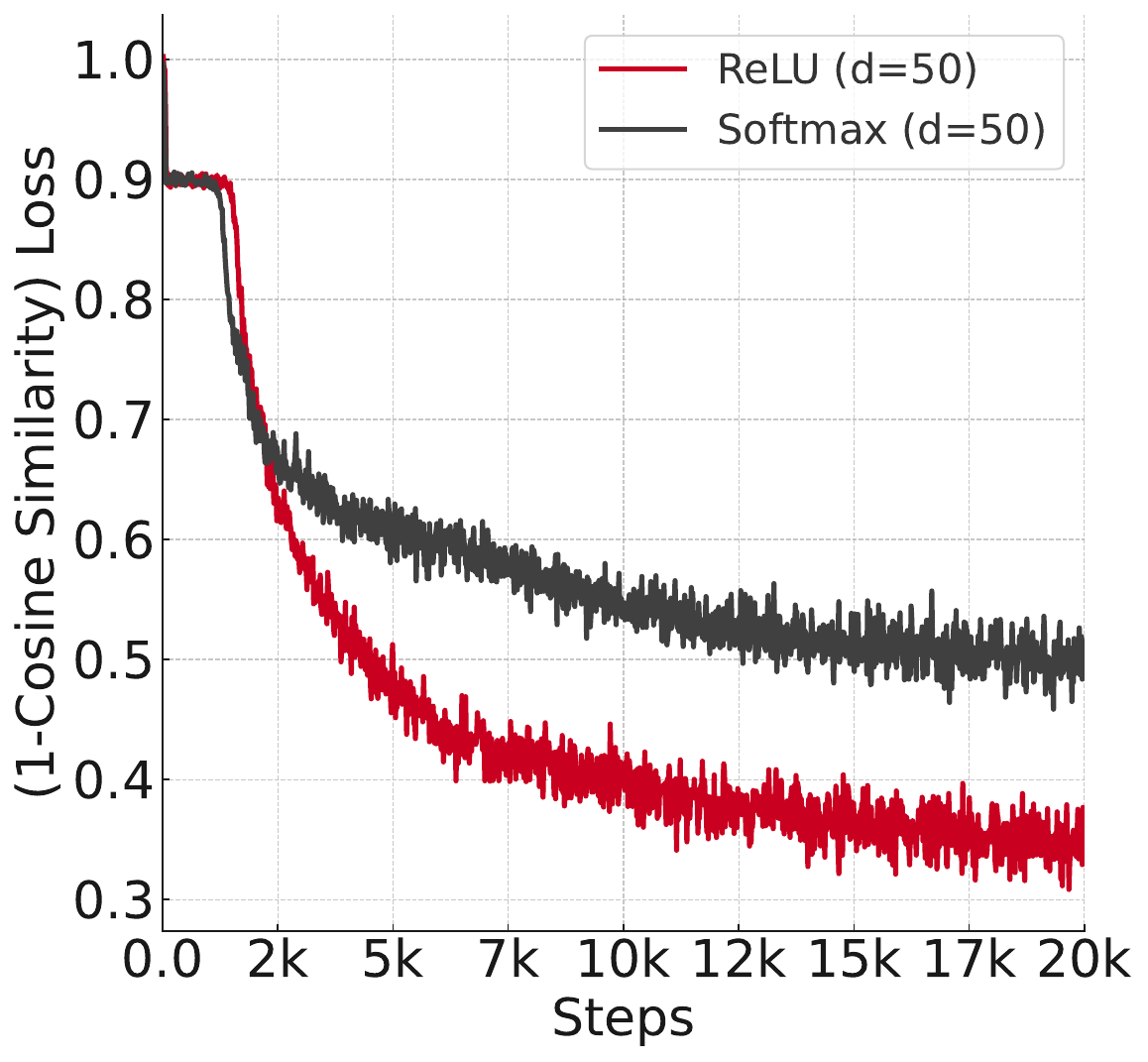}
    \endminipage
    \vspace{-1em}
    \caption{{\textbf{Test Set Loss Curve Comparison between Softmax and ReLU Transformers (Top-1 Eigenvector Prediction).}
    We use a $3$-layer, $2$ head, $64$ hidden dimension Transformer to predict the top-$1$ eigenvector across all experiments in this figure. We observe that ReLU-based Transformers consistently outperform their Softmax counterparts. The performance gap grows as $d$ increases.}}
    \label{fig:softmax_loss}
\end{figure}

\begin{figure}[!h]
    \centering
    \begin{minipage}{0.49\textwidth}
        \centering
        \includegraphics[width=\linewidth]{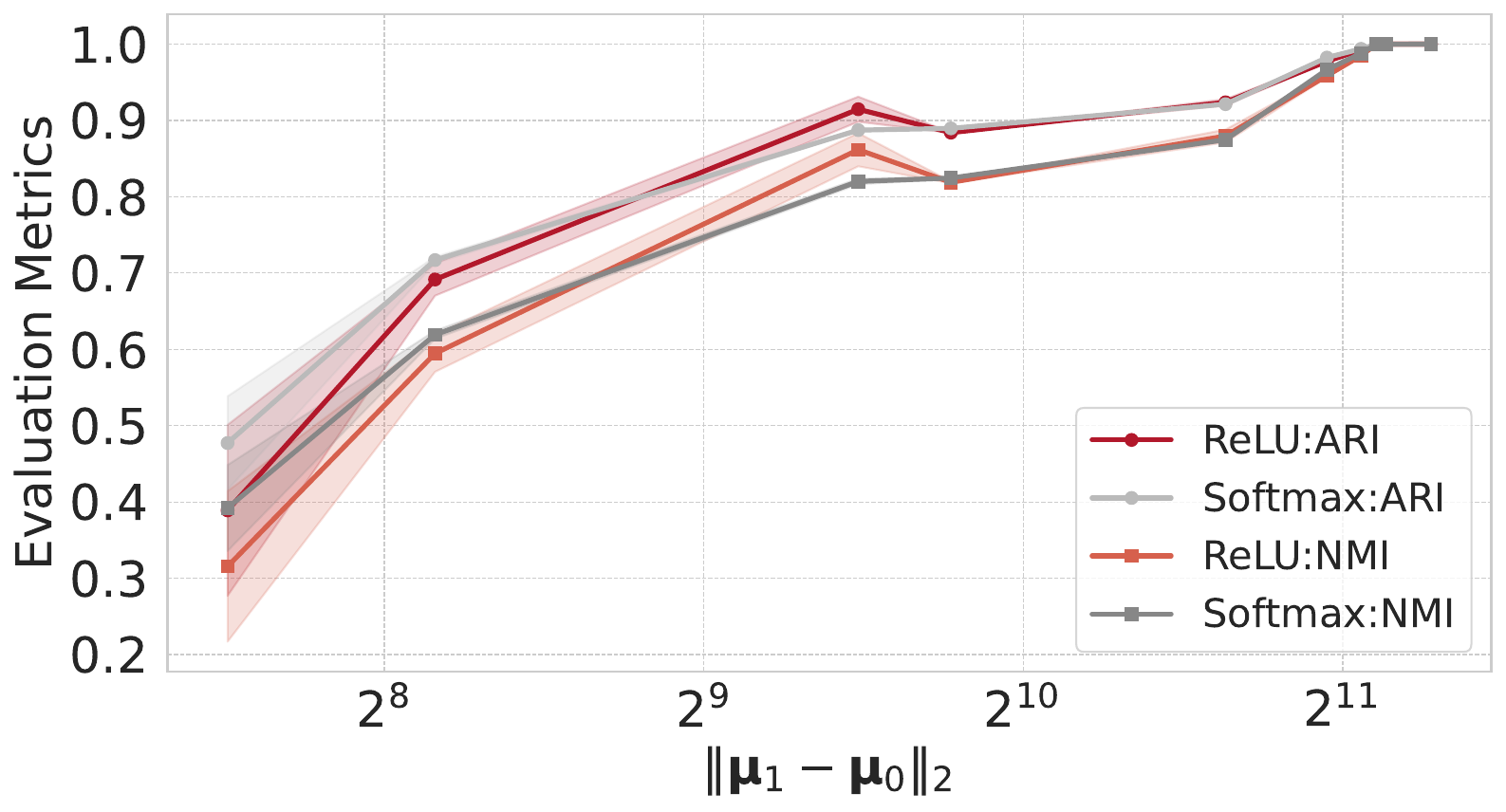}
    \end{minipage}
    \hfill 
    \begin{minipage}{0.49\textwidth}
        \centering
        \includegraphics[width=\linewidth]{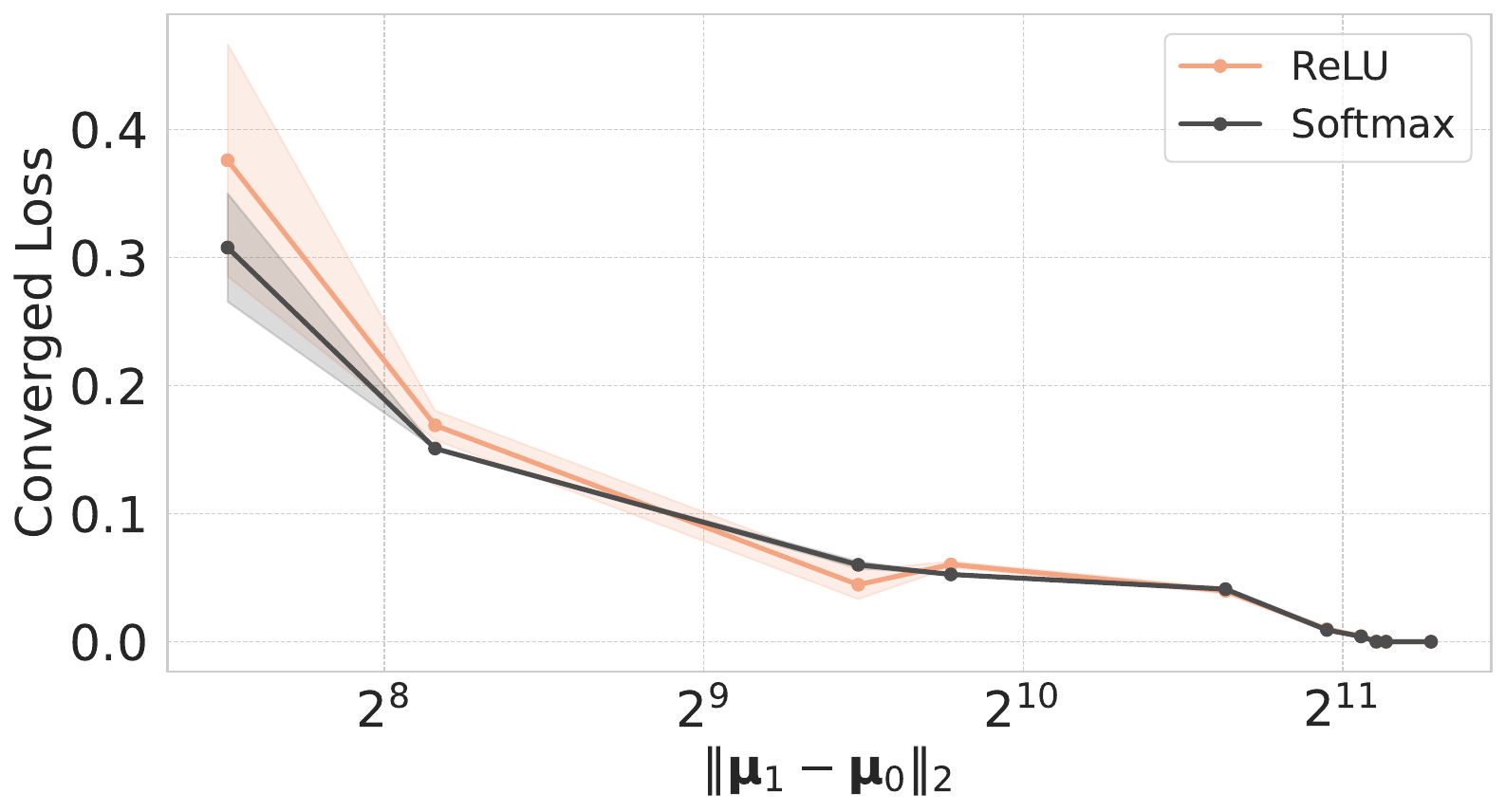}
    \end{minipage}
    \caption{{\textbf{Comparison of the ReLU and Softmax Transformers on the Real World Dataset.}
    We compare the performance of ReLU Transformer and Softmax Transformer on 2-class clustering problems on the Covertype dataset.
    \emph{Left: Performance Comparison on ARI and NMI metrics.}
    \emph{Right: Performance Comparison on the Converged Loss.}
    }
    }
\label{fig:relu_softmax_covtype}
\end{figure}

\begin{figure}[!h]
    \centering
    \begin{minipage}{0.49\textwidth}
        \centering
        \includegraphics[width=\linewidth]{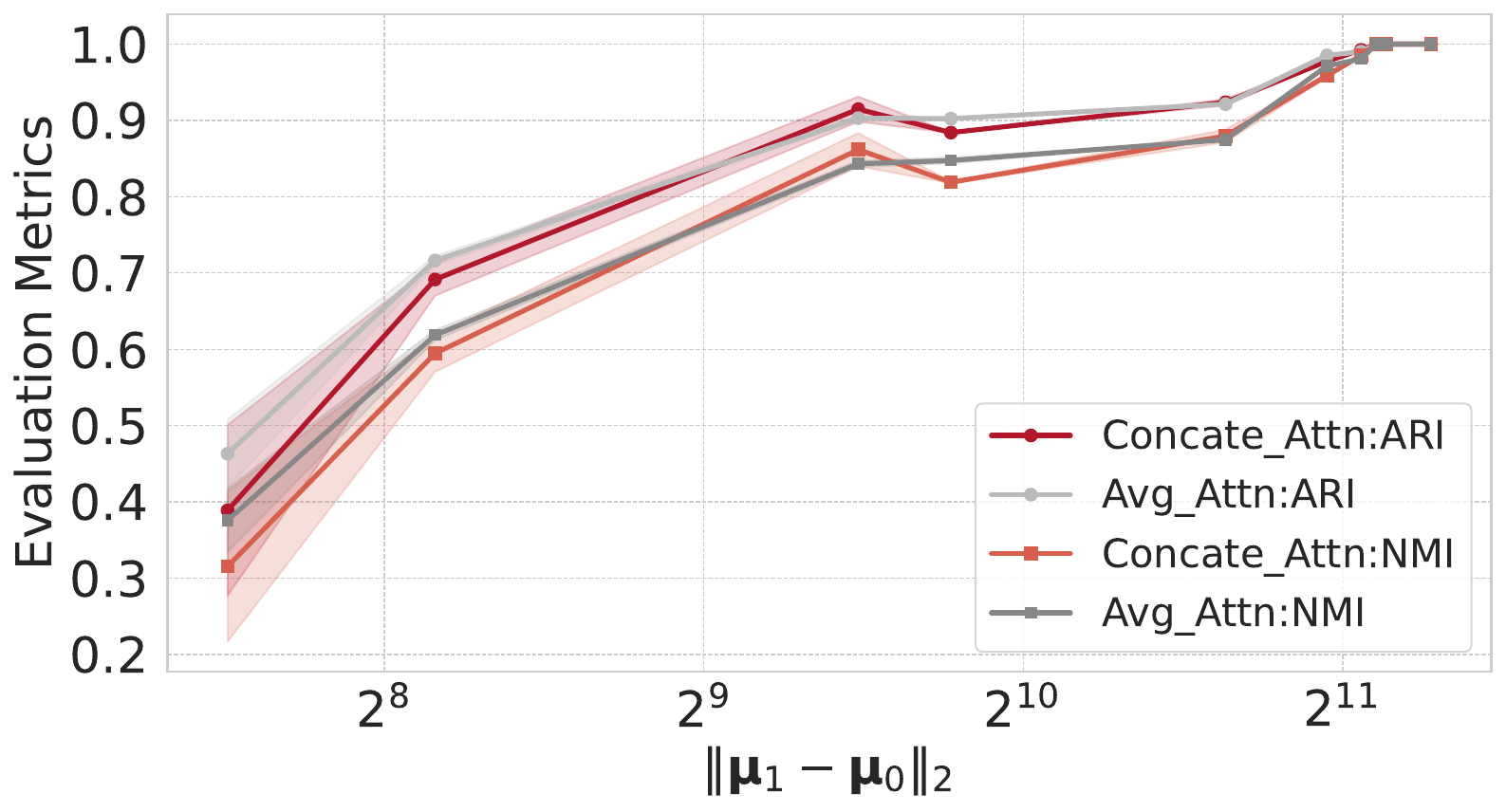}
    \end{minipage}
    \hfill 
    \begin{minipage}{0.49\textwidth}
        \centering
        \includegraphics[width=\linewidth]{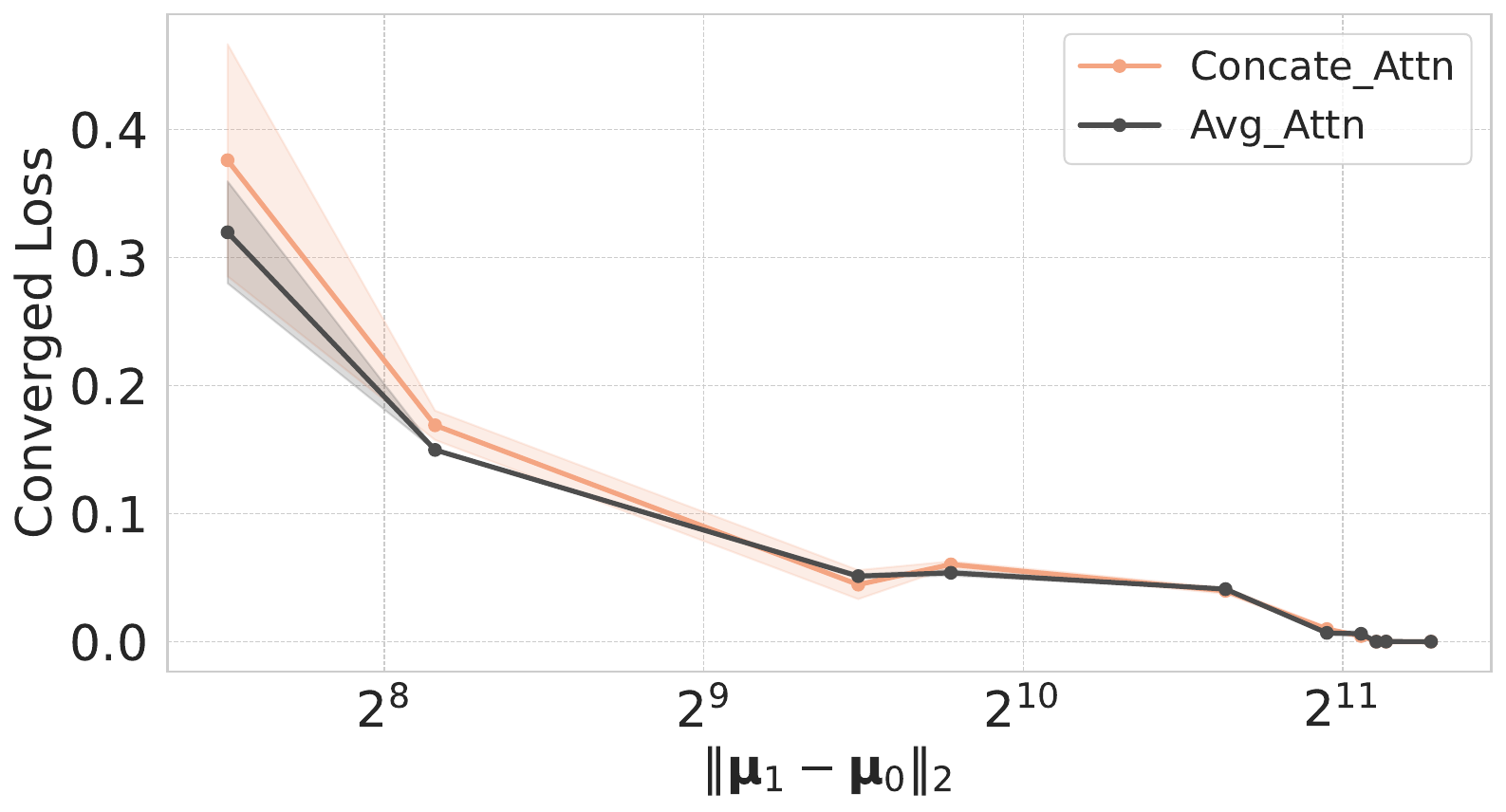}
    \end{minipage}
    \caption{{\textbf{Comparison of Concatenated Attention and Averaged Attention on the Real World Dataset.}
    \emph{Left: Performance Comparision on ARI and NMI.}
    \emph{Right: Performance Comparision on Converged Loss.}
    }
    }
\label{fig:concate_averaged_covtype}
\end{figure}

This section provides additional details on the simulations from three different perspectives: (1) The comparison between Transformers versus the Power Methods in solving the PCA problem; (2) The comparison between the ReLU Attention and Softmax Attention in the experiments; (3) The comparison between the Concatenated multi-head Attention and the Averaged multi-head Attention in the theoretical results of this work. This section hopes to give empirical evidence bridging the assumptions of our theoretical results and the Transformer networks used in practice.

\paragraph{The Power Method versus Transformers.} 
We compare the predicted eigenvector of the Transformer and the Power Method in figure \ref{fig:mnist_power_method_eigenvector} and table \ref{tab:mul_k_vector_encoder}.
We want to understand the sharpness of our construction of the Transformers in the approximation of the Power Method. When choosing the number of iterations $\tau$ of the Power Method for the Transformer to compare with, we use $L = 2\tau + 4(k-1) +1$ as derived in theorem \ref{thm3.1}, but with $k$ replaced by $k-1$ since we only need to remove principal components when we want to predict $k \geq 2$ principal components. In the left and middle subplots of figure \ref{fig:mnist_power_method_eigenvector} and table \ref{tab:mul_k_vector_encoder}, we observe a linear relationship between the number of layers of Transformers and the number of iterations of the Power Method. This verifies our theoretical result. We evaluate this correspondence on the MNIST dataset, as shown in the right plot of figure \ref{fig:mnist_power_method_eigenvector}. Our empirical evaluation justifies the tightness of our approximation.

\paragraph{Softmax versus ReLU Attention.}
We perform experiments comparing the performance of the ReLU Attention used in the theoretical results and the Softmax Attention that is commonly used in practice.
{We run two different experiments on PCA and Clustering respectively to show that the choice of the ReLU Attention in our theoretical construction is equivalent to the Softmax version on the empirical performance. The first experiment is in figure \ref{fig:softmax_loss} where we report the training loss of predicting top-$1$ principal components with different input dimensions $d$. The second experiment is in figure \ref{fig:relu_softmax_covtype} where we compare between the ReLU and Softmax Attentions on the Covertype dataset experiments with different $\Vert\bfa\mu_1-\bfa\mu_0\Vert_2$.}

\paragraph{Concatenated versus Averaged Multi-head Attentions.}
Similar to the previous paragraph, here we hope to compare the performance between the Concatenated multi-head Attention and the Averaged multi-head Attention studied in the theoretical results of this work. {We compare the Averaged Attention with the Concatenated Attention \cite{vaswani2017attention} on a two-class clustering problem using the Covtype dataset, as shown in figure \ref{fig:concate_averaged_covtype}. We observe that the difference in the clustering performance with different $\Vert\bfa\mu_1-\bfa\mu_0\Vert_2$ is very small regardless of the metrics used.}

\section{Discussions}\label{sect5}
This section discusses the limitations in this work and potential future working directions.  Our limitations in the theoretical results can be summarized as follows: \textbf{(1)} From the theoretical perspective, our results guarantee the performance of ERM solutions whereas the true estimator is obtained through the stochastic gradient descent method; \textbf{(2)} The sharpness of our results might not be easily verifiable as the approximation of the Transformer part is subject to constraints from both the network structure and the algorithms that achieve the sharp rate. We believe that future working directions include: \textbf{(1)} understanding the Multi-clustering Problem on Transformers; \textbf{(2)} replacing the ReLU Attentions with Softmax Attentions and try to obtain similar results; \textbf{(3)} generalizing the results in this work to other unsupervised learning problems.

\label{sec:meth}

\addtolength{\textheight}{-.3in}%

\label{sec:conc}


\newpage
\begin{center}
{\large\bf SUPPLEMENTARY MATERIAL}
\end{center}

\appendix

{
\setlength{\parskip}{-0em}
\startcontents[sections]
\printcontents[sections]{ }{1}{}
}


\section{Additional Theoretical Background}

\begin{definition}[Sufficiently Smooth $d$-variate function]
    Denote $\msf B_{\infty}^d(R):=[-R,R]^d$ as the standard $\ell_\infty$ ball in $\bb R^d$.
    We say a function $g:\bb R^d\to\bb R$ is $(R,C_\ell)$ smooth if for $s=\lceil(d-1)/2\rceil+2$, $g$ is a $C^s$ function on $\msf B^d_\infty(\bb R)$ and
    \begin{align*}
        \sup_{\bfa z\in\msf B_{\infty}^d(R)}\Vert\nabla^dg(\bfa z)\Vert_{\infty}=\sup_{\bfa z\in\msf B^d_\infty(R)}\max_{j_1,\ldots,j_i\in[d]}\lef|\pta_{x_{j_1}\ldots x_{j_{i}}}g(\bfa x)\rig|\leq L_i
    \end{align*}
    for all $i\in\{0,1,\ldots,s\}$, with $\max_{0\leq i\leq s}L_iR^i\leq C_\ell$.
\end{definition}
\begin{definition}[Approximability by sum of ReLUs \citep{bai2024transformers}]\label{def6} A function $g:\bb R^k\to\bb R$ is $(\epsilon_{approx}, R,M,C)$-approximable by sum of ReLUs if there exists a function $f_{M,C}$ such that
\begin{align*}
    f_{M,C}(\bfa z)=\sum_{m=1}^Mc_m\sigma(\bfa a_m^\top[\bfa z;1])\text{ with }\sum_{m=1}^M|c_m|\leq C,\;\max_{m\in[M]}\Vert \bfa a_m\Vert_1\leq 1,\quad \bfa a_m\in\bb R^{k+1},\; c_m\in\bb R,
\end{align*}
such that $\sup_{\bfa z\in\msf B_\infty^k(R)}|g(\bfa z)-f_{M,C}(\bfa z)|\leq \epsilon_{approx}$.
\end{definition}

\section{Proofs}
\subsection{Proof of Proposition \ref{genbound}}
\begin{proof}
    The proof follows from \citep{wainwright2019high}, using the fact that for all $\bfa\theta\in \Theta(B_{\bfa\theta},B_M)$, we have 
    \begin{align*}
        \frac{1}{n}\sum_{j=1}^nL\lef(TF_{\wha\theta}(\bfa H_i),\bfa V_i\rig)\leq\frac{1}{n}\sum_{j=1}^n L\lef(TF_{\bfa\theta}(\bfa H_i),\bfa V_i\rig),
    \end{align*} 
    it is not hard to show that
    \begin{align*}
        L\lef(TF_{\wha\theta}(\bfa H),\bfa V\rig)\leq \inf_{\bfa\theta\in \Theta(B_{\bfa\theta},B_M)}\bb E\lef[L(TF_{\wha\theta}(\bfa H),\bfa V)\rig]+2\sup_{\bfa\theta\in \Theta(B_{\bfa\theta},B_M)}| X_{\bfa\theta}|,
    \end{align*}
    where $ X_{\bfa\theta}=\frac{1}{n}\sum_{j=1}^nL(TF_{\bfa\theta}(\bfa H_i),\bfa V_i)-\bb E[L(TF_{\bfa\theta}(\bfa H),\bfa V)]$ is the empirical process indexed by $\bfa\theta$.
    The tail bound for empirical process requires us to verify a few regularity conditions \citep{gine2016mathematical} on the function $L$ and the set $\Theta$, given as follows:
    \begin{enumerate}
        \item The metric entropy of an operator norm ball 
        $$\log N(\delta, B_{\Vert\cdot\Vert_{op}}(\delta),\Vert\cdot\Vert_{op})\leq CLB_M D^2\log\lef(1+2(B_{\theta}+B_X+k)/\delta\rig).$$
        \item $L(TF_{\bfa\theta}(\bfa H),\bfa V)\leq C\sqrt k$.
        \item The Lipschitz condition of Transformers satisfies that for all $\bfa\theta_1,\bfa\theta_2\in\Theta(B_{\bfa\theta}, B_{M})$, we have $L(TF_{\bfa\theta_1}(\bfa H),\bfa V) - L(TF_{\bfa\theta_2}(\bfa H),\bfa V)\leq C L B_1^{L}\Vert\bfa\theta_1-\bfa\theta_2\Vert_{op}$ where $B_1=B_{\theta}^4B_X^3$.
    \end{enumerate}
    The first and second verifications follow immediately from J.2 in \citep{bai2024transformers}. The third verification is given upon noticing that as $L(\bfa x,\bfa y)=\Vert\bfa x-\bfa y\Vert_2$,
    \begin{align*}
        \sup_{\bfa\theta,\bfa H,\bfa V}L(TF_{\bfa\theta}(\bfa H),\bfa V)\leq C\sqrt k,\qquad \Vert\nabla_{\bfa x} L\Vert\leq C.
    \end{align*}
    Further note that $\Vert\tda W_0\Vert_2\asymp\Vert\tda W_1\Vert_2\asymp 1$.
    Given the above result, and corollary J.1 in \citep{bai2024transformers}, we can show that
    \begin{align*}
        L(TF_{\bfa\theta_1}(\bfa H),\bfa V)-L(TF_{\bfa\theta_2}(\bfa H),\bfa V)\leq C L B_1^{L}\Vert\bfa\theta_1-\bfa\theta_2\Vert_{op},
    \end{align*}
    where $B_1 = B^4_{\theta}B^3_{X}$.
    Therefore, using the uniform concentration bound given by proposition A.4 we can show that with probability at least $1-\delta$, we have
    \begin{align*}
        \sup_{\bfa\theta\in\Theta(B_{\bfa\theta},B_M)} |X_{\theta}|\leq C\sqrt k\sqrt{\frac{LB_MD^2\log(B_{\theta}+B_X+k)+\log(1/\delta)}{n}}.
    \end{align*}
    Therefore, replacing $D$ with $Ckd$ we complete the proof.
\end{proof}

\subsection{Proof of Theorem \ref{thm3.1}}
\begin{proof}
    Our proof can be disected into the following setps: 1. We construct a Transformer with fixed parameters that performs
    (1) The computation of the symmetrized covariate matrix; (2) The approximation of the power method; (3) The removal of the principal eigenvectors; (4) Adjust the dimension of the output through multiplying the two matrices $\tda W_0$ and $\tda W_1$ on the left and right.
    \begin{center}
        \textbf{1. The Covariate Matrix.}
    \end{center}
    To compute the covariate matrix $\bfa X\bfa X^\top$, we construct $\bfa H=\begin{bmatrix}
        \bfa X_1,\ldots,\bfa X_N\\
        \tda p_{1,1},\ldots,\tda p_{1,N}\\
        \tda p_{2,1},\ldots,\tda p_{2,N}\\
        \vdots\\
        \tda p_{\ell,1},\ldots,\tda p_{\ell,N}
    \end{bmatrix}=\begin{bmatrix}
        \bfa X\\
        \bfa P
    \end{bmatrix}$
    we let the number of heads $m=2$ and construct the first covariate layer as follows,
    \begin{align}
     & \bfa V_1^{cov}=I_D=-\bfa V_2^{cov},\quad \bfa Q_1^{cov,\top}\bfa K_1^{cov}=-\bfa Q_2^\top\bfa K_2=\begin{bmatrix}
         \bfa 0_{N+1\times d} & I_d &\bfa 0\\
         \bfa 0&\bfa 0&\bfa 0
     \end{bmatrix}\in\bb R^{D\times D},\nnb\\
     &\tda p_{1,\ell,j}=\bfa 0,\qquad\tda p_{2,\ell,j}=\begin{cases}
         \mbbm 1_{\ell=j}\quad&\text{ when }\ell\leq d\\
         0&\text{ when }\ell>d
     \end{cases}.
    \end{align}
    Under the above construction, we obtain that
    \begin{align*}
        &\bfa Q_1^\top\bfa K_1\bfa H=\begin{bmatrix}
            I_{d}&\bfa 0\\
            \bfa 0&\bfa 0
        \end{bmatrix}\in\bb R^{D\times N},\quad\bfa Q_2^\top\bfa K_2\bfa H=\begin{bmatrix}
            -I_{d}&\bfa 0\\
            \bfa 0&\bfa 0
        \end{bmatrix}\in\bb R^{D\times N},\\
        &\sigma(\bfa H^\top\bfa Q_1^\top\bfa K_1\bfa H)+\sigma(\bfa H^\top\bfa Q_2^\top\bfa K_2\bfa H)=\begin{bmatrix}
            \bfa X^\top,\bfa 0
        \end{bmatrix}\in\bb R^{N\times N}.
    \end{align*}
   We further obtain that
    \begin{align*}
        \frac{1}{N}\sum_{m=1}^M(\bfa V_m\bfa H)\times \sigma\lef((\bfa Q_m\bfa H)^\top(\bfa K_m\bfa H)\rig)=\begin{bmatrix}
        \bfa 0&\bfa 0\\
            \bfa X\bfa X^\top\in\bb R^{d\times d}&\bfa 0\\
            \bfa 0&\bfa 0
        \end{bmatrix}\in\bb R^{D\times D}.
    \end{align*}
    Therefore, the output is given by  $\tda H^{cov}=\begin{bmatrix}
            \bfa X\\
            \bfa X\bfa X^\top,\bfa 0\\
            \tda p_{2,1},\ldots,\tda p_{2,N}\\
            \tda p_{\ell,1},\ldots,\tda p_{\ell,N}
        \end{bmatrix}$. 
    \begin{center}
        \textbf{2. The Power Iteration.}
    \end{center}
    Then we consider constructing a single attention layer that approximates the power iteration. This step involves two important operations: (1) Obtaining the vector given by $\bfa X\bfa X^\top\bfa v$. (2) Approximation of the value of the inverse norm given by $1/\Vert \bfa X\bfa X^\top\bfa v\Vert_2$. We show that one can use the multihead ReLU Transformer to achieve both goals simulatenously, whose parameters are given by 
    \begin{align*}
        \bfa V_{1}^{pow,1}&=-\bfa V_{2}^{pow,1}=\begin{bmatrix}
            \bfa 0_{(3d+1)\times (2d+1)}&\bfa 0&\bfa 0\\
            \bfa 0_{(d)\times(2d+1)}&I_d&\bfa 0\\
            \bfa 0&\bfa 0&\bfa 0
        \end{bmatrix},\\
        \bfa Q_1^{pow,1}&=-\bfa Q_1^{pow,1}=\begin{bmatrix}
           \bfa 0_{(d+1)\times (d+1)}& \bfa 0&\bfa 0\\
            \bfa 0_{d\times (d+1)}&I_d&\bfa 0\\
            \bfa 0&\bfa 0&\bfa 0
        \end{bmatrix},\\
        \bfa K_1^{pow,1}&=\bfa K_2^{pow,1}=\begin{bmatrix}
            \bfa 0_{(3d+1)\times(3d+1)}&\bfa 0&\bfa 0\\
            \bfa 0_{d\times(3d+1)}&I_d&\bfa 0\\
            \bfa 0&\bfa 0&\bfa 0
        \end{bmatrix},\qquad\tda p_{4,j} = \bfa 0\text{ for all }j\in[N].
    \end{align*}
    Given the above formulation, we are able to show that 
    \begin{align*}
       &\bfa Q_2^{pow,1}\tda H^{cov}=-\bfa Q_1^{pow,1}\tda H^{cov} = \begin{bmatrix}
        \bfa 0_{(2d+1)\times N}\\
            \bfa X\bfa X^\top,\bfa 0\\
            \bfa 0
        \end{bmatrix},\\
        &\bfa K_2^{pow,1}\tda H^{cov}=\bfa K_1^{pow,1}\tda H^{cov}=\begin{bmatrix}
            \bfa 0_{2d+1}\\
            \tda p_{3,1},\bfa 0\\
            \bfa 0
        \end{bmatrix},
    \end{align*}
    which implies that 
    \begin{align*}
    \sum_{m\in\{1,2\}}\sigma((\bfa Q_m^{pow, 1}\tda H^{cov})^\top\bfa K_m^{pow,1}\tda H^{cov})=\begin{bmatrix}
           \bfa 0& \bfa 0_{d\times(2d+1)}&\\
            \bfa X\bfa X^\top\tda p_{3,1}&\bfa 0_{d\times(N-1)}\\
            \bfa 0&\bfa 0&
        \end{bmatrix}.
    \end{align*}
    Then we can show that 
    \begin{align*}
        \tda H^{pow,1}-\tda H^{cov}&=\sum_{m\in\{1,2\}}\bfa V_m^{pow,1}\tda H^{cov}\times\sigma((\bfa Q_m^{pow,1}\tda H^{cov})^\top\bfa K_m^{pow,1}\tda H^{cov})\\
        &=\begin{bmatrix}
            \bfa 0_{3d+1}\\
            \bfa X\bfa X^\top\tda p_{3,1},\bfa 0_{d\times(N-1)}\\
            \bfa 0
        \end{bmatrix}.
    \end{align*}
    Therefore, we conclude that the output of the first power iteration layer is given by
    \begin{align*}
        \tda H^{pow,1}=\begin{bmatrix}
            \bfa X\\
            \tda y^\top\\
            \bfa X\bfa X^\top,\bfa 0\\
            \tda p_{2,1},\ldots,\tda p_{2,N}\\
            \tda p_{3,1},\ldots,\tda p_{3,N}\\
            \bfa X\bfa X^\top\tda p_{3,1},\bfa 0\\
            \tda p_{5,1},\ldots,\tda p_{5,N}\\
            \vdots\\
            \tda p_{\ell,1},\ldots,\tda p_{\ell,N}
        \end{bmatrix}.
    \end{align*}
    Then, using lemma \ref{reluapprox}, we design an extra attention layer that performs the normalizing procedure, with the following parameters for all $m\in[M]$,
    \begin{align*}
        &\bfa V_{m}^{pow,2}=\begin{bmatrix}
            \bfa 0_{d\times (4d+1)}&c_m\bfa I_d&\bfa 0\\
            \bfa 0&\bfa 0&\bfa 0
        \end{bmatrix},\qquad\bfa Q_{m}^{pow,2}=\begin{bmatrix}
            \bfa 0_{d\times(2d+1)}&\bfa I_{d}&\bfa 0\\
            \bfa 0&\bfa 0&\bfa 0
        \end{bmatrix},\\
        &\bfa K_m^{pow,2}=\begin{bmatrix}
            \bfa 0_{1\times (3d+1)}&\bfa a_m^\top&\bfa 0\\
            \vdots\\
            \bfa 0_{1\times(3d+1)}&\bfa a_m^\top&\bfa 0\\
            \bfa 0_{(D-d)\times(3d+1)}&\bfa 0&\bfa 0
        \end{bmatrix}.
    \end{align*}
    Under the above construction, we obtain that
    \begin{align*}
        (\bfa Q_m^{pow,2}\tda H^{pow,1})^\top =\begin{bmatrix}
            I_{d\times d}&\bfa 0\\
            \bfa 0&\bfa 0
        \end{bmatrix},\qquad \bfa K_m^{pow,2}\tda H^{pow,1} = \begin{bmatrix}
            \bfa a_m^\top\bfa X\bfa X^\top\tda p_{3,1}&\bfa 0\\
            \vdots\\
            \bfa a_m^\top\bfa X\bfa X^\top\tda p_{3,1}&\bfa 0\\
            \bfa 0_{(D-d)\times 1}&\bfa 0
        \end{bmatrix}.
    \end{align*}
    Then, given $\bfa V_m^{pow,2}$ we can show that under the condition given by lemma \ref{reluapprox}, we have
    \begin{align*}
        \bigg\Vert&\sum_{m=1}^M\bfa V_m^{pow,2}\tda H^{pow,1}\sigma\lef((\bfa Q_m^{pow,2}\tda H^{pow,1})^\top(\bfa K_m^{pow,2}\tda H^{pow,1})\rig)-\begin{bmatrix}
        \bfa 0_{4d+1}\\
            \frac{\bfa X\bfa X^\top\tda p_{3,1}}{\Vert\bfa X\bfa X^\top\tda p_{3,1}\Vert_2}-\bfa X\bfa X^\top\tda p_{3,1},\bfa 0\\
            \bfa 0
        \end{bmatrix}\bigg\Vert_{\infty}\\
        &<\epsilon,
    \end{align*}
    Moreover, we can further achieve that
    \sm{\begin{align*}
        \Bigg\Vert\sum_{m=1}^M\bfa V_m^{pow,2j}\tda H^{pow,1}\sigma\lef((\bfa Q_m^{pow,2}\tda H^{pow,1})^\top (\bfa K_m^{pow,2}\tda H^{pow,1})\rig)&-\begin{bmatrix}
            \bfa 0_{4d+1}\\
            \frac{\bfa X\bfa X^\top\tda p_{3,1}}{\Vert\bfa X\bfa X^\top\tda p_{3,1}\Vert_2}-\bfa X\bfa X^\top\tda p_{3,1},\bfa 0\\
            \bfa 0
        \end{bmatrix}\Bigg\Vert_2\\
        &<\epsilon\Vert\bfa X\bfa X^\top\tda p_{3,1}\Vert_2.
    \end{align*}}
    Hence, using the fact that $\tda H^{pow,2}=\tda H^{pow,1}+\sum_{i=1}^m\bfa V_m^{pow,2}\tda H^{pow,1}\sigma\lef((\bfa Q_m^{pow,2}\tda H^{pow,1})^\top(\bfa K_m^{pow,2}\tda H^{pow,1})\rig)$, we obtain that
    \begin{align*}
        \Bigg\Vert\tda H^{pow,2}-\begin{bmatrix}
            \bfa X\\
            \tda y\\
            \bfa X\bfa X^\top,\bfa 0\\
            \tda p_{2,1},\ldots,\tda p_{2,N}\\
            \tda p_{3,1},\ldots,\tda p_{3,N}\\
            \frac{\bfa X\bfa X^\top\tda p_{3,1}}{\Vert\bfa X\bfa X^\top\tda p_{3,1}\Vert_2},\ldots\bfa 0\\
            \vdots
        \end{bmatrix}\Bigg\Vert_{2}<\epsilon\Vert\bfa X\bfa X^\top\tda p_{3,1}\Vert_2.
    \end{align*}
    Then we construct another attention layer, which performs similar calculations as that of $pow,1$ but switch the rows of $\tda p_{3,1}$ with that of $\frac{\bfa X\bfa X^\top\tda p_{3,1}}{\Vert\bfa X\bfa X^\top\tda p_{3,1}\Vert_2}$. Our construction for the third layer is given by
    \begin{align*}
        \bfa V_{1}^{pow,3}&=-\bfa V_{2}^{pow,3}=\begin{bmatrix}
            \bfa 0_{(3d+1)\times (2d+1)}&\bfa 0&\bfa 0\\
            \bfa 0_{d\times(2d+1)}&I_d&\bfa 0\\
            \bfa 0&\bfa 0&\bfa 0
        \end{bmatrix},\\
        \bfa Q_1^{pow,3}&=-\bfa Q_2^{pow,3}=\begin{bmatrix}
           \bfa 0_{(3d+1)\times (d+1)}&\bfa 0& \bfa 0\\
            \bfa 0_{d\times (d+1)}&I_d&\bfa 0\\
            \bfa 0&\bfa 0&\bfa 0
        \end{bmatrix},\\
        \bfa K_1^{pow,3}&=\bfa K_2^{pow,3}=\begin{bmatrix}
            \bfa 0_{(4d+1)\times(4d+1)}&\bfa 0&\bfa 0\\
            \bfa 0_{d\times(4d+1)}&I_d&\bfa 0\\
            \bfa 0&\bfa 0&\bfa 0
        \end{bmatrix},\qquad\tda p_{4,j} = \bfa 0\text{ for all }j\in[N].
    \end{align*}
    Given the above construction, we can show that
    \begin{align*}
       &\bfa Q_2^{pow,3}\tda H^{pow,2}=-\bfa Q_1^{pow,3}\tda H^{pow,2}=\begin{bmatrix}
           &\bfa 0_{(3d+1)\times N}&\\
           \bfa 0_{}&\bfa X\bfa X^\top &\bfa 0\\
           &\bfa 0&
       \end{bmatrix},\quad \bfa K_2^{pow,3}\tda H^{pow,2}=\bfa K_1^{pow,3}\tda H^{pow,2},\\
       &\Bigg\Vert\bfa K_2^{pow,3}\tda H^{pow,2}-\begin{bmatrix}
            \bfa 0_{(3d+1)\times N}\\
            \frac{\bfa X\bfa X^\top\tda p_{3,1}}{\Vert\bfa X\bfa X^\top\tda p_{3,1}\Vert_2},\bfa 0\\
            \bfa 0
        \end{bmatrix}\Bigg\Vert_{2}\leq\epsilon\Vert\bfa X\bfa X^\top\tda p_{3,1}\Vert_2.
    \end{align*}
    Then, using the fact that given $\bfa X_1,\bfa X_2$ with $\Vert \bfa X_1-\bfa X_2\Vert_2\leq\delta_0$, we have
$        \Vert\bfa X\bfa X^\top(\bfa X_1-X_2)\Vert_2\leq \Vert\bfa X\bfa X^\top\Vert_2\delta_0 $. 
    Hence, collecting the above pieces, we have
    \begin{align*}
      \Bigg\Vert\sum_{m=1}^2\bfa V_m^{pow,3}\tda H^{pow,2}\sigma\lef((\bfa Q_2^{pow,3}\tda H^{pow,2})^{\top}\bfa K_2^{pow,3}\tda H^{pow,2}\rig)&-\begin{bmatrix}
          \bfa 0_{(3d+1)\times N}\\
          \frac{(\bfa X\bfa X^\top)^2\tda p_{3,1}}{\Vert\bfa X\bfa X^\top\tda p_{3,1}\Vert_2}-\frac{\bfa X\bfa X^\top\tda p_{3,1}}{\Vert\bfa X\bfa X^\top\tda p_{3,1}\Vert_2},\bfa 0\\
          \bfa 0
      \end{bmatrix}\Bigg\Vert_{2}\\
      &\leq\epsilon\Vert\bfa X\bfa X^\top\Vert_2\lef\Vert\bfa X\bfa X^\top\tda p_{3,1}\rig\Vert_2.
    \end{align*}
    Henceforth, one can further show that
$      \Bigg\Vert \tda H^{pow,3}- \begin{bmatrix}
            \bfa X\\
            \tda y\\
            \bfa X\bfa X^\top,\bfa 0\\
            \tda p_{2,1},\ldots,\tda p_{2,N}\\
            \tda p_{3,1},\ldots,\tda p_{3,N}\\
            \frac{(\bfa X\bfa X^\top)^2\tda p_{3,1}}{\Vert\bfa X\bfa X^\top\tda p_{3,1}\Vert_2},\bfa 0\\
            \tda p_{5,1},\ldots,\tda p_{5,N}\\
            \vdots\\
            \tda p_{\ell,1},\ldots,\tda p_{\ell,N}
        \end{bmatrix}\Bigg\Vert_{2}\leq\epsilon \Vert\bfa X\bfa X^\top\Vert_2\Vert\bfa X\bfa X^\top\tda p_{3,1}\Vert_2$.
    Consider we are doing in total of $\tau$ power iterations, we can set for all $\tau\in\bb N^*$,
    \begin{align*}
        &\bfa V_m^{pow,2\tau+1}=\bfa V_m^{pow,3},\quad \bfa Q_m^{pow,2\tau+1}=\bfa Q_m^{pow,3},\quad\bfa K_m^{pow,2\tau+1}=\bfa K_m^{pow,3},\\
        &\bfa V_m^{pow,2\tau+2}=\bfa V_m^{pow,4},\quad\bfa Q_m^{pow,2\tau+2}=\bfa Q_m^{pow,4},\quad\bfa K_m^{pow,2\tau+2}=\bfa K_m^{pow,4}.
    \end{align*}
    Therefore, taking another layer of normalization, we can show that
    \begin{align*}
        \Bigg\Vert \tda H^{pow,3}- \begin{bmatrix}
            \bfa X\\
            \tda y\\
            \bfa X\bfa X^\top,\bfa 0\\
            \tda p_{2,1},\ldots,\tda p_{2,N}\\
            \tda p_{3,1},\ldots,\tda p_{3,N}\\
            \frac{(\bfa X\bfa X^\top)^2\tda p_{3,1}}{\Vert\bfa X\bfa X^\top\tda p_{3,1}\Vert_2^2},\bfa 0\\
            \tda p_{5,1},\ldots,\tda p_{5,N}\\
            \vdots\\
            \tda p_{\ell,1},\ldots,\tda p_{\ell,N}
        \end{bmatrix}\Bigg\Vert_{2}\leq2\epsilon \Vert\bfa X\bfa X^\top\Vert_2.
    \end{align*}
    Then, using the sublinearity of errors, we can show that for $\tau\in\bb N$,
    \begin{align*}
              \Bigg\Vert \tda H^{pow,2\tau+2}- \begin{bmatrix}
            \bfa X\\
            \tda y\\
            \bfa X\bfa X^\top,\bfa 0\\
            \tda p_{2,1},\ldots,\tda p_{2,N}\\
            \tda p_{3,1},\ldots,\tda p_{3,N}\\
            \tda p_{3,1}^{(\tau)},\bfa 0\\
            \tda p_{5,1},\ldots,\tda p_{5,N}\\
            \vdots\\
            \tda p_{\ell,1},\ldots,\tda p_{\ell,N}
        \end{bmatrix}\Bigg\Vert_{\infty}\leq \tau\epsilon\Vert\bfa X\bfa X^\top\Vert_2,\quad\tda p^{(\tau)}_{3,1}=\frac{\bfa X\bfa X^\top\tda p_{3,1}^{(\tau-1)}}{\lef\Vert\bfa X\bfa X^\top\tda p_{3,1}^{(\tau-1)}\rig\Vert_2},\quad\tda p_{3,1}^{(0)}=\tda p_{3,1}.
    \end{align*}
    If we denote $\bfa v_i$ as the eigenvector corresponds to the $i$ th largest eigenvalue of $\bfa X\bfa X^\top$. Let the eigenvalues of $\bfa X\bfa X^\top$ be denoted by $\lambda_1>\lambda_2>\cdots>\lambda_n$. Given $|\tda p_{3,1}^\top\bfa v_1|>\delta$ and $|\sqrt{\lambda_1}-\sqrt{\lambda_2}|=\Omega(1)$. Theorem 3.11 in \citep{blum2020foundations} page 53 shows that given $k=\frac{\log(1/\epsilon_0\delta)}{2\epsilon_0}$ and $\Vert\tda p_{3,1}^{(\tau)}\Vert_2 = \Vert\bfa v_1\Vert_2=1$, one immediately obtains that
    \begin{align*}  
    \tda p_{3,1}^{(\tau),\top}\bfa v_1\geq 1- \epsilon_0,\qquad\Vert\tda p_{3,1}^{(\tau)}-\bfa v_1\Vert_2 = \sqrt{2-2\bfa v_1^\top\tda p_{3,1}^{(\tau)}}=\sqrt{2\epsilon_0}.
    \end{align*}
    And we also consider the approximation of the maximum eigenvalue. Note that using $\Vert\bfa v_1\Vert_2=1$, we have 
    \begin{align*}
       \Vert\bfa X\bfa X^\top\Vert_2&=\Vert\bfa X\bfa X^\top\bfa v_1\Vert_2=\lef\Vert\bfa X\bfa X^\top\tda p_{3,1}^{(\tau)}+\bfa X\bfa X^\top(\bfa v_1-\tda p_{3,1}^{(\tau)})\rig\Vert_2\\
       &\leq\Vert\bfa X\bfa X^\top\tda p_{3,1}^{(\tau)}\Vert_2+\Vert\bfa X\bfa X^\top(\bfa v_1-\tda p_{3,1}^{(\tau)})\Vert_2\\
       &\leq\Vert\bfa X\bfa X^\top\tda p_{3,1}^{(\tau)}\Vert_2+\Vert\bfa X\bfa X^\top\Vert_2\Vert\bfa v_1-\tda p_{3,1}^{(\tau)}\Vert_2.
   \end{align*}
   Similarly we can also derive that $\Vert\bfa X\bfa X^\top\Vert_2\geq \Vert\bfa X\bfa X^\top\tda p_{3,1}^{(\tau)}\Vert_2-\Vert\bfa X\bfa X^\top\Vert_2\Vert\bfa v_1-\tda p_{3,1}\Vert_2$. Then we show that
   \begin{align*}
       \Big|\Vert\bfa X\bfa X^\top\Vert_2-\Vert\bfa X\bfa X^\top\tda p_{3,1}^{(\tau)}\Vert_2\Big|\leq\Vert\bfa X\bfa X^\top\Vert_2\Vert\bfa v_1-\tda p_{3,1}^{(\tau)}\Vert_2\leq\sqrt{2\epsilon_0}\Vert\bfa X\bfa X^\top\Vert_2.
   \end{align*}
    \begin{center}
        \textbf{3. The Removal of principal Eigenvectors.}
    \end{center}
    After $\tau$ iterates on the power method, we need to remove the principal term from the matrix $\bfa X\bfa X^\top$, achieved through two important steps: (1) The computation of the estimated eigenvalue $\Vert\bfa X\bfa X^\top\tda p_{3,1}\Vert_2$. (2) The construction of the low rank update $\tda p_{3,1}\tda p_{3,1}^\top$. For step (1), we consider the following construction:
    \begin{align*}
        &\bfa V_1^{rpe,1}=-\bfa V_2^{rpe,1}=\begin{bmatrix}
            \bfa 0_{(3d+1)\times(2d+1)}&\bfa 0&\bfa 0\\
            \bfa 0_{d\times(2d+1)}&I_d&\bfa 0\\
            \bfa 0&\bfa 0&\bfa 0
        \end{bmatrix},\quad\bfa Q_1^{rpe,1} =-\bfa Q_2^{rpe,1}= \begin{bmatrix}
            \bfa 0_{(d+1)\times(d+1)}&\bfa 0&\bfa 0\\
            \bfa 0_{d\times(d+1)}& I_d&\bfa 0\\
            \bfa 0&\bfa 0&\bfa 0
        \end{bmatrix},\\
        &\bfa K_1^{rpe,1}=\bfa K_2^{rpe,1} = \begin{bmatrix}
            \bfa 0_{(4d+1)\times(4d+1)}&\bfa 0&\bfa 0\\
            \bfa 0_{d\times(4d+1)}& I_d&\bfa 0\\
            \bfa 0&\bfa 0&\bfa 0
        \end{bmatrix}.
    \end{align*}
    Note that the above construction is similar to the first layer of the power method. Under this construction, we can show that
    \begin{align}\label{diffhrpe1}
        &\tda H^{rpe,1}=\tda H^{pow,2\tau+2}+\sum_{m\in\{1,2\}}\bfa V_m^{rpe,1}\sigma((\bfa Q_m^{rpe,1}\tda H^{pow,2\tau+2})^\top(\bfa K_m^{rpe,1}\tda H^{pow,2\tau+2})),\nnb\\
        &\Bigg\Vert\tda H^{rpe,1}-\ub{\begin{bmatrix}
            \bfa X\\
            \tda y\\
            \bfa X\bfa X^\top,\bfa 0\\
            \tda p_{2,1},\ldots,\tda p_{2,N}\\
            \tda p_{3,1},\ldots,\tda p_{3,N}\\
            \tda p_{3,1}^{(\tau)},\bfa 0\\
            \bfa X\bfa X^\top\tda p_{3,N}^{(\tau)},\bfa 0\\
            \tda p_{6,1},\ldots,\tda p_{6,N}\\
            \vdots\\
            \tda p_{\ell,1},\ldots,\tda p_{\ell,N}
        \end{bmatrix}}_{=:\bfa H^{rpe,1}}\Bigg\Vert_2\leq C\tau\epsilon\Vert\bfa X\bfa X^\top\Vert_2^2,\qquad\tda p_{5,i}=\bfa 0,\quad\forall i\in[N].
    \end{align}
    Then, we construct the next layer, using the notations in lemma \ref{reluapprox}, for $M\geq \Vert\bfa X\bfa X^\top\Vert_2^d\frac{C(d)}{\epsilon^2}$ for all $m\in[M]$ we have 
    \begin{align*}
       &\bfa V_m^{rpe,2}=\begin{bmatrix}
            \bfa 0_{d\times(4d+1)}&d_m\bfa I_d&\bfa 0\\
            \bfa 0&\bfa 0&\bfa 0
       \end{bmatrix},\qquad \bfa Q_m^{rpe,2} =\begin{bmatrix}
           \bfa 0_{d\times(2d+1)}&\bfa I_d&\bfa 0\\
           \bfa 0&\bfa 0&\bfa 0
       \end{bmatrix},\\
       &\bfa K_m^{rpe,2} =\begin{bmatrix}
           \bfa 0_{1\times(5d+1)}&\bfa b_m^\top&\bfa 0\\
           \vdots&&\\
           \bfa 0_{1\times(5d+1)}&\bfa b_m^\top&\bfa 0 \\
           \bfa 0_{(D-d)\times(5d+1)}&\bfa 0&\bfa 0
       \end{bmatrix}.
    \end{align*}
    Given the above construction, we subsequently show that
    \begin{align*}
       (\bfa Q_m^{rpe,2}\tda H^{rpe,1})^\top=\begin{bmatrix}
           I_{d\times d}&\bfa 0\\
           \bfa 0&\bfa 0
       \end{bmatrix},\qquad \bfa K_m^{rpe,2}\tda H^{rpe,1} = \begin{bmatrix}
            \bfa b_m^\top\bfa X\bfa X^\top\tda p_{3,1}^{(\tau)}&\bfa 0\\
            \vdots &\vdots\\
            \bfa b_m^\top\bfa X\bfa X^\top\tda p_{3,1}^{(\tau)}&\bfa 0\\
            \bfa 0_{(D-d)\times 1}&\bfa 0
        \end{bmatrix}.
    \end{align*}
    Hence, given the construction of $\bfa V_m^{rpe,2}$, we can show that $
        \tda H^{rpe,2} $ satisfies
        \begin{align*}
           \tda H^{rpe,2} &= \tda H^{rpe,1}+\sum_{m\in[M]}\bfa V_m^{rpe,2}\tda H^{rpe,1}\times\sigma\lef((\bfa K^{rpe,2}_m\tda H^{rpe,1})^\top(\bfa Q_m^{rpe}\tda H^{rpe,1})\rig)\\
           &=\ub{\bfa H^{rpe,1}+\sum_{m\in[M]}\bfa V_m^{rpe,2}\bfa H^{rpe,1}\times\sigma\lef((\bfa K_m^{rpe,2}\bfa H^{rpe,1})^\top(\bfa Q_m^{rpe,2}\bfa H^{rpe,1})\rig)}_{=:\wha H^{rpe,1}}\\
           &+\lef(\tda H^{rpe,1}-\bfa H^{rpe,1}\rig)+\sum_{m\in[M]}\bfa V_m^{rpe,2}\tda H^{rpe,1}\times\sigma\lef((\bfa K_m^{rpe,2}\tda H^{rpe,1})^\top\bfa Q_m^{rpe,2}\tda H^{rpe,1}\rig)\\
           &-\sum_{m\in[M]}\bfa V_m^{rpe,2}\tda H^{rpe,1}\times\sigma\lef((\bfa K_m^{rpe,2}\tda H^{rpe,1})^\top\bfa Q_m^{rpe,2}\tda H^{rpe,1}\rig).
        \end{align*}
    We note that by lemma \ref{reluapprox} we can show that
    \begin{align*}
        \Bigg\Vert\wha H^{rpe,1}-\begin{bmatrix}
            \bfa X\\
            \tda y\\
            \bfa X\bfa X^\top,\bfa 0\\
            \tda p_{2,1}\ldots,\tda p_{2,N}\\
            \tda p_{3,1}\ldots,\tda p_{3,N}\\
            \tda p_{3,1}^{(\tau)},\bfa 0\\
            \Vert\bfa X\bfa X^\top\tda p_{3,N}^{(\tau)}\Vert_2^{\frac{1}{2}}\tda p_{3,1}^{(\tau)},\bfa 0\\
            \vdots\\
            \tda p_{\ell,1},\ldots\tda p_{\ell, N}
        \end{bmatrix}\Bigg\Vert_2\leq C\tau\epsilon\Vert\bfa X\bfa X^\top\Vert_2^2.
    \end{align*}
    Then the rest of the proof focuses on showing that the rest of the terms are small. Note that using \eqref{diffhrpe1}, we show that
    \begin{align*}
        \lef\Vert\tda H^{rpe,1}-\bfa H^{rpe,1}\rig\Vert_2\leq\tau\epsilon\Vert\bfa X\bfa X^\top\Vert_2^2.
    \end{align*}
    And for the last term, we can show that
    \begin{align*}
        \Big\Vert\sum_{m\in[M]}\bfa V_m^{rpe,2}\tda H^{rpe,1}&\times\sigma\lef((\bfa K_m^{rpe,2}\tda H^{rpe,1})^\top(\bfa Q_m^{rpe,2}\tda H^{rpe,1})\rig)\\
        &-\sum_{m\in[M]}\bfa V_m^{rpe,2}\bfa H^{rpe,1}\times\sigma\lef((\bfa K_m^{rpe,2}\bfa H^{rpe,1})^\top(\bfa Q_m^{rpe,2}\bfa H^{rpe,1})\rig)\Big\Vert_2\\
        &\leq C\tau\epsilon\Vert\bfa X\bfa X^\top\Vert_2^2. 
    \end{align*}
    Collecting the above pieces, we finally show that
    \begin{align*}
        \Bigg\Vert\tda H^{rpe,2}-\begin{bmatrix}
            \bfa X\\
            \bfa X\bfa X^\top,\bfa 0\\
            \tda p_{2,1},\ldots,\tda p_{2,N}\\
            \tda p_{3,1},\ldots,\tda p_{3,N}\\
            \tda p_{3,1}^{(\tau)},\bfa 0\\
            \Vert\bfa X\bfa X^\top\tda p_{3,1}^{(\tau)}\Vert_2^{\frac{1}{2}}\tda p_{3,1}^{(\tau)},\bfa 0\\
            \tda p_{6,1},\ldots,\tda p_{6,N}\\
            \vdots\\
            \tda p_{\ell,1},\ldots,\tda p_{\ell,N}
        \end{bmatrix} \Bigg\Vert_2\leq C\tau\epsilon\Vert\bfa X\bfa X^\top\Vert_2^2.
    \end{align*}
    Then we construct another layer to remove the principal components from the matrix $\bfa X\bfa X^\top$, given by
    \begin{align*}
        &-\bfa V_1^{rpe,3}=\bfa V_2^{rpe,3}= \begin{bmatrix}
            \bfa 0_{(d+1)\times(4d+1)}&\bfa 0&\bfa 0\\
            \bfa 0& I_d&\bfa 0\\
            \bfa 0&\bfa 0&\bfa 0
        \end{bmatrix},\qquad\bfa Q_1^{rpe,3}=-\bfa Q_2^{rpe,3}=\begin{bmatrix}
            \bfa 0_{d\times(4d+1)}&I_d&\bfa 0\\
            \bfa 0&\bfa 0&\bfa 0
        \end{bmatrix},\\
        &\bfa K_1^{rpe,3}=\bfa K_2^{rpe,3}=\begin{bmatrix}
            \bfa 0_{d\times(4d+1)}& I_d&\bfa 0\\
            \bfa 0&\bfa 0&\bfa 0
        \end{bmatrix}.
    \end{align*}
    Then we can show that
    \begin{align*}
        (\bfa Q_1^{rpe,3}\tda H^{rpe,2})^\top =\begin{bmatrix}
            \Vert\bfa X\bfa X^\top\tda p_{3,1}^{(\tau)}\Vert_2^{\frac{1}{2}}\tda p_{3,1}^{(\tau),\top}&\bfa 0\\
            \bfa 0&\bfa 0
        \end{bmatrix},\qquad\bfa K_1^{rpe,3}\tda H^{rpe,2}=\begin{bmatrix}
            \bfa 0&\bfa 0\\
            I_d&\bfa 0\\
            \bfa 0&\bfa 0
        \end{bmatrix}.
    \end{align*}
    Then it is further noted that
    $
       -(\bfa Q_2^{rpe,3}\tda H^{rpe,2})^\top\bfa K_2^{rpe,3}\tda H^{rpe,2}= (\bfa Q_1^{rpe,3}\tda H^{rpe,2})^\top\bfa K_1^{rpe,3}\tda H^{rpe,2} 
    $ satisfies
    \begin{align*}
        \Bigg\Vert(\bfa Q_1^{rpe,3}\tda H^{rpe,2})^\top\bfa K_1^{rpe,3}\tda H^{rpe,2}-\begin{bmatrix}
            \Vert\bfa X\bfa X^\top\tda p_{3,1}^{(\tau)}\Vert_2^{\frac{1}{2}}\tda p_{3,1}^{(\tau),\top}&\bfa 0\\
            \bfa 0&\bfa 0
        \end{bmatrix}\Bigg\Vert_2\leq C\tau\epsilon\Vert\bfa X\bfa X^\top\Vert_2^2.
    \end{align*}
    And therefore, combining our construction for $\bfa V_m$, it is noted that
    \begin{align*}
        \Bigg\Vert\sum_{m=1}^2\bfa V_m\tda H^{rpe,2}\times\sigma((\bfa Q_1^{rpe,3}\tda H^{rpe,2})^\top\bfa K_1^{rpe,3}\tda H^{rpe,2})&-\begin{bmatrix}
            \bfa 0_{(d+1)\times N}\\
            -\Vert\bfa X\bfa X^\top\tda p_{3,1}^{(\tau)}\Vert_2\tda p_{3,1}^{(\tau)}\tda p_{3,1}^{(\tau),\top},\bfa 0\\
            \bfa 0
        \end{bmatrix}\Bigg\Vert_2\\
        &\leq C\tau\epsilon\Vert\bfa X\bfa X^\top\Vert_2^2.
    \end{align*}
    Therefore, we can further show that
    \begin{align*}
        \tda H^{rpe,3} =\tda H^{rpe,2}+ \sum_{m=1}^2\bfa V_{m}^{rpe,3}\tda H^{rpe,2}\times\sigma\lef((\bfa Q_m^{rpe,3}\tda H^{rpe,2})^\top\bfa K_m^{rpe,3}\tda H^{rpe,2}\rig)
    \end{align*}
    satisfies
    \begin{align*}
       \Bigg\Vert \tda H^{rpe,3}-\begin{bmatrix}
            \bfa X\\
            \bfa X\bfa X^\top-\Vert\bfa X\bfa X^\top\tda p_{3,1}^{(\tau)}\Vert_2\tda p_{3,1}^{(\tau)}\tda p_{3,1}^{(\tau),\top},\bfa 0\\
            \tda p_{2,1},\ldots,\tda p_{2,N}\\
            \tda p_{3,1},\ldots,\tda p_{3,N}\\
            \tda p_{3,1}^{(\tau)},\bfa 0\\
            \tda p_{5,1},\ldots,\tda p_{5,N}\\
            \vdots\\
            \tda p_{\ell,1},\ldots,\tda p_{\ell,N}
        \end{bmatrix}\Bigg\Vert_2\leq C\tau\epsilon\Vert\bfa X\bfa X^\top\Vert_2^2.
    \end{align*}
    And we can construct another layer to remove the term $\Vert\bfa X\bfa X^\top\tda p_{3,1}^{(\tau)}\Vert_2^{\frac{1}{2}}\tda p_{3,1}^{(\tau)}$, which is achieved by 
    \begin{align*}
        &-\bfa V_1^{rpe,4} = \bfa V_2^{rpe,4} =  \begin{bmatrix}
            \bfa 0_{(4d+1)\times(4d+1)}&\bfa 0&\bfa 0\\
            \bfa 0_{d\times(4d+1)}& I_d&\bfa 0\\
            \bfa 0&\bfa 0&\bfa 0
        \end{bmatrix},\\
        &\bfa Q_1^{rpe,4} = -\bfa Q_2^{rpe,4}=\begin{bmatrix}
            &\bfa 0_{(3d+1)\times D}&\\
            \bfa 0_{d\times (2d+1)}& I_d&\bfa 0\\
            &\bfa 0&
        \end{bmatrix},\\
        &\bfa K_1^{rpe,4}= \bfa K_2^{rpe,4} =\begin{bmatrix}
            &\bfa 0_{(3d+1)\times D}&\\
            \bfa 0_{d\times (2d+1)}& I_d&\bfa 0\\
            &\bfa 0&
        \end{bmatrix}.
    \end{align*}
    Using the above construction, we can further show that
    \begin{align*}
        \Bigg\Vert\tda H^{rpe,4}-\begin{bmatrix}
            \bfa X\\
            \bfa X\bfa X^\top-\Vert\bfa X\bfa X^\top\tda p_{3,1}^{(\tau)}\Vert_2\tda p_{3,1}^{(\tau)}\tda p_{3,1}^{(\tau),\top},\bfa 0\\
            \tda p_{2,1},\ldots,\tda p_{2,N}\\
            \tda p_{3,1},\ldots,\tda p_{3,N}\\
            \tda p_{3,1}^{(\tau)},\bfa 0\\
            \tda p_{5,1},\ldots,\tda p_{5,N}\\
            \vdots\\
            \tda p_{\ell,1},\ldots,\tda p_{\ell,N}
        \end{bmatrix}\Bigg\Vert_2\leq C\tau\epsilon\Vert\bfa X\bfa X^\top\Vert_2^2.
    \end{align*}
    And then we proceed to recover the rest of the $k$ principal eigenvectors using similar model architecture given by the ones used by the Power Iterations. For the computation over the $\tau$-th eigenvector, we denote $\tda H^{pow, \eta, 1}$ till $\tda H^{pow, \eta, \tau}$ to be the intermediate states corresponding to the $\eta$-th power iteration. We denote $\tda H^{rpe,\eta,\tau_0}$ to be the output of $\eta$-th removal of principal eigenvector layers for the $\tau$-th eigenvector. Furthermore, we iteratively define
    \begin{align*}
        \bfa A_1 =\bfa X\bfa X^\top-\Vert\bfa X\bfa X^\top\tda p_{3,1}^{(\tau)}\Vert_2\tda p_{3,1}^{(\tau)}\tda p_{3,1}^{(\tau),\top},\qquad \bfa A_{i+1}=\bfa A_{i}-\Vert\bfa A_i\tda p_{3,i}^{(\tau)}\Vert_2\tda p_{3,i}^{(\tau)}\tda p_{3,i}^{(\tau),\top},\qquad\forall i\in[k].
    \end{align*}
    Then, applying the subadditivity of the $2$-norm, we can show that
    \begin{align*}
        \Bigg\Vert\tda H^{rpe,4,k}-\begin{bmatrix}
            \bfa X\\
            \bfa A_{k+1},\bfa 0\\
            \tda p_{2,1},\ldots,\tda p_{2,N}\\
            \tda p_{3,1},\ldots,\tda p_{3,N}\\
            \tda p_{3,1}^{(\tau)},\bfa 0\\
            \tda p_{3,2}^{(\tau)},\bfa 0\\
            \vdots\\
            \tda p_{3,k}^{(\tau)},\bfa 0
        \end{bmatrix}\Bigg\Vert_2\leq C\tau k\epsilon\Vert\bfa X\bfa X^\top\Vert_2^2.
    \end{align*}
    For simplicity, we denote $\tda A=\begin{bmatrix}
        \bfa X\\
        \bfa A_{k+1},\bfa 0\\
        \tda p_{2,1},\ldots,\tda p_{2,N}\\
        \tda p_{3,1},\ldots,\tda p_{3,N}
    \end{bmatrix}$ and $\tda P=\begin{bmatrix}
        \tda p_{3,1}^{(\tau)}\\
        \tda p_{3,2}^{(\tau)}\\
        \vdots\\
        \tda p_{3,k}^{(\tau)}
    \end{bmatrix}$ from here. 
    \begin{center}
        \textbf{4. Finishing Up.}
    \end{center}
    The finishing up phase considers constructing $\tda W_0$ and $\tda W_1$ that adjust the final output format. Our construction gives the following 
    \begin{align*}
        \tda W_0 = \begin{bmatrix}
            \bfa 0,I_{kd}
        \end{bmatrix},\qquad\tda W_1=\begin{bmatrix}
            1\\
            \bfa 0_{N-1}
        \end{bmatrix}.
    \end{align*}
    And we can show that
    \begin{align*}
        \Bigg\Vert\tda W_0\tda H^{rpe,4,k}\tda W_1-\begin{bmatrix}
            \tda p_{3,1}^{(\tau)}\\
            \tda p_{3,2}^{(\tau)}\\
            \vdots\\
            \tda p_{3,k}^{(\tau)}
        \end{bmatrix}\Bigg\Vert_2\leq C\tau k\epsilon\Vert\bfa X\bfa X^\top\Vert_2^2.
    \end{align*}
        We further use the result given by lemma \ref{lm3.2}, denote $a_{\eta}:=\lef\Vert\bfa v_{\eta}-\tda p_{3,\eta}^{(\tau)}\rig\Vert_2$, $\wh\lambda_{\eta}=\lef\Vert \bfa A_{\eta}\tda p_{3,\eta}^{(\tau)}\rig\Vert_2$, and $b_{\eta}:=|\lambda_{\eta}-\wh \lambda_{\eta}|$ for $\eta\in[k]$, we obtain that for all $\eta\geq 1$, given the number of iterations $\tau\geq C\frac{\log(1/\epsilon_0\delta)}{2\epsilon_0}$ where the constant value $C$ depends on $d$,
    \begin{align*}
    a_{\eta+1}\leq\frac{\max_{i\in[\eta]}b_{i}+\sum_{i=1}^{\eta}2\lambda_i a_i}{\Delta},\qquad b_{\eta+1}\leq\frac{2\lambda_{\eta+1}}{\Delta}\bl\max_{i\in[\eta]}b_{\eta}+\sum_{i=1}^{\eta} 2\lambda_ia_i\br+\lambda_{\eta+1}\sqrt{2\epsilon_0}.
    \end{align*}
    Further note that the starting point is given by
    $a_{1}\leq \sqrt{2\epsilon_0}$, $b_1\leq\lambda_{1}\sqrt{2\epsilon_0}$. Introducing $A_{\eta}=\sum_{i=1}^{\eta} 2\lambda_ia_i$, we obtain that
$        A_{\eta+1} =\sum_{i=1}^{\eta+1}2\lambda_ia_i = A_{\eta}+2\lambda_{\eta+1}a_{\eta+1}$ which alternatively implies that
    \begin{align*}
        \frac{1}{2\lambda_{\eta+1}}(A_{\eta+1}-A_{\eta})\leq\frac{\max_{i\in[\eta]}b_i+A_{\eta}}{\Delta},\qquad b_{\eta+1}\leq\frac{2\lambda_{\eta+1}}{\Delta}\bl\max_{i\in[\eta]}b_{\eta}+A_{\eta}\br+\lambda_{\eta+1}\sqrt{2\epsilon_0}.
    \end{align*}
    We use the fact 
        $\frac{\lambda_{\eta}}{\Delta}>1$ for all $\eta\in[k]$
    to show the following
    \begin{align*}
A_{\eta+1}+\max_{i\in[\eta+1]}b_i\leq\frac{5\lambda_{\eta+1}}{\Delta}\lef(A_{\eta}+\max_{i\in[\eta]} b_i\rig)+\lambda_1\sqrt{2\epsilon_0}, \qquad A_1+b_1=2\lambda_1\sqrt{2\epsilon_0},
    \end{align*}
    which implies that
    \begin{align}\label{iterativeeq}
    A_{\eta+1}+\max_{i\in[\eta+1]}b_i+&\frac{\lambda_1\sqrt{2\epsilon_0}}{\frac{5\lambda_{1}}{\Delta}-1}\leq\frac{5\lambda_{1}}{\Delta}\bl A_{\eta}+\max_{i\in[\eta]}b_i +\frac{\lambda_1\sqrt{2\epsilon_0}}{\frac{5\lambda_{1}}{\Delta}-1}\br,\nnb\\
        A_{\eta+1}+\max_{i\in[\eta+1]}b_i+&\frac{\lambda_1\sqrt{2\epsilon_0}}{\frac{5\lambda_{1}}{\Delta}-1}\leq\bl A_1+b_1+\frac{\lambda_1\sqrt{2\epsilon_0}}{\frac{5\lambda_1}{\Delta}-1}\br\prod_{i=1}^{\eta}\bl\frac{5\lambda_{i+1}}{\Delta}\br\nnb\\
        &=\lambda_1\sqrt{2\epsilon_0}\bl 2+\frac{1}{\frac{5\lambda_1}{\Delta}-1} \br\prod_{i=1}^{\eta}\bl\frac{5\lambda_{i+1}}{\Delta}\br.
    \end{align}
    Therefore, applying the inequality given by \eqref{iterativeeq} we can show that, for $\eta\leq k$, we have for all $\eta\in[k-1]$,
    \begin{align*}
        a_{\eta+1}&\leq\frac{1}{\Delta}\bl \lambda_1\sqrt{2\epsilon_0}\bl 2+\frac{1}{\frac{5\lambda_1}{\Delta}-1}\br\prod_{i=1}^{\eta}\bl\frac{5\lambda_{i+1}}{\Delta}\br-\frac{\lambda_1\sqrt{2\epsilon_0}}{\frac{5\lambda_1}{\Delta}-1}\br,\\
        b_{\eta+1}&\leq\frac{2\lambda_{\eta}\lambda_1\sqrt{2\epsilon_0}}{\Delta}\bl 2+\frac{1}{\frac{5\lambda_1}{\Delta}-1}\br\prod_{i=1}^{\eta}\bl\frac{5\lambda_{i+1}}{\Delta}\br+\lambda_{\eta+1}\sqrt{2\epsilon_0}.
    \end{align*}
    Therefore collecting pieces, we conclude that there exists a transformer with number of layers $2\tau+4k+1$ and number of heads $M\leq \lambda_1^d\frac{C(d)}{\epsilon^2}$ such that
    the final output $\wha v_1,\ldots,\wha v_{k}$ given by the Transformer model satisfy $\forall \eta\in[k-1]$,
    \begin{align*}
        \lef\Vert\wha v_{\eta+1}-\bfa v_{\eta+1}\rig\Vert_2\leq C\tau \epsilon\lambda_1^2+\frac{1}{\Delta}\bl\lambda_1\sqrt{2\epsilon_0}\bl 2+\frac{1}{\frac{5\lambda_1}{\Delta}-1}\br\prod_{i=1}^{\eta}\bl\frac{5\lambda_{i+1}}{\Delta}\br-\frac{\lambda_1\sqrt{2\epsilon_0}}{\frac{5\lambda_1}{\Delta}-1}\br.
    \end{align*}
    And the rest of the result directly follows.
    \end{proof}

    \subsection{Proof of Lemma \ref{lm3.1}}
        \begin{proof}
        To prove the above result, we consider two events
$            A_1 =\lef\{\Vert \bfa y\Vert_2\geq \sqrt{\frac{1}{\epsilon}}\rig\}$, $A_2=\lef\{|\bfa y^\top\bfa v|\leq \sqrt{\epsilon}\rig\}$, then we can show that
        \begin{align*}
           \lef\{ |\bfa v^\top\bfa x|\leq\frac{1}{\sqrt\epsilon}\rig\}\subset A_1\cup A_2\quad\Rightarrow\quad\bb P\bl|\bfa v^\top\bfa x|\leq\sqrt{\epsilon}\br\leq\bb P(A_1)+\bb P(A_2).
        \end{align*}
        And we use the tail bound for Chi-square given by \citep{laurent2000adaptive} to obtain that as $\epsilon<d^{-1}$,
        \begin{align*}
            \bb P(A_1) = \bb P\lef(\Vert\bfa y\Vert_2^2\geq \epsilon^{-1}\rig)\leq\exp\lef(-C\epsilon^{-1}\rig).
        \end{align*}
        And similarly, consider the event $A_2$, note that $\bfa y^\top\bfa v\sim N(0,1)$, we use the cdf of the folded normal distribution to obtain that
        \begin{align*}
            \bb P\lef(A_2\rig) = \bb P\lef(\lef|\bfa v^\top\bfa y\rig|\leq\sqrt{\epsilon}\rig)=erf\lef(\frac{\sqrt\epsilon}{\sqrt 2}\rig)=\frac{2}{\sqrt{\pi}}\bl \sqrt\epsilon-\frac{(\sqrt\epsilon)^3}{3}+\frac{(\sqrt\epsilon)^5}{10}-\frac{(\sqrt\epsilon)^7}{42}\br\leq\frac{\sqrt\epsilon}{\sqrt{\pi}}.
        \end{align*}
        Then we obtain that 
        \begin{align*}
            \bb P\bl|\bfa v^\top\bfa x|\leq\sqrt\epsilon\br\leq \frac{\sqrt\epsilon}{\sqrt{\pi}}+\exp\lef(-C\epsilon^{-1}\rig).
        \end{align*}
        Consider in total of $k$ independent random vectors $\bfa X_1,\ldots,\bfa X_k$, and arbitrary $k$ vectors $\bfa v_1,\ldots,\bfa v_k$, we can show that
        \begin{align*}
            \bb P\bl\exists i\text{ such that }&\bfa X_i^\top\bfa v_i\leq\epsilon\br\leq k\bb P\bl \bfa X_1^\top\bfa v_1\leq\epsilon\br\leq\frac{k\sqrt{\epsilon}}{\sqrt{\pi}}+k\exp(-C\epsilon^{-1}).
        \end{align*}
    \end{proof}

\subsection{Proof of Lemma \ref{lm3.2}}

\begin{lemma}\label{lm3.2}
Assume that the correlation matrix $\bfa X\bfa X^\top$ has eigenvalues $\lambda_1>\lambda_2>\ldots>\lambda_{k}$. Assume that the eigenvectors are given by $\bfa v_1,\bfa v_2,\ldots,\bfa v_n$ and the eigenvalues satisfy $\inf_{i\neq j}|\lambda_i-\lambda_j|=\Delta$. Then, given that the estimate for the first $\tau$ eigenvectors satisfy
$    \bfa v_i^\top\wha v_i\geq 1-\epsilon_i $ and the eigenvalues satisfy $|\lambda_i-\wh\lambda_i|\leq\delta_i$, the principal eigenvector of $\bfa X\bfa X^\top-\sum_{i=1}^{\tau}\wh\lambda_i\wha v_i\wha v_i^\top$ denoted by $\tda v_{\tau+1}$ satisfies
\begin{align*}
    \Vert\tda v_{\tau+1}-\bfa v_{\tau+1}\Vert_2 \leq\frac{\max_{i\in[\tau]}\delta_i+\sum_{i=1}^\tau\sqrt8\lambda_i\sqrt{\epsilon_i}}{\Delta}.
\end{align*}
Alternatively, we can also show that the eigenvector $\wha v_{\tau+1}$ returned by power method with $k= \frac{\log(1/\epsilon_0\delta)}{2\epsilon_0}$ that is initialized by  satisfies
\begin{align*}
    \wha v_{\tau+1}^\top\bfa v_{\tau+1} \geq 1-\epsilon_{\tau+1}:=  1-\frac{1}{2}\Big(\frac{\max_{i\in[\tau]}\delta_i+\sum_{i=1}^{\tau}\sqrt{8}\lambda_i\sqrt{\epsilon_i}}{\Delta}+\sqrt{2\epsilon_0}\Big)^2,
\end{align*}
\end{lemma}
    \begin{proof}
    Our proof is given by inductive arguments. Consider our obtained estimates $\{\wha v_i\}_{i\in[k]}$ for the eigenvectors $\{\bfa v_i\}_{i\in[k]}$ satisfy
    \begin{align*}
        \bfa v_i^\top\wha v_i\geq 1-\epsilon_i\qquad \forall i\in[\tau],\qquad|\lambda_i-\wh\lambda_i|\leq\delta_i.
    \end{align*}
    We note that for the eigenvectors, we have for a vector $\bfa v_0$,
    \begin{align*}
        \Vert\bfa v_i\bfa v_i^\top-\wha v_i\wha v_i^\top\Vert_2&=\sup_{\bfa v_0\in\bb S^{d-1}}\bfa v_0^\top(\bfa v_i\bfa v_i^\top-\wha v_i\wha v_i^\top)\bfa v_0=\sup_{\bfa v_0\in\bb S^{d-1}}(\bfa v_0^\top\bfa v_i)^2-(\bfa v_0^\top\wha v_i)^2\\
        &=\sup_{\bfa v_0\in\bb S^{d-1}}(\bfa v_0^\top(\bfa v_i-\wha v_i))(\bfa v_0^\top(\bfa v_i+\wha v_i^\top))\\
        &\leq 2\Vert\bfa v_i-\wha v_i\Vert_2=2\sqrt{\Vert\bfa v_i-\wha v_i\Vert_2^2} = 2\sqrt{\Vert\bfa v_i\Vert_2^2+\Vert\wha v_i\Vert_2^2-2\bfa v_i^\top\wha v_i} = 2\sqrt{2\epsilon_i}.
    \end{align*}
    Then, we can show by the subadditivity of the spectral norm,
    \begin{align*}
        \Big\Vert\bfa X\bfa X^\top-\sum_{i=1}^\tau\wh\lambda_i\wha v_i\wha v_i^\top\Big\Vert_2&=\Big\Vert\sum_{i=1}^k\lambda_i\bfa v_i\bfa v_i^\top-\sum_{i=1}^{\tau}\wh\lambda_i\wha v_i\wha v_i^\top\Big\Vert_2\\
        &\leq\Big\Vert\sum_{i=1}^k\lambda_i\bfa v_i\bfa v_i^\top-\sum_{i=1}^\tau\lambda_i\wha v_i\wha v_i^\top\Big\Vert_2+\Big\Vert\sum_{i=1}^{\tau}\delta_i\wha v_i\wha v_i^\top\Big\Vert_2\\
        &\leq\Big\Vert\sum_{i=\tau+1}^k\lambda_i\bfa v_i\bfa v_i^\top\Big\Vert_2+\Big\Vert\sum_{i=1}^\tau\delta_i\wha v_i\wha v_i^\top\Big\Vert_2+\Big\Vert\sum_{i=1}^{\tau}\lambda_i\bfa v_i\bfa v_i^\top-\sum_{i=1}^{\tau}\lambda_i\wha v_i\wha v_i^\top \Big\Vert_2\\
        &\leq\lambda_{\tau+1} + \max_{i\in[\tau]}\delta_i+ \Big\Vert\sum_{i=1}^{\tau}\lambda_i(\bfa v_i\bfa v_i^\top-\wha v_i\wha v_i^\top)\Big\Vert_2 \\
        &\leq\lambda_{\tau+1}+\max_{i\in[\tau]}\delta_i+\sum_{i=1}^{\tau}\lambda_i\Big\Vert\bfa v\bfa v_i^\top-\wha v_i\wha v_i^\top\Big\Vert_2\\
        &\leq\lambda_{\tau+1}+\max_{i\in[\tau]}\delta_i+\sum_{i=1}^{\tau}\sqrt{8}\lambda_i\sqrt{\epsilon_i}.
    \end{align*}
    By similar argument, we can also show that
    \begin{align*}
        \Big\Vert\bfa X\bfa X^\top-\sum_{i=1}^{\tau}\wh\lambda_i\wha v_i\wha v_i^\top\Big\Vert_2\geq\lambda_{\tau+1}-\max_{i\in[\tau]}\delta_i-\sum_{i=1}^{\tau}\sqrt 8\lambda_i\sqrt{\epsilon_i}.
    \end{align*}
    To study the convergence of the eigenvectors, we notice that by Davis-Kahan Theorem by \citep{yu2015useful} we can show that the principal eigenvector $\tda v_{\tau+1}=\argmax_{\bfa v\in\bb S^{d-1}}$ satisfies
    \begin{align*}
        \Vert\tda v_{\tau+1}-\bfa v_{\tau+1}\Vert_2\leq\frac{\Big|\lef\Vert\bfa X\bfa X^\top-\sum_{i}^{\tau}\wh\lambda_i\wha v_i\wha v_i^\top\rig\Vert-\lambda_{\tau+1}\Big|}{\max\{|\lambda_{\tau+1}-\lambda_{\tau}|,|\lambda_{\tau-1}-\lambda_{\tau}|\}}\leq\frac{\max_{i\in[\tau]}\delta_i+\sum_{i=1}^{\tau}\sqrt 8\lambda_i\sqrt{\epsilon_i}}{\Delta}.
    \end{align*}
    Consider the eigenvector returned by the power method, we can show by the subadditivity of $L_2$ norm, we obtain that $\Vert\wha v_{\tau+1}-\bfa v_{\tau+1}\Vert_2\leq \Vert\tda v_{\tau+1}-\wha v_{\tau+1}\Vert_2+\Vert\tda v_{\tau+1}-\bfa v_{\tau+1}\Vert_2\leq \frac{\max_{i\in[\tau]}\delta_i+\sum_{i=1}^{\tau}\sqrt{8}\lambda_i\sqrt{\epsilon_i}}{\Delta}+\sqrt{2\epsilon_0}$
    \begin{align*}
        \wha v_{\tau+1}^\top\bfa v_{\tau+1}&=\frac{1}{2}\lef(2-\Vert\bfa v_{\tau+1}-\wha v_{\tau+1}\Vert_2^2\rig)\geq\frac{1}{2}\lef(2-\lef(\Vert\tda v_{\tau+1}-\wha v_{\tau+1}\Vert_2+\Vert\bfa v_{\tau+1}-\tda v_{\tau+1}\Vert_2\rig)^2\rig)\\
        &=1-\frac{1}{2}\Big(\frac{\max_{i\in[\tau]}\delta_i+\sum_{i=1}^{\tau}\sqrt{8}\lambda_i\sqrt{\epsilon_i}}{\Delta}+\sqrt{2\epsilon_0}\Big)^2.
    \end{align*}
    Moreover, consider the estimate of the eigenvalue, we have 
    \begin{align*}
        &\Big\Vert\Big(\bfa X\bfa X^\top -\sum_{i=1}^{\tau}\wh\lambda_i\wha v_i\wha v_i^\top\Big)\wha v_{\tau+1}\Big\Vert_2\\
        &\leq\Big\Vert\Big(\bfa X\bfa X^\top-\sum_{i=1}^{\tau}\lambda_i\bfa v_i\bfa v_i^\top\Big)\wha v_{\tau+1}\Big\Vert_2+\Big\Vert\sum_{i=1}^\tau\lambda_i\bfa v_i\bfa v_i^\top-\sum_{i=1}^{\tau}\wh\lambda_i\wha v_i\wha v_i^\top \Big\Vert_2\\
        &\leq \Big\Vert\Big(\bfa X\bfa X^\top-\sum_{i=1}^{\tau}\lambda_i\bfa v_i\bfa v_i^\top\Big)\wha v_{\tau+1} \Big\Vert_2+\Big\Vert\sum_{i=1}^{\tau}\lambda_i\bfa v_i\bfa v_i^\top-\sum_{i=1}^{\tau}\lambda_i\wha v_i\wha v_i^\top\Big\Vert_2+\Big\Vert \sum_{i=1}^{\tau}\lef(\lambda_i-\wh\lambda_i\rig)\wha v_i\wha v_i^\top\Big\Vert_2\\
        &\leq\Big\Vert\Big(\bfa X\bfa X^\top-\sum_{i=1}^{\tau}\lambda_i\bfa v_i\bfa v_i^\top\Big)\bfa v_{\tau+1}\Big\Vert_2+\Big\Vert\Big(\bfa X\bfa X^\top-\sum_{i=1}^{\tau}\lambda_i\bfa v_i\bfa v_i^\top\Big)\Big\Vert_2\Vert\wha v_{\tau+1}-\bfa v_{\tau+1}\Vert_2\\
        &+\max_{i\in[\tau]}\delta_i+\sum_{i=1}^{\tau}\sqrt{8}\lambda_i\sqrt{\epsilon_i}\\
        &=\lambda_{\tau+1}+\lambda_{\tau+1}\Big(\frac{\max_{i\in[\tau]}\delta_i+\sum_{i=1}^{\tau}\sqrt 8\lambda_i\sqrt{\epsilon_i}}{\Delta}+\sqrt{2\epsilon_0}\Big)+\max_{i\in[\tau]}\delta_i+\sum_{i=1}^{\tau}\sqrt 8\lambda_i\sqrt{\epsilon_i}.
    \end{align*}
    Therefore, by similar arguments, we can show that
    \begin{align*}
        \Big|\Big\Vert \Big(\bfa X\bfa X^\top-\sum_{i=1}^{\tau}\wh\lambda_i\wha v_i\wha v_i^\top\Big)\wha v_{\tau+1}\Big\Vert_2-\lambda_{\tau+1}\Big|\leq \frac{2\lambda_{\tau+1}}{\Delta}\Big(\max_{i\in[\tau]}\delta_i+\sum_{i=1}^{\tau}\sqrt{8}\lambda_i\sqrt{\epsilon_i}\Big)+\lambda_{\tau+1}\sqrt{2\epsilon_0}.
    \end{align*}
\end{proof}

\subsection{Proof of Lemma \ref{reluapprox}}

    \begin{lemma}[Approximation of norm by sum of ReLU activations by Transformer networks]\label{reluapprox}
    Assume that there exists a constant $C$ with $\Vert\bfa v\Vert_2\leq C$. There exists a multihead ReLU attention layer with number of heads $M<\lef(\frac{\overline R}{\underline R}\rig)^d\frac{C(d)}{\epsilon^2}\log(1+C/\epsilon)$ such that there exsits $\{\bfa a_m\}_{m\in[M]}\subset\bb S^{N-1}$ and $\{c_m\}_{m\in[M]}\subset\bb R$ where for all $\bfa v$ with $\overline R\geq\Vert\bfa v\Vert_2\geq \underline R$, we have 
    \begin{align*}
        \bigg|\sum_{m=1}^Mc_m\sigma(\bfa a_m^\top\bfa v)-\frac{1}{\Vert\bfa v\Vert_2}+1\bigg|\leq\epsilon.
    \end{align*}
    Similarly, there exists a multihead ReLU attention layer with number of heads $M\leq\overline R^{\frac{d}{2}}\frac{C(d)}{\epsilon^2}\log\lef(1+C/\epsilon\rig)$, a set of vectors $\{\bfa b_m\}_{m\in[M]}\subset\bb S^{N-1}$ and $\{d_m\}_{m\in[M]}\subset\bb R$ such that 
    \begin{align*}
        \Big|\sum_{m=1}^Md_m\sigma(\bfa b_m^\top\bfa v)-\Vert\bfa v\Vert_2^{1/2}+1\Big|\leq\epsilon.
    \end{align*}
\end{lemma}

\begin{proof}
    Consider a set $\msf C^d(\overline R) :=\msf B^d_{\infty}(\overline R)\setminus\msf B_{2}^d(\underbar R)$, then it is not hard to check that given $\Vert\bfa v\Vert_2>C$ with some $C(d)>0$ depending on $d$ such that we have
    \begin{align*}
        \sup_{\bfa v\in\msf C^d(\overline R)}\pta_{v_{j_1},\ldots,v_{j_i}\in[d]}\bl\frac{1}{\Vert\bfa v\Vert_2}\br\leq\frac{C(d)}{\Vert\bfa v\Vert_2^d}\leq\frac{C(d)}{\underline R^d}.
    \end{align*}
    Therefore, consider the definition \ref{def6}, we have $C_{\ell}=\lef(\frac{\overline R}{\underline R}\rig)^d C(d)$.  Note that by proposition A.1 in \citep{bai2024transformers} shows that for a function that is $(R,C_{\ell})$ smooth with $R\geq 1$ is $(\epsilon_{approx}, R, M,C)$ approximable with $M\leq C(d)C_{\ell}\log(1+C_{\ell}/\epsilon_{approx})/\epsilon_{approx}^2$, we complete the proof. 

    Then we consider the function $\Vert\bfa v\Vert_2^{\frac{1}{2}}$, note that
    \begin{align*}
        \sup_{\bfa v\in\msf C^d(\overline R)}\pta_{v_{j_1},\ldots,v_{j_i}\in[d]}\Vert\bfa v\Vert_2^{\frac{1}{2}}\leq C\Vert\bfa v\Vert_2^{-\frac{1}{2}}\leq C\overline R^{-\frac{1}{2}}.
    \end{align*}
    And the rest of the proof follows similarly to the previous step.
\end{proof}
\subsection{Proof of Thereom \ref{thm411}}
    Our algorithm goes by first splitting the data into two parts $\{\bfa X_i\}_{i\in[N_1]}$ and $\{\bfa X_i\}_{i\in\{N_1+1,\ldots, N\}}$. And using the first part we construct an estimate for the principal direction of variance. Then we cluster the rest of the samples through projecting them to the principal direction. 
    
    We apply the matrix perturbation theory to show that the empirical covariance matrix is close to the expected one. And the principal component corresponds to the direction with the large variance. We note that the covariance matrix of the mixture model is given by 
    \begin{align*}
        \bb E&\lef[(\bfa X_1-\bb E[\bfa X_1])(\bfa X_1-\bb E[\bfa X_1])^\top\rig]=\bb E[\bfa X_1\bfa X_1^\top]-\bb E[\bfa X_1]\bb E[\bfa X_1]^\top\\
        &=\frac{1}{2}\bb E[\bfa X_1\bfa X_1^\top|z_1=1]+\frac{1}{2}\bb E[\bfa X_1X^\top_1|z_1=0]-\frac{(\bfa\mu_1+\bfa\mu_2)(\bfa\mu_1+\bfa\mu_2)^\top}{4}\\
        &=I_d+ \frac{1}{2}(\bfa\mu_1\bfa\mu_1^\top+\bfa\mu_2\bfa\mu_2^\top)-\frac{(\bfa\mu_1+\bfa\mu_2)(\bfa\mu_1+\bfa\mu_2)^\top}{4}\\
        &=I_d+\frac{1}{4}(\bfa\mu_1\bfa\mu_1^\top+\bfa\mu_2\bfa\mu_2^\top)-\frac{1}{4}\bfa\mu_1\bfa\mu_2^\top-\frac{1}{4}\bfa\mu_2\bfa\mu_1^\top\\
        &= I_d+\frac{1}{4}(\bfa\mu_1-\bfa\mu_2)(\bfa\mu_1-\bfa\mu_2)^\top.
    \end{align*}
    Therefore, it is noted that the top-$1$ principal eigenvector is given by $\frac{\bfa\mu_1-\bfa\mu_2}{\Vert\bfa \mu_1-\bfa\mu_2\Vert_2}$, corresponding to the eigenvalue of $1+\frac{1}{4}\Vert\bfa\mu_1-\bfa\mu_2\Vert_2^2$. We further consider the empirical covariance matrix, given by 
    \begin{align*}
        \frac{1}{{N_1}}\sum_{i=1}^{N_1}(\bfa X_i-\bar{\boldsymbol{X}})(\bfa X_i-\bar{\boldsymbol{X}})^\top=\frac{1}{N_1}\sum_{i=1}^{N_1}\bfa X_i\bfa X_i^\top-\bar{\boldsymbol{X}}\bar{\boldsymbol{X}}^\top.
    \end{align*}
    To understand the statistical behavior of $\frac{1}{{N_1}-1}\sum_{i=1}^{N_1}\bfa X_i\bfa X_i^\top-\bar{\boldsymbol{X}}\bar{\boldsymbol{X}}^\top$, we first study the sub-Gaussian property of $\la v,\bfa X_1\ra$ for any $v$. Using the fact that
    $\exp(x)+\exp(-x) \leq 2\exp\lef(\frac{1}{8}x^2\rig)$, we can show that for all $v\in\bb S^{d-1}$ we have
    \begin{align*}
        &\bb E\lef[\exp(\lambda\la v,\bfa X_1\ra )\rig]= \frac{1}{2}\Big(\bb E\lef[\exp(\lambda\la v, \bfa X_1\ra)|z_1 = 0\rig]+\bb E\lef[\exp(\lambda\la v, \bfa X_1\ra)|z_1 = 1\rig]\Big)\\
        &=\frac{1}{2}\Big(\exp\Big(\lambda v^\top\bfa\mu_0+\frac{1}{2}\lambda^2\Big)+\exp\Big(\lambda v^\top\bfa\mu_1+\frac{1}{2}\lambda^2\Big) \Big)\\
        &=\frac{1}{2}\exp\Big(\frac{1}{2}\lambda^2\Big)\lef(\exp(\lambda v^\top\bfa\mu_1)+\exp(\lambda v^\top\bfa\mu_0)\rig)\\
        &=\frac{1}{2}\exp\Big(\frac{1}{2}\lambda^2+\frac{1}{2}\lambda v^\top(\bfa\mu_1+\bfa\mu_0)\Big)\lef(\exp\Big(\frac{1}{2}\lambda v^\top(\bfa\mu_1-\bfa\mu_0)\Big)+\exp\Big(-\frac{1}{2}\lambda v^\top(\bfa\mu_1-\bfa\mu_0)\Big)\rig)\\
        &\leq\exp\Big(\frac{1}{2}\lambda^2+\frac{1}{2}\lambda v^\top(\bfa\mu_1+\bfa\mu_0)+\frac{1}{8}\lambda^2\Vert\bfa\mu_1-\bfa\mu_0\Vert_2^2\Big).
    \end{align*}
    Hence, we can additionally show that $\bfa X_1-\bb E[\bfa X_1]$ is sub-Gaussian with $\sigma^2 = 1+\frac{1}{4}\Vert\bfa\mu_1-\bfa\mu_0\Vert_2^2$. Then, defining $\wh\Sigma = (X-\bb E[\bfa X])(X-\bb E[\bfa X])^\top$ and $\Sigma = \bb E[\bfa X_1\bfa X_1^\top]-\bb E[\bfa X_1]\bb E[\bfa X_1]^\top$, we can show that
    \begin{align}\label{diffwhsigmasigma}
        \bb P\bl\frac{\Vert\wh\Sigma-\Sigma\Vert_2}{1+\frac{1}{4}\Vert\bfa\mu_1-\bfa\mu_0\Vert_2^2}\geq C\Big(\sqrt{\frac{d}{{N_1}}}+\frac{d}{{N_1}}+\delta\Big)\br\leq\exp(-C{N_1}\min(\delta,\delta^2)).
    \end{align}
    Denote $\tde\Sigma:=\frac{1}{{N_1}}\sum_{i=1}^{N_1}(\bfa X_i-\bar{\boldsymbol{X}})(\bfa X_i-\bar{\boldsymbol{X}})$, we consider the difference between $\tde\Sigma$ and $\wh\Sigma$, it is noted that
    \begin{align}\label{diffsigmatde}
        \lef\Vert\tde\Sigma-\wh\Sigma\rig\Vert_2&= \lef\Vert\frac{1}{{N_1}}\sum_{i=1}^{N_1}(\bfa X_i-\bar{\boldsymbol{X}})(\bfa X_i-\bar{\boldsymbol{X}})^\top-\frac{1}{{N_1}}\sum_{i=1}^{N_1}(\bfa X_i-\bb E[\bfa X])(\bfa X_i-\bb E[\bfa X])^\top\rig\Vert_2\nnb\\
        &=\lef\Vert 2\bar{\boldsymbol{X}}\bb E[\bfa X]^\top-\bar{\boldsymbol{X}}\bar{\boldsymbol{X}}^\top-\bb E[\bfa X]\bb E[\bfa X]^\top\rig\Vert_2=\lef\Vert(\bar{\boldsymbol{X}}-\bb E[\bfa X])(\bar{\boldsymbol{X}}-\bb E[\bfa X])^\top\rig\Vert_2\nnb\\
        &=\Vert\bar{\boldsymbol{X}}-\bb E[\bfa X]\Vert_2^2.
    \end{align}
    To analyze the difference between $\bar{\boldsymbol{X}}$ and $\bb E[\bfa X]$ we introduce the projection $P:=\frac{(\bfa\mu_0-\bfa\mu_1)(\bfa\mu_0-\bfa\mu_1)^\top}{\Vert\bfa\mu_0-\bfa\mu_1\Vert_2^2}$ and $P^\perp = I-P$.
    We further notice that $P(\bar{\boldsymbol{X}}-\bb E[\bfa X]) = \frac{1}{{N_1}}\sum_{i=1}^{N_1}P\bfa X_i-\bb E[P\bfa X_i]$ where $P\bfa X_i$s are i.i.d. univariate sub-Gaussian random variables with  $\sigma^2 = 1+\frac{1}{4}\Vert\bfa\mu_1-\bfa\mu_0\Vert_2^2$. We therefore utilize Hoeffding's inequality to obtain that for all $t>0$,
    \begin{align*}
        \bb P\bl(P\bar{\boldsymbol{X}}-\bb E[P\bfa X])^2\geq t^2\br =\bb P\bl|P\bar{\boldsymbol{X}}-\bb E[P\bfa X]|\geq t\br\leq 2\exp\bl-\frac{{N_1}t^2}{2\lef(1+\frac{1}{4}\Vert\bfa\mu_1-\bfa\mu_0\Vert_2^2\rig)}\br.
    \end{align*}
    And, it is noted that the distribution of $P^\perp(\bar{\boldsymbol{X}}-\bb E[\bfa X])$ is ${N_1}\lef(0, \frac{1}{{N_1}}I_{d-1}\rig)$. Utilizing the $\chi^2(d-1)$ tail bound given by \citep{laurent2000adaptive},
    \begin{align*}
        \bb P\Big(\lef\Vert P^\perp \bar{\boldsymbol{X}}\rig\Vert_2^2\geq  \frac{1}{{N_1}}(d-1+\sqrt{2(d-1)x}+2x)\Big)\leq\exp(-x).
    \end{align*}
    Then, using the union bound, we can check that for $a,b>0$, define $\ca E_{a,b} =\{(P\bar{\boldsymbol{X}}-\bb E[P\bfa X])^2\leq a, \Vert P^\perp\bar{\boldsymbol{X}}\Vert_2^2\leq b\}$, then we can show that
    \begin{align*}
        \bb P\lef(\ca E_{a,b}\rig)
        &\geq 1-\bb P\Big((P\bar{\boldsymbol{X}}-\bb E[P\bfa X])^2>a\text{ or } \Vert P^\perp\bar{\boldsymbol{X}}\Vert_2^2>\frac{d-1}{{N_1}}+b\Big)\\
        &\geq 1-\bb P\Big((P\bar{\boldsymbol{X}}-\bb E[P\bfa X])^2> a\Big)-\bb P\Big(\Vert P^\perp \bar{\boldsymbol{X}}\Vert_2^2 > \frac{d-1}{{N_1}}+b\Big)\\
        &\geq 1 - 2\exp\Big(-\frac{C{N_1}a}{1+\frac{1}{4}\Vert\bfa\mu_1-\bfa\mu_0\Vert_2^2}\Big) -\exp\Big( C\min\Big\{\frac{{N_1}^2b^2}{d-1}, {N_1}b\Big\}\Big).
    \end{align*}
    Let $a = \frac{1+\frac{1}{4}\Vert\bfa\mu_1-\bfa\mu_0\Vert_2^2}{{N_1}}$ and $b = \sqrt{\frac{d-1}{N_1}}$. Then, with probability at least $1-C\delta$, we have
    \begin{align}\label{diffxex}
        \Vert \bar{\boldsymbol{X}}-\bb E[\bfa X]\Vert_2^2&= \Vert P(\bar{\boldsymbol{X}}-\bb E[\bfa X])\Vert_2^2 + \Vert P^\perp(\bar{\boldsymbol{X}}-\bb E[\bfa X])\Vert_2^2\nnb\\
        &\leq \frac{1+\frac{1}{4}\Vert\bfa\mu_1-\bfa\mu_0\Vert_2^2}{{N_1}}\log(1/\delta)+\frac{d-1}{{N_1}}+\frac{\log(1/\delta)}{{N_1}}\vee \sqrt{\frac{(d-1)\log(1/\delta)}{N_1}}.
    \end{align}
    Then, collecting \eqref{diffwhsigmasigma}, \eqref{diffsigmatde}, and \eqref{diffxex} we can show that with probability at least $1-C\delta$,
    \begin{align*}
        \Vert\tde\Sigma-\Sigma\Vert_2&\leq\Vert\tde\Sigma-\wh\Sigma\Vert_2+\Vert\wh\Sigma-\Sigma\Vert_2\leq \frac{1+\frac{1}{4}\Vert\bfa\mu_1-\bfa\mu_0\Vert_2^2}{{N_1}}\log(1/\delta)\\
        &+\frac{d-1}{{N_1}}+\frac{\log(1/\delta)}{{N_1}}\vee\sqrt{\frac{(d-1)\log(1/\delta)}{{N_1}}}+\sqrt{\frac{d}{{N_1}}}\Big(1+\frac{1}{4}\Vert\bfa\mu_1-\bfa\mu_0\Vert_2^2\Big)\\
        &\leq \frac{C\log(1/\delta)}{{N_1}}\vee\sqrt{\frac{(d-1)\log(1/\delta)}{{N_1}}}+\sqrt{\frac{d}{{N_1}}}\Big(1+\frac{1}{4}\Vert\bfa\mu_1-\bfa\mu_0\Vert_2^2\Big).
    \end{align*}
    Denote the principal eigenvector of $\Sigma$ as $\bfa v_1$ and the principal eigenvector of $\wh\Sigma$ as $\wha v_1$, then we make use the handy version of the Davis-Kahan theorem provided by \citep{yu2015useful} to obtain that with probability at least $1-\delta$, we have
    \begin{align}\label{dkthms}
        \sqrt{1-(\bfa v_1^\top\wha v_1)^2}&\leq \frac{2\Vert\wh\Sigma-\tde\Sigma\Vert_{2}}{\Vert\bfa\mu_1-\bfa\mu_0\Vert_2^2}\leq C\sqrt{\frac{d(1+\log(1/\delta))}{{N_1}}}\Big(\Vert\bfa\mu_1-\bfa\mu_0\Vert_2^{-2}+\frac{1}{4}\Big)+\frac{1}{\Vert\bfa\mu_1-\bfa\mu_0\Vert_2^2}\frac{C}{{N_1}}.
    \end{align}
    Analogously we can show that with probability at least $1-\delta$, we have
    \begin{align*}
        \lef\Vert\wha v_1-\bfa v_1\rig\Vert_2 &= \sqrt{\Vert\wha v_1\Vert_2^2 +\Vert\bfa v_1\Vert_2^2 -2\bfa v_1^\top\wha v_1}=2\sqrt{1-\bfa v_1^\top\wha v_1}\\
        &\leq C\sqrt{\frac{d(1+\log(1/\delta))}{{N_1}}}\Big(\Vert\bfa\mu_1-\bfa\mu_0\Vert_2^{-2}+\frac{1}{4}\Big)+\frac{C}{\Vert\bfa\mu_1-\bfa\mu_0\Vert_2^2{N_1}}.
    \end{align*}
    And we denote the above event as $\ca E$.
    Then we analyze the approximate mean estimator given by $\bar{\boldsymbol{X}}^\top\wha v_1$, note that by Cauchy-Schwartz inequality we obtain that
    \begin{align*}
        \Big| \bar{\boldsymbol{X}}^\top\wha v_1&-\frac{(\bfa\mu_0+\bfa\mu_1)^\top\bfa v_1}{2}\Big|=\bar{\boldsymbol{X}}^\top\wha v_1-\bar{\boldsymbol{X}}^\top\bfa v_1+\bar{\boldsymbol{X}}^\top\bfa v_1-\frac{(\bfa\mu_0+\bfa\mu_1)^\top\bfa v_1}{2}\\
        &=\bar{\boldsymbol{X}}^\top(\wha v_1-\bfa v_1)+\Big(\bar{\boldsymbol{X}}-\frac{1}{2}(\bfa\mu_0+\bfa\mu_1) \Big)^\top\bfa v_1\leq\lef\Vert\bar{\boldsymbol{X}}\rig\Vert_2\Vert\wha v_1-\bfa v_1\Vert_2+(\bar{\boldsymbol{X}}-\bb E[\bfa X])^\top\bfa v_1.
    \end{align*}
    Consider the $L_2$ norm of $\bar{\boldsymbol{X}}$, it is noted that $\bar{\boldsymbol{X}}$'s two norm under $\ca E_{a,b}$ with $a=\frac{1+\frac{1}{4}\Vert\bfa\mu_1-\bfa\mu_0\Vert_2^2}{{N_1}}$ and $b = \frac{\sqrt{d-1}}{{N_1}}$ satisfies the following with probability at least $1-C \delta$,
    \begin{align*}
        \Vert\bar{\boldsymbol{X}}\Vert_2&\leq\Vert\bar{\boldsymbol{X}}-\bb E[\bfa X]\Vert_2 +\Vert\bb E[\bfa X]\Vert_2\leq\frac{\Vert\bfa\mu_1-\bfa\mu_0\Vert_2}{2} + \sqrt{\frac{1+\frac{1}{4}\Vert\bfa\mu_0-\bfa\mu_1\Vert_2^2}{{N_1}}\log(1/\delta)}\\
        &+\sqrt{\frac{d-1}{{N_1}}} +\sqrt{\frac{\log(1/\delta)}{{N_1}}}\vee\sqrt{\frac{(d-1)\log(1/\delta)}{N_1}}.
    \end{align*}
    For the term $T_2$, we can show that $(\bar{\boldsymbol{X}}-\bb E[\bfa X])^\top\bfa v_1$ is a $\frac{1}{4}\Vert\bfa\mu_0-\bfa\mu_1\Vert_2^2+1$ sub-Gaussian random variable. By Hoeffding's inequality, we can show that
    \begin{align*}
     \bb P\Big(\lef|(\bar{\boldsymbol{X}}-\bb E[\bfa X])^\top\bfa v_1\rig|\geq t\Big)&\leq 2\exp\lef(-\frac{{N_1}t^2}{2(\frac{1}{4}\Vert\bfa\mu_0-\bfa\mu_1\Vert_2^2+1)}\rig).
    \end{align*}
    Hence, with probability at least $1-C\delta$, we have
    \begin{align*}
        \lef|(\bar{\boldsymbol{X}}-\bb E[\bfa X])^\top\bfa v_1\rig|\leq\sqrt{\frac{1}{{N_1}}\Big(\frac{1}{2}\Vert\bfa\mu_0-\bfa\mu_1\Vert_2^2+2\Big)\log(1/\delta)}.
    \end{align*}
    Therefore, we can show that with probability at least $1-C\delta$, as ${N_1}\gtrsim d$ we can show that
    \begin{align*}
        \Big|\bar{\boldsymbol{X}}^\top\wha v_1&-\bb E[\bfa X]^\top\bfa v_1\Big|\leq \bl\frac{1}{2}\Vert\bfa\mu_1-\bfa\mu_0\Vert_2+\sqrt{\frac{1+\frac{1}{4}\Vert\bfa\mu_0-\bfa\mu_1\Vert_2^2}{N}\log(1/\delta)}\\
        &+\sqrt{\frac{d-1}{{N_1}}}+\sqrt{\frac{\log(1/\delta)}{{N_1}}}\vee \sqrt{\frac{(d-1)\log(1/\delta)}{{N_1}}}\br\\
        &\cdot\bl C\sqrt{\frac{d}{{N_1}}}\Big(\Vert\bfa\mu_1-\bfa\mu_0\Vert_2^{-2}+4\Big)+\Vert\bfa\mu_1-\bfa\mu_0\Vert_2^{-2}\frac{C}{{N_1}}\br\\
        &+\sqrt{\frac{1}{{N_1}}\Big(\frac{1}{2}\Vert\bfa\mu_0-\bfa\mu_1\Vert_2^2+2\Big)\log(1/\delta)}\\
        &\lesssim \sqrt{\frac{1}{{N_1}}(\Vert\bfa\mu_0-\bfa\mu_1\Vert_2^2+4)\log(1/\delta)}+\sqrt{\frac{d}{{N_1}}}\lef(\Vert\bfa\mu_1-\bfa\mu_0\Vert_2^{-1}+4\Vert\bfa\mu_1-\bfa\mu_0\Vert_2\rig)\\
        &+\sqrt{\frac{d-1}{{N_1}}}\log(1/\delta).
    \end{align*}
    We consider the oracle estimator given by the smoothed sign $\tde F(\bfa X_i):=\tanh(\beta(\bfa X_i-\bb E[\bfa X])^\top\bfa v_1)$. We also define $\wh F(\bfa X_i):=\tanh(\beta(\bfa X_i-\bar{\boldsymbol{X}})^\top\wha v_1)$ as our approximate estimator. Then we can show that, given $\wh z_i:= \wh F(\bfa X_i)$,
    \begin{align*}
        \bb E\bigg[\min_{\tde z\in\{z,-z\}}\frac{1}{N}\sum_{i=N_1+1}^{N}L(\wh z_i,z_i)\bigg]&= \bb E\bigg[\min_{z_0\in\{\wh z,-\wh z\}}\frac{1}{N}\sum_{i=1}^N L(z_{0,i},z_i)\bigg]\leq \bb E\lef[L(\wh F(\bfa X_i),z_i)\rig]\\
        &\leq\ub{\bb E\lef[L(\tde F(\bfa X_i),z_i)\rig]}_{T_1}+\ub{\bb E\Big[ \lef|\wh F(\bfa X_i)-\tde F(\bfa X_i)\rig|\Big]}_{T_2}.
    \end{align*}
    To analyze the loss, we first consider the approximation error of the $\tanh(\beta x)$ function with respect to the $\sign$ function. Note that
    \begin{align*}
        |\tanh(\beta x)&-\sign(x)|= \mbbm 1_{x>0}\Big|\frac{1-\exp(-2\beta x)}{1+\exp(-2\beta x)}-1\Big| + \mbbm 1_{x\leq 0}\Big|\frac{-1+\exp(2\beta x)}{1+\exp(2\beta x)}+1\Big|\\
        &=2\mbbm 1_{x>0}\exp(-2\beta x)+ 2\mbbm 1_{x\leq 0}\exp(2\beta x)+ O(\exp(-4\beta|x|))\\
        &=2\exp(-2\beta|x|)+O(\exp(-4\beta|x|)).
    \end{align*}
    For the first term, we have
    \begin{align}\label{t1upperbound}
        T_1 &= \bb E\Big[L(\tde F(\bfa X_i), z_i)\Big]\leq\bb E\Big[\Big|\tanh(\beta(\bfa X_i-\bb E[\bfa X])^\top\bfa v_1)-\sign(\beta(\bfa X_i-\bb E[\bfa X])^\top\bfa v_1)\Big|\Big]\nnb\\
        &+\bb E\Big[\Big|\sign((\bfa X_i-\bb E[\bfa X])^\top\bfa v_1) - z_i\Big|\Big]\nnb\\
        &\leq 2\bb E\Big[\exp\Big(-2\beta|(\bfa X_i-\bb E[\bfa X])^\top\bfa v_1|\Big)\Big]+O\Big(\bb E\Big[\exp(-4\beta|(\bfa X_i-\bb E[\bfa X])^\top\bfa v_1|)\Big]\Big)\nnb\\
        &+2\bb P\lef((\bfa X_i-\bb E[\bfa X])^\top\bfa v_1>0|z_i=-1\rig)+2\bb P((\bfa X_i-\bb E[\bfa X])^\top\bfa v_1<0|z_i=1\nnb\\
        &\leq C\exp\lef(-C\beta^2\Vert\bfa\mu_0-\bfa\mu_1\Vert_2^2\rig) + \frac{C}{\Vert\bfa\mu_0-\bfa\mu_1\Vert_2}\exp\Big(-\frac{1}{4}\Vert\bfa\mu_0-\bfa\mu_1\Vert_2^2\Big).
    \end{align}
    Then, we consider the second term to obtain that there exists an event $\ca E$ with $\bb P(\ca E)\geq 1-\delta$, such that
    \begin{align}\label{diffff}
        T_2&=\bb E\Big[\lef|\wh F(\bfa X_i)-\tde F(\bfa X_i)\rig|\ca E\Big]\bb P(\ca E)+\bb E\Big[\lef|\wh F(\bfa X_i)-\tde F(\bfa X_i)\rig|\ca E^c\Big]\bb P(\ca E^c)\nnb\\
        &\leq \bb E\Big[\Big|\tanh(\beta(\bfa X_i-\bb E[\bfa X])^\top\bfa v_1)-\tanh(\beta(\bfa X_i-\bar{\boldsymbol{X}})^\top\wha v_1)\Big|\Big|\ca E\Big]\bb P(\ca E)+C\delta\nnb\\
        &=\bb E\Big[(1-\tanh^2(X-\bb E[\bfa X])^\top\bfa v_1)\beta\Big| (X-\bb E[\bfa X])^\top\bfa v_1-(X-\bar{\boldsymbol{X}})^\top\wha v_1\Big|\Big|\ca E\Big]\bb P(\ca E)+C\delta\nnb\\
        &\leq\bb E\Big[(1-\tanh^2(X-\bb E[\bfa X])^\top\bfa v_1)\beta\Big(| X^\top(\bfa v_1-\wha v_1)|+\Big|\bb E[\bfa X]^\top\bfa v_1-\bar{\boldsymbol{X}}^\top\wha v_1\Big|\Big)\Big|\ca E\Big]\bb P(\ca E)+C\delta\nnb\\
        &\leq \beta\Big(C\sqrt{\frac{1}{{N_1}}(\Vert\bfa\mu_0-\bfa\mu_1\Vert_2^2+4)\log(1/\delta)}+\sqrt{\frac{d}{{N_1}}}\lef(\Vert\bfa\mu_1-\bfa\mu_0\Vert_2^{-1}+4\Vert\bfa\mu_1-\bfa\mu_0\Vert_2\rig)\nnb\\
        &+\sqrt{\frac{d-1}{{N_1}}}\log(1/\delta) \Big)+\beta\bb E[(1-\tanh^2(\beta(X-\bb E[\bfa X])^\top\bfa v_1))X^\top(\bfa v_1-\wha v_1)]+C\delta.
    \end{align}
    For the last term we can show that, if we denote $\bfa a:=\bfa v_1-\wha v_1$, then we note that
    \sm{\begin{align*}
       &\bb E\lef[(1-\tanh^2(\beta(X-\bb E[\bfa X])^\top\bfa v_1))|X^\top(\bfa v_1-\wha v_1)|\Big|\ca E\rig]
       = \bb E\lef[(1-\tanh^2(\beta(X-\bb E[\bfa X])^\top\bfa v_1))|X^\top\bfa a|\Big|\ca E\rig]\\
       &=\bb E\lef[(1-\tanh^2(\beta(X-\bb E[\bfa X])^\top\bfa v_1))|X^\top(P\bfa a+P^\perp \bfa a)|\Big|\ca E \rig]\\
       &=\bb E\lef[(1-\tanh^2(\beta(X-\bb E[\bfa X])^\top\bfa v_1))|X^\top P\bfa a|\Big|\ca E\rig]+\bb E\lef[(1-\tanh^2(\beta(X-\bb E[\bfa X])^\top\bfa v_1))|X^\top P^\perp \bfa a|\Big|\ca E\rig]\\
       &\lesssim\bb E\lef[(1-\tanh^2(\beta(X-\bb E[\bfa X])^\top\bfa v_1))|X^\top P\bfa a|\Big|\ca E\rig]+\sqrt d\bb E\lef[\Vert\bfa a\Vert_2|\ca E\rig]\\
       &\lesssim\bb E\lef[C\beta\exp(-C\beta(X-\bb E[\bfa X])^\top\bfa v_1)|X^\top\bfa v_1|\rig]+\sqrt d\bb E\lef[\Vert\bfa a\Vert_2|\ca E\rig]\\
       &\lesssim\bb E\lef[C\beta\exp(-C\beta(X-\bb E[\bfa X])^\top\bfa v_1)|(X-\bb E[\bfa X])^\top\bfa v_1|\rig]\\
       &+C\beta(\bfa\mu_0-\bfa\mu_1)^\top\bfa v_1\exp(-\beta^2\Vert\bfa\mu_0-\bfa\mu_1\Vert_2^2)\\
       &\lesssim\sqrt d\bb E\lef[C\beta\exp(-C\beta(X-\bb E[\bfa X])^\top\bfa v_1)|(X-\bb E[\bfa X])^\top\bfa v_1|\rig]+C\sqrt{d}\beta\Vert\bfa\mu_0-\bfa\mu_1\Vert_2^2\exp(-\beta^2\Vert\bfa\mu_0-\bfa\mu_1\Vert_2^2)\\
       &\lesssim C\beta\Vert\bfa\mu_0-\bfa\mu_1\Vert_2^2\exp(-\beta^2\Vert\bfa\mu_0-\bfa\mu_1\Vert_2^2).
    \end{align*}}
    Therefore, we conclude that there exsits an event $\ca E$ with $\bb P(\ca E)\geq 1-C\delta$ such that
    \begin{align*}
        &\bb E\Big[\min_{\tde z\in\{z,-z\}}\frac{1}{N}\sum_{i=N_1+1}^NL(\wh z_i,z_i)\Big|\ca E\Big]\lesssim \lef(1+\beta^2 \Vert\bfa\mu_0-\bfa\mu_1\Vert_2^2\rig)\exp\lef(-C\beta^2\Vert\bfa\mu_0-\bfa\mu_1\Vert_2^2\rig)\\
        &+ \frac{2}{\Vert\bfa\mu_0-\bfa\mu_1\Vert_2}\exp\Big(-\frac{1}{4}\Vert\bfa\mu_0-\bfa\mu_1\Vert_2^2\Big)+\sqrt{\frac{d}{N_1}}\lef(\Vert\bfa\mu_1-\bfa\mu_0\Vert_2^{-1}+4\Vert\bfa\mu_1-\bfa\mu_0\Vert_2+\log(1/\delta)\rig).
    \end{align*}
    Then we finally conclude that
    \begin{align}\label{lossub}
    \bb E\Big[\min_{\tde z\in\{z,-z\}}&\frac{1}{N}\sum_{i=1}^NL(\wh z_i,z_i)\Big]\leq\bb E\Big[\min_{\tde z\in\{z,-z\}}\frac{1}{N}\sum_{i=N_1+1}^{N}L(\wh z_i,z_i)\Big]+\frac{CN_1}{N}\nnb\\
    &\lesssim \bb E\Big[\min_{\tde z\in\{z,-z\}}\frac{1}{N}\sum_{i=N_1+1}^N L(\wh z_i,z_i)\Big|\ca E\Big]\bb P(\ca E)+\bb P(\ca E^c)+\frac{CN_1}{N}\nnb\\
    &\lesssim\frac{N_1}{N}+\lef(1+\beta^2 \Vert\bfa\mu_0-\bfa\mu_1\Vert_2^2\rig)\exp\lef(-C\beta^2\Vert\bfa\mu_0-\bfa\mu_1\Vert_2^2\rig)+\delta\nnb\\
    &+\sqrt{\frac{d}{N_1}}\lef(\Vert\bfa\mu_1-\bfa\mu_0\Vert_2+\log(1/\delta)\rig)+\frac{1}{\Vert\bfa\mu_0-\bfa\mu_1\Vert_2}\exp\Big(-\frac{1}{4}\Vert\bfa\mu_0-\bfa\mu_1\Vert_2^2\Big).
    \end{align}
    Optimizing over $N_1$ and $\delta$, we let $N_1=d^{\frac{1}{3}}N^{\frac{2}{3}}\lef(\Vert\bfa\mu_1-\bfa\mu_0\Vert_2+\log N\rig)^{\frac{2}{3}}$ and $\delta = N_1/N$, then the above expectation reduces to 
    \begin{align*}
        \bb E[L_{GMM}(\wh z, z)]&\lesssim d^{\frac{1}{3}}N^{-\frac{1}{3}}\lef(\Vert\bfa\mu_1-\bfa\mu_0\Vert_2+\log N\rig)^{\frac{2}{3}}+\Vert\bfa\mu_0-\bfa\mu_1\Vert_2^{-1}\exp\Big(-\frac{1}{4}\Vert\bfa\mu_0-\bfa\mu_1\Vert_2^2\Big)\\
        &+(1+\beta^2\Vert\bfa\mu_0-\bfa\mu_1\Vert_2^2)\exp\lef(-C\beta^2\Vert\bfa\mu_0-\bfa\mu_1\Vert_2^2\rig).
    \end{align*}
    Hence, given that $\Vert\bfa\mu_0-\bfa\mu_1\Vert_2\gtrsim \sqrt{\log (N/d)}\vee\sqrt{\log(d)}$, we can show that algorithm \ref{alg:gmm} achieves
    \begin{align*}
        \bb E[L_{GMM}(\wh z,z)]\lesssim \lef(\frac{d\log^2 N}{N}\rig)^{\frac{1}{3}}.
    \end{align*}    \label{proof:gmm}


\subsection{Proof of Lemma \ref{lm:4.1}}
        We first consider the covariate matrix built in the proof of theorem \ref{thm3.1}, noticing that
        \begin{align*}
            &\bfa V_{1}^{cov}= I_D = -\bfa V_{2}^{cov} ,\quad \bfa Q_1^{cov}=-\bfa Q_2^{cov} =\begin{bmatrix}
                \bfa 0_{N_1\times(2d)}&I_{N_1}&\bfa 0\\
                &\bfa 0&
            \end{bmatrix}, \\
            &\bfa K_1^{cov}=\bfa K_2^{cov}=\begin{bmatrix}
                \bfa a_1&\ldots&\bfa a_{N_1}&\bfa 0_{D\times(D-N_1)}\end{bmatrix}\times\begin{bmatrix}
                \bfa 0_{N\times(2d)}&I_{N_1}&\bfa 0\\
                &\bfa 0&
            \end{bmatrix}.
        \end{align*}
        Given the above construction, we can show that
        \begin{align*}
            (\bfa Q_1^{cov}\bfa H)^\top\bfa K_1^{cov}\bfa H = \begin{bmatrix}
                I_{N_1}&\bfa 0_{N_1\times(N-N_1)}\\
                \bfa 0_{(N-N_1)\times N_1}&\bfa 0
            \end{bmatrix}\times\begin{bmatrix}
                \bfa a_1&\ldots&\bfa a_{N_1}&\bfa 0_{D\times (N-N_1)}
            \end{bmatrix}
        \end{align*}
        Hence, it is subsequently shown that
        \begin{align*}
            &\tda H^{cov, 1} = \bfa H+\sum_{i=1}^4\bfa V_i^{cov}\bfa H\times\sigma\lef((\bfa Q_i^{cov}\bfa H)^\top\bfa K_i^{cov}\bfa H\rig)\\
            &=\bfa H+\begin{bmatrix}
                \bfa 0&\bfa 0\\
                I_{d} &\bfa 0\\
                \bfa 0
            \end{bmatrix}\times\begin{bmatrix}
                \bfa X\\
                \bfa P
            \end{bmatrix}\times\begin{bmatrix}
                I_{N_1}&\bfa 0_{N_1\times(N-N_1)}\\
                \bfa 0_{(N-N_1)\times N_1}&\bfa 0
            \end{bmatrix}\times\begin{bmatrix}
                \bfa a_1&\ldots&\bfa a_{N_1}&\bfa 0_{D\times (N-N_1)}
            \end{bmatrix}\\
            &+\begin{bmatrix}
                I_d&\bfa 0\\
                \bfa 0&\bfa 0\\
                \bfa 0
            \end{bmatrix}\times\begin{bmatrix}
                \bfa X\\
                \bfa P
            \end{bmatrix}\times \begin{bmatrix}
                I_{N_1}& \bfa 0_{N_1\times (N-N_1)}\\
                \bfa 0_{(N-N_1)\times N_1}&\bfa 0
            \end{bmatrix}\times\begin{bmatrix}
                \bfa b&\ldots&\bfa b
            \end{bmatrix}\\
            &=\bfa H+\begin{bmatrix}
                &\bfa 0_{d\times N}&\\
                 \bfa X_1&\ldots& X_{N_1}&\bfa 0\\
                &\bfa 0&
            \end{bmatrix}\times\begin{bmatrix}
                \bfa a_1&\ldots&\bfa a_{N_1}&\bfa 0_{D\times (N-N_1)}
            \end{bmatrix}\\
            &+\begin{bmatrix}
                \bfa X_1&\ldots& X_{N_1}&\bfa 0\\
                &\bfa 0&&
            \end{bmatrix}\times \begin{bmatrix}
                \bfa b&\ldots&\bfa b
            \end{bmatrix}\\
            &=\bfa H+\begin{bmatrix}
                &\bfa 0_{d\times N}\\
                 \bfa X_1-\bar{\boldsymbol{X}}&\ldots& X_{N_1}-\bar{\boldsymbol{X}}&\bfa 0\\
                 &\bfa 0
            \end{bmatrix}+\begin{bmatrix}
                -\bar{\boldsymbol{X}} & -\bar{\boldsymbol{X}}&\ldots & - \bar{\boldsymbol{X}}\\
                &\bfa 0&&
            \end{bmatrix}\\
            &=\begin{bmatrix}
                \bfa X_1-\bar{\boldsymbol{X}}&\ldots& X_{N_1}-\bar{\boldsymbol{X}}&\ldots& X_N-\bar{\boldsymbol{X}}\\
                \bfa X_1-\bar{\boldsymbol{X}}&\ldots &X_{N_1}-\bar{\boldsymbol{X}}&\bfa 0&\bfa 0\\
                &&\bfa P&
            \end{bmatrix},
        \end{align*}
        with $\bfa a$ given by $\bfa a_{i,j}=\frac{N_1-1}{N_1}$ when $j=i$ and $-\frac{1}{N_1}$ when $j\neq i$. And $\bfa b$ is given by $\bfa b_{ij}=-\frac{1}{N_1}$ for all $j\in[N_1]$ and $\bfa b_{ij}=0$ for all $j\notin[N_1]$.
        Then the rest of the layer design easily follows from that of the proof of the theorem \ref{thm3.1}, where we show that there exists a sub-Transformer network with number of layers $L=2\tau+5$, number of heads $M\leq\lambda_1^d\frac{C}{\epsilon^2}$ and $\tau\leq\frac{\log(1/\epsilon_0\delta)}{\epsilon_0}$ that outputs
        \begin{align*}
           \bfa H^{spect}:= \begin{bmatrix}
                \bfa X_1-\bar{\boldsymbol{X}}&\ldots &X_{N_1}-\bar{\boldsymbol{X}}&\ldots &X_N-\bar{\boldsymbol{X}}\\
                \wha v_1^{TF}&\bfa 0 &\bfa 0&\bfa 0&\bfa 0\\
                I_d&&\bfa 0&&\\
                \bfa 0&\bfa 0&\bfa 0&\bfa 0&\bfa 0
            \end{bmatrix}
        \end{align*}
        where the following holds with probability at least $1-\sqrt{\frac{\delta}{\pi}}-\exp(-C\delta^{-1/2})$, where $\delta<\frac{1}{2} d^{-1}$
        \begin{align}\label{diffv1tfb1pm}
            \lef\Vert\wha v_1^{TF}-\wha v_1^{PM}\rig\Vert_2\lesssim \tau\epsilon \lambda_1^2+\frac{\lambda_1\sqrt{\epsilon_0}}{\Delta},
        \end{align}
        where $\wha v_1^{TF}$ is the Transformer output, $\wha v_1^{PM}$ is the power method output. We let $\lambda_i$ be the $i$-th biggest eigenvalue of $\wh\Sigma$ and $\Delta\asymp \max_{i}|\lambda_i-\lambda_{i+1}|$.  Together with the result given by theorem 3.11 in \citep{blum2020foundations}, we can show that
        \begin{align*}
            \Vert\wha v_1^{TF}-\wha v_1\Vert_2&\leq \lef\Vert\wha v_1^{TF}-\wha v_1^{PM}\rig\Vert_2+\Vert\wha v_1-\wha v_1^{PM}\Vert_2\lesssim \frac{\log(1/\epsilon_0\delta)}{\epsilon_0}\epsilon\lambda_1^2+\frac{C\lambda_1\sqrt{\epsilon_0}}{\Delta}+\epsilon_0.
        \end{align*}
        Then the following layer returns our estimate $\bfa v_1^\top(\bfa X_i-\bar{\boldsymbol{X}})$ for $i\in[N]$,
        \begin{align*}
            \bfa V^{est}_1= -\bfa V^{est}_2 = \begin{bmatrix}
                &\bfa 0_{3d\times D}&\\
                \bfa 0_{d\times 2d}& I_d&\bfa 0\\
                &\bfa 0&
            \end{bmatrix},\quad\bfa Q^{est}_1=\begin{bmatrix}
                \bfa 0_{d\times d} & I_d&\bfa 0
            \end{bmatrix},\quad \bfa K_1^{est}=I_d.
        \end{align*}
        Using the above construction we can show that
        \begin{align*}
           \bfa H^{spect,2}&= \bfa H^{spect}+ \sum_{i=1}^2\bfa V_1^{est}\bfa H^{spect}\sigma\lef((\bfa Q_1^{est}\bfa H^{spect})^\top(\bfa K_1^{est}\bfa H^{spect})\rig)\\
           &=\bfa H^{spect} +\begin{bmatrix}
               &\bfa 0_{3d\times N}&\\
               \wha v_1^{TF,\top}(\bfa X_1-\bar{\boldsymbol{X}}) &\ldots&\wha v_1^{TF,\top}(X_N-\bar{\boldsymbol{X}})\\
               &\bfa 0&
           \end{bmatrix}\\
           &=\begin{bmatrix}
               \bfa X_1-\bar{\boldsymbol{X}}&\ldots&X_{N_1}-\bar{\boldsymbol{X}}&\ldots &X_N-\bar{\boldsymbol{X}}\\
               \wha v_1^{TF}&\bfa 0&\bfa 0&\bfa 0&\bfa 0\\
               I_d&&\bfa 0&&\\
               \wha v_1^{TF,\top}(\bfa X_1-\bar{\boldsymbol{X}})&&\ldots&&\wha v_1^{TF,\top}(X_N-\bar{\boldsymbol{X}})\\
               &&\bfa 0&&
           \end{bmatrix}.
        \end{align*}
        We construct $W_2=\begin{bmatrix}
            \bfa 0_{1\times 3d}&1&\bfa 0
        \end{bmatrix}$ and the resulting output satisfies
        \begin{align*}
            W_2\bfa H^{spect, 2}= \begin{bmatrix}
                \wha v_1^{TF,\top}(\bfa X_1-\bar{\boldsymbol{X}})&\ldots&\wha v_1^{TF,\top}(X_N-\bar{\boldsymbol{X}})
            \end{bmatrix}.
        \end{align*}
    Then, applying the $\tanh(\beta x)$ nonlinearities we obtain the final conclusion.

\subsection{Proof of Theorem \ref{thm3.2}}
\begin{proof}[Proof of Theorem \ref{thm3.2}]
        We first consider the upper bound for the upper bound on the eigenvalues of $\wh\Sigma$. Denote $\wh\Sigma^{(i)}$ to be the empirical covariance matrix built upon the $i$-th pretraining instance. Similarly we denote $\wha v^{(i)}$ as the corresponding estimated eigenvector for $i$-th pretraining instance. By union bound we can show that
        \begin{align*}
            \bb P\lef(\exists i\in[n], {s.t.  }\frac{\Vert\wh\Sigma^{(i)}-\Sigma^{(i)}\Vert_2}{1+\frac{1}{4}\Vert\bfa\mu_1^{(i)}-\bfa\mu_0^{(i)}\Vert_2^2}\geq C\Big(\sqrt{\frac{d}{N_1^{(i)}}}+\frac{d}{N_1^{(i)}}+\delta\Big)\rig)\leq n\exp\lef(-CN_1^{(i)}\min(\delta,\delta^2)\rig).
        \end{align*}
        By \eqref{diffsigmatde} with probability at least $1-\delta$, simultaneously for all $i\in[n]$,
        \begin{align}\label{ubwhsigmai}
            \Vert\wh\Sigma^{(i)}\Vert_2&\leq \Vert\wh\Sigma^{(i)}-\Sigma^{(i)}\Vert_2+\Vert\Sigma^{(i)}\Vert_2\nnb\\
            &=\Big(1+\frac{1}{4}\Vert\bfa\mu_1^{(i)}-\bfa\mu_0^{(i)}\Vert_2^2\Big)\bigg(1+\sqrt{\frac{d}{N_1^{(i)}}}+\frac{d}{N_1^{(i)}}+\sqrt{\frac{\log(n/\delta)}{N_1^{(i)}}}\vee\frac{\log(n/\delta)}{N_1^{(i)}}\bigg).
        \end{align}
        And we denote the event in \eqref{ubwhsigmai} as $\ca E_{\delta}^{1}$. Then we consider the difference between the output and the target vector. Using \eqref{diffv1tfb1pm} and the theorem 3.11 in \citep{blum2020foundations} we can show that given $\tau\geq\frac{\log(n/\epsilon \delta)}{2\epsilon}$, we have with probability at least $1-\delta$,
        \begin{align*}
           \Vert\wha v_1-\wha v_1^{PM}\Vert_2 = 2-2 \wha v_1^\top\wha v_1^{PM}\leq 2\epsilon.
        \end{align*}
        And we denote the above event as $\ca E_\delta^2$. 
        Then we immediately obtain that under $\ca E_\delta^1\cup\ca E_\delta^2$, the following holds
        \begin{align*}
            \lambda_1^{(i)}&=\Vert\wh\Sigma^{(i)}\Vert_2 \leq\Vert\wh\Sigma^{(i)}-\Sigma^{(i)}\Vert_2+\Vert\Sigma^{(i)}\Vert_2\\
            &\leq\Big(1+\frac{1}{4}\Vert\bfa\mu_0^{(i)}-\bfa\mu_1^{(i)}\Vert_2^2\Big)\Big( 1+C\sqrt{\frac{d}{N_1^{(i)}}}+\frac{d}{N_1^{(i)}}+\sqrt{\frac{\log(n/\delta)}{N_1^{(i)}}}\vee\frac{\log(n/\delta)}{N_1^{(i)}}\Big)\\
            &\leq \ub{\Vert\bfa\mu_0^{(i)}-\bfa\mu_1^{(i)}\Vert_2^2\Big(1+\sqrt{\frac{d}{N_1^{(i)}}}+\sqrt{\frac{\log(n/\delta)}{N_1^{(i)}}}\vee\frac{\log(n/\delta)}{N_1^{(i)}}\Big)}_{\tde\lambda^{(i)}_1(\delta)}.
        \end{align*}
        And similarly we can show that $\Delta^{(i)}$ is lower bounded by
        \begin{align*}
            \Delta^{(i)}&\geq \frac{1}{4}\Vert\bfa\mu_0^{(i)}-\bfa\mu_1^{(i)}\Vert_2^2-2\Big(1+\frac{1}{4}\Vert\bfa\mu_0^{(i)}-\bfa\mu_1^{(i)}\Vert_2^2\Big)\Big(\sqrt{\frac{d}{N_1^{(i)}}}+\frac{d}{N_1^{(i)}}+\sqrt{\frac{\log(n/\delta)}{N_1^{(i)}}}\vee\frac{\log(n/\delta)}{N_1^{(i)}}\Big)\\
            &\gtrsim\ub{\Vert\bfa\mu_0^{(i)}-\bfa\mu_1^{(i)}\Vert_2^2\Big(1-C\sqrt{\frac{d}{N_1^{(i)}}}- C\sqrt{\frac{\log(n/\delta)}{N_1^{(i)}}}\vee\frac{\log(n/\delta)}{N_1^{(i)}}\Big)}_{\tde\Delta(\delta)},
        \end{align*}
        with probability at least $1-\delta$.
        Collecting the above pieces, we can show that with probability at least $1-\delta$, simulatneously for all $i\in[n]$,
        \begin{align*}
            \Vert\wha v_1^{TF,(i)}-\wha v_1^{(i)}\Vert_2&\lesssim \frac{\log(1/(\epsilon_0\delta))}{\epsilon_0}\epsilon\tde\lambda_1^{{(i)},2}(\delta)+\frac{\lambda_1^{(i)}(\delta)\sqrt{\epsilon_0}}{\tde\Delta^{(i)}(\delta)}+\epsilon_0.
        \end{align*}
        Collecting the above pieces and using the fact that with probability at least $1-\delta$ simultaneously for all $i\in[n]$ we have
        \begin{align*}
            \Vert\wha v_1^{TF,(i)}-\bfa v_1^{(i)}\Vert_2&\leq\Vert\wha v_1^{TF,(i)}-\wha v_1^{(i)}\Vert_2+\Vert\wha v_1^{(i)}-\bfa v_1^{(i)}\Vert_2\\
            &\lesssim \frac{\log(1/\epsilon)+\log(1/\delta)}{\epsilon_0}\epsilon\tde\lambda_1^{(i),2}(\delta)+\frac{\lambda_1^{(i)}(\delta)\sqrt{\epsilon_0}}{\tde\Delta^{(i)}(\delta)}+\epsilon_0\\
            &+\sqrt{\frac{d(1+\log(1/\delta))}{N_1^{(i)}}}\Big(\Vert\bfa\mu^{(i)}_1-\bfa\mu^{(i)}_0\Vert_2^{-2}+1\Big)+\frac{1}{\Vert\bfa\mu_1^{(i)}-\bfa\mu_0^{(i)}\Vert_2^2N_1^{(i)}}.
        \end{align*}
        From here on we consider the loss upper bound given by a general sample and omit the superscript $(i)$ that index the $i$-th pre-training instance.
        
        Then, we apply the same proof idea of theorem \ref{thm411} by modifying \eqref{diffff} as
        \begin{align*}
            \bb E&[L_{GMM}(TF_{\bfa\theta}(\bfa H^{(1)}), z^{(1)})]= \bb E\bigg[\min_{\tde z\in\{z,-z\}}\frac{1}{N}\sum_{i=1}^NL(TF_{\bfa\theta}(\bfa H^{(1)})_i,z_i)\bigg]\leq T_1 + T_2^\prime,
        \end{align*}
        where for $T_1$ we are able to use \eqref{t1upperbound} to show that
        \begin{align*}
            T_1\leq C\exp\lef(-C\beta^2\Vert\bfa\mu_0-\bfa\mu_1\Vert_2^2\rig)+\frac{C}{\Vert\bfa\mu_0-\bfa\mu_1\Vert_2}\exp\Big(-\frac{1}{4}\Vert\bfa\mu_0-\bfa\mu_1\Vert_2^2\Big).
        \end{align*}
        And for $T_2^\prime$, we can show that
        \begin{align*}
            T_2^\prime &\leq\bb E\Big[(1-\tanh^2(\beta(X-\bb E[\bfa X])^\top\bfa v_1)\beta\Big(| X^\top(\bfa v_1-\wha v_1^{TF})|+|\bb E[\bfa X]^\top\bfa v_1-\bar{\boldsymbol{X}}^\top\wha v_1^{TF}|\Big)\Big]\\
            &=\bb E\Big[(1-\tanh^2(\beta(X-\bb E[\bfa X])^\top\bfa v_1)\beta|X^\top(\bfa v_1-\wha v_1^{TF})|\Big]\\
            &+\bb E\Big[(1-\tanh^2(\beta(X-\bb E[\bfa X])^\top\bfa v_1)\beta \lef|\bb E[\bfa X]^\top\bfa v_1-\bar{\boldsymbol{X}}^\top\wha v_1^{TF}\rig|\Big]\\
            &=\bb E\Big[(1-\tanh^2(\beta(X-\bb E[\bfa X])^\top\bfa v_1)\beta|X^\top(\bfa v_1-\wha v_1^{TF})|\Big]\\
            &+\bb E\Big[(1-\tanh^2(\beta(X-\bb E[\bfa X])^\top\bfa v_1)\beta\Big]\bb E\Big[\lef|\bb E[\bfa X]^\top\bfa v_1-\bar{\boldsymbol{X}}^\top\wha v_1^{TF}\rig|\Big].
        \end{align*}
        Note that for the distribution we can show that with probability at least $1-\delta$,
        \tny{\begin{align*}
        \Big|&\bb E[\bfa X]^\top\bfa v_1- \bar  X^\top\wha v_1\Big|= \Vert \bar{\boldsymbol{X}}\Vert_2 \Vert\wha v_1^{TF}-\bfa v_1\Vert_2+ (\bar{\boldsymbol{X}}-\bb E[\bfa X])^\top\bfa v_1\\
        &\lesssim\sqrt{\frac{1}{N_1}\bl\frac{1}{2}\Vert\bfa\mu_0-\bfa\mu_1\Vert_2^2+2\br\log(n/\delta)}+\bl\frac{\Vert\bfa\mu_1-\bfa\mu_0\Vert_2}{2}+\sqrt{\frac{(d+\frac{1}{4}\Vert\bfa\mu_0-\bfa\mu_1\Vert_2^2)\log(n/\delta)}{N_1}}\br\\
        &\cdot\Big(\frac{\log(1/(\epsilon_0\delta))}{\epsilon_0}\epsilon\tde\lambda_1^2(\delta)+\frac{\tde\lambda_1(\delta)\sqrt{\epsilon_0}}{\tde\Delta(\delta)}+\epsilon_0+\sqrt{\frac{d(1+\log(n/\delta))}{N_1}}\Big(\Vert\bfa\mu_1-\bfa\mu_0\Vert_2^{-2}+1\Big)+\frac{1}{\Vert\bfa\mu_1-\bfa\mu_0\Vert_2^2N_1}\Big)\\
        &\lesssim \sqrt{\frac{1}{N_1}\lef(\Vert\bfa\mu_0-\bfa\mu_1\Vert_2^2+4\rig)\log(n/\delta)}+\sqrt{\frac{d}{N_1}}\lef(\Vert\bfa\mu_1-\bfa\mu_0\Vert_2^{-1}+4\Vert\bfa\mu_1-\bfa\mu_0\Vert_2\rig)+\frac{1}{2}\Vert\bfa\mu_1-\bfa\mu_0\Vert_2\\
        &+\Big(\frac{\log(1/(\epsilon_0\delta))}{\epsilon_0}\epsilon\tde\lambda^2_1(\delta)+\frac{\tde\lambda_1(\delta)\sqrt{\epsilon_0}}{\tde\Delta(\delta)}+\epsilon_0\Big)\cdot\bigg(\frac{1}{2}\Vert\bfa\mu_1-\bfa\mu_0\Vert_2+\sqrt{\frac{d\log(n/\delta)}{N_1}}\bigg).
        \end{align*}}
        Then, we collect the above terms and compare it with the result in \eqref{diffff} to show that
        \tny{\begin{align*}
            &T_2^\prime \lesssim T_2+\beta\bigg(\frac{\log(1/(\epsilon_0\delta))}{\epsilon_0}\epsilon\tde\lambda_1^2(\delta)+\frac{\tde\lambda_1(\delta)\sqrt{\epsilon_0}}{\tde\Delta(\delta)}+\epsilon_0\bigg)\bigg(\frac{1}{2}\Vert\bfa\mu_0-\bfa\mu_1\Vert_2+\sqrt{\frac{d\log(n/\delta)}{N_1}}\bigg).
        \end{align*}}
        
        Then we consider the expected loss induced by the construction, defining $\tde F(\bfa X_i):=\tanh(\beta(\bfa X_i-\bb E[\bfa X])^\top\bfa v_1)$, we can show that following \eqref{lossub} we obtain that
        \begin{align*}
            \bb E&\lef[L_{GMM}(TF_{\bfa\theta},z)\rig]\lesssim\bb E\lef[L_{GMM}(TF_{\bfa\theta},z)|\ca E_{\delta}^1\cup\ca E_{\delta}^2\rig]\bb P\lef(\ca E_{\delta}^1\cup\ca E_{\delta}^2\rig)+\bb P\lef((\ca E_{\delta}^1\cup\ca E_{\delta}^2)^c\rig) \\
            &\lesssim\frac{N_1}{N}+(1+\beta^2\Vert\bfa\mu_0-\bfa\mu_1\Vert_2^2)\exp\lef(-C\beta^2\Vert\bfa\mu_0-\bfa\mu_1\Vert_2^2\rig)+\delta \\
            &+\sqrt{\frac{d}{N_1}}\lef(\Vert\bfa\mu_1-\bfa\mu_0\Vert_2+\log(n/\delta)\rig)+\frac{1}{\Vert\bfa\mu_0-\bfa\mu_1\Vert_2}\exp\Big(-\frac{1}{4}\Vert\bfa\mu_0-\bfa\mu_1\Vert_2^2\Big)\\
            &+\beta\Big(\frac{\log(1/(\epsilon_0\delta))}{\epsilon_0}\epsilon\tde\lambda_1^2(\delta)+\frac{\tde\lambda_1(\delta)\sqrt{\epsilon_0}}{\tde\Delta(\delta)}+\epsilon_0\Big)\Big(\frac{1}{2}\Vert\bfa\mu_1-\bfa\mu_0\Vert_2+\sqrt{\frac{d\log(n/\delta)}{N_1}}\Big).
        \end{align*}
        Then we consider the generalization error. Using the machine created in the proof of proposition \ref{genbound} we can show that
        \begin{enumerate}
            \item With probability at least $1-\delta$, simultaneously for all $[n]$ we have $\lambda_1\lesssim \Vert\bfa\mu_1-\bfa\mu_0\Vert_2^2\Big(1+\sqrt{\frac{\log(n/\delta)}{N_1}}\vee\frac{\log(n/\delta)}{N_1}\Big)$.
            \item The entropy of the operator norm ball is given by
            \begin{align*}
                \log N\lef(\delta, B_{\Vert\cdot\Vert_{op}}(\delta),\Vert\cdot\Vert_{op}\rig)\leq CLB_MD^2\log(1+2(B_{\theta}+B_{\bfa\mu}^2)/\delta).
            \end{align*}
            \item The loss is upperbounded as $L(TF_{\bfa\theta}(\bfa H),z)\leq C$.
            \item The Lipschitz condition of Transformers satisfies that for all $\bfa\theta_1,\bfa\theta_2\in\Theta(B_{\bfa\theta}, B_M)$ we have
            $L\lef(TF_{\bfa\theta_1}(\bfa H),z\rig)-L\lef(TF_{\bfa\theta_2}(\bfa H),z\rig)\leq C\beta LB_1^L\Vert\bfa\theta_1-\bfa\theta_2\Vert_{op}$,
            where $B_1=B_{\theta}^4B_{\bfa\mu}^6$.
        \end{enumerate}
        Therefore, we can show that with probability at least $1-\delta$, we have
        \begin{align*}
            L\lef(TF_{\bfa\theta}(\bfa H),z\rig)\leq\inf_{\bfa\theta\in\Theta(B_{\bfa\theta},B_M)}\bb E[L(TF_{\wha\theta(\bfa H)},z)]+2C\sqrt{\frac{LB_Md^2\log(B_{\theta}+B_{\bfa\mu})+\log(1/\delta)}{n}}.
        \end{align*}
        Replacing the $\tde\lambda_1(\delta)=\Vert\bfa\mu_0-\bfa\mu_1\Vert_2^2\Big(1+\sqrt{\frac{d}{N_1}}+\sqrt{\frac{\log(n/\delta)}{N_1}}\Big)$ and $\tde\Delta(\delta)=\Vert\bfa\mu_0-\bfa\mu_1\Vert_2^2\Big(1-C\sqrt{\frac{d}{N_1}}- C\sqrt{\frac{\log(n/\delta)}{N_1}}\vee\frac{\log(n/\delta)}{N_1}\Big)$, we can show that as we can show that the ERM estimator $\wha\theta_{GMM}$ satisfies
        \sm{\begin{align*}
            \bb E[L_{GMM}(&TF_{\bfa\theta_{GMM}}(\bfa H),z)]\lesssim \frac{N_1}{N}+\beta\sqrt{\epsilon\log(1/\sqrt{\epsilon})}B_{\bfa\mu}^2+\beta^2\Vert\bfa\mu_1-\bfa\mu_0\Vert_2^2\exp(-C\beta^2\Vert\bfa\mu_0-\bfa\mu_1\Vert_2^2)\\
            &+\sqrt{\frac{d}{N_1}}\lef(B_{\bfa\mu}+\log(n/\delta)\rig)+2C\sqrt{\frac{(1+\frac{1}{\sqrt{\epsilon}})B_Md^2\log(B_{\theta}+B_{\bfa\mu})+\log(1/\delta)}{n}}+\delta,
        \end{align*}}
        where we already let $\epsilon_0 = \frac{1}{\sqrt{\epsilon
        }}$.
        Then we consider taking the values of $B_M \leq B_{\bfa\mu}^{2d}\frac{C}{\epsilon^2}$ and taking $\epsilon=\lef(n^{-\frac{1}{2}}B_{\bfa\mu}^{d-2}d(\log B_{\bfa\mu})^{\frac{1}{2}}\rig)^{\frac{4}{7}}$, $\delta\asymp \Big(\frac{d\log^2N}{N}\Big)^{\frac{1}{3}} + B_{\bfa\mu}^{\frac{2}{7}d} d^{\frac{2}{7}}n^{-\frac{1}{7}}(\log B_{\bfa\mu})^{\frac{1}{7}}(\beta B_{\bfa\mu}^2)^{\frac{4}{7}}$, $\beta\asymp 1$, $N_1\asymp d^{1/3}N^{2/3}(\Vert\bfa\mu_1-\bfa\mu_0\Vert_2+\log N)^{2/3}$, we conclude that
        \begin{align*}
            \bb E[L_{GMM}(&TF_{\bfa\theta_{GMM}}(\bfa H),z)]\lesssim \Big(\frac{d\log^2N}{N}\Big)^{\frac{1}{3}} + B_{\bfa\mu}^{\frac{2}{7}d} d^{\frac{2}{7}}n^{-\frac{1}{7}}(\log B_{\bfa\mu})^{\frac{1}{7}}(\beta B_{\bfa\mu}^2)^{\frac{4}{7}}.
        \end{align*}
    \end{proof}


\section{Experimental Details}\label{sec:exp_details}

\subsection{Model Architecture Detail for Seciton \ref{sect401}.}
We use slightly different architectures to predict eigenvalues and principal components.
For eigenvalues prediction, we flatten the transformer output $TF_{\bfa\theta}(\bfa X) \in \R^{N \times \text{embd\_dim}}$ and use a linear layer $\bfa W_\lambda \in \R^{(N\cdot \text{embd\_dim}) \times k}$ to readout the top $k$ eigenvalues.
As for principal components, we use one more linear layer $\bfa W_v \in \R^{(N\cdot \text{embd\_dim}) \times (k\cdot d)}$ to readout $k$ eigenvectors concatenated in a $1$-dimension vector.

\subsection{Additional Experimental Results}\label{sec:add_exp}

\begin{table}[t]
\centering
\caption{\textbf{Multi-$k$ Principal Components Prediction on Synthetic Dataset.}
We train a large transformer ($L=12, {\rm embd} = 256, {\rm head} = 8$) on synthetic dataset to predict top-$k$ principal components, and use cosine similarity as metric. This is the detailed result of the right subplot in figure \ref{fig:encoder_eigenvector}.
}
\begin{tabular}{lcccc}
\toprule
\textbf{k-th eigenvec.} & \textbf{k=1} & \textbf{k=2} & \textbf{k=3} & \textbf{k=4} \\ \hline
$k_{\text{train}}=4$    & $0.9525 (0.0065)$ & $0.8648 (0.0092)$ & $0.7286 (0.0142)$ & $0.4471 (0.0722)$ \\
$k_{\text{train}}=3$    & $0.9515 (0.0085)$ & $0.8454 (0.0273)$ & $0.6574 (0.0041)$ & - \\
$k_{\text{train}}=2$   & $0.9598 (0.0017)$ & $0.8326 (0.0143)$ & - & - \\
$k_{\text{train}}=1$   & $0.9404 (0.0033)$ & - & - & - \\
\bottomrule
\end{tabular}
\label{tab:mul_k_vector_encoder}
\end{table}

\begin{figure}[!h]
    \centering
    \minipage{0.3\textwidth}
        \includegraphics[width=\linewidth]{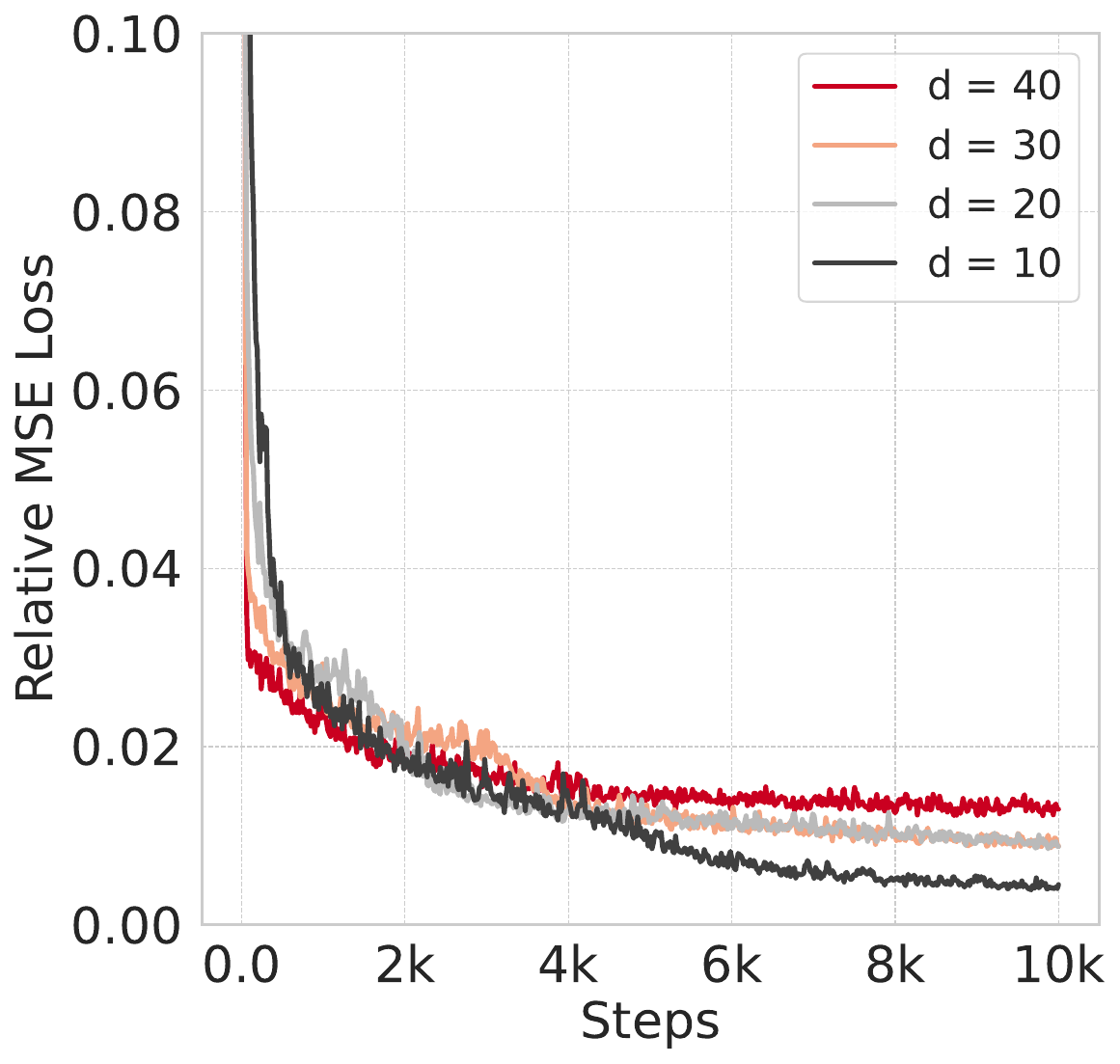}
    \endminipage\hfill
    \minipage{0.3\textwidth}
        \includegraphics[width=\linewidth]{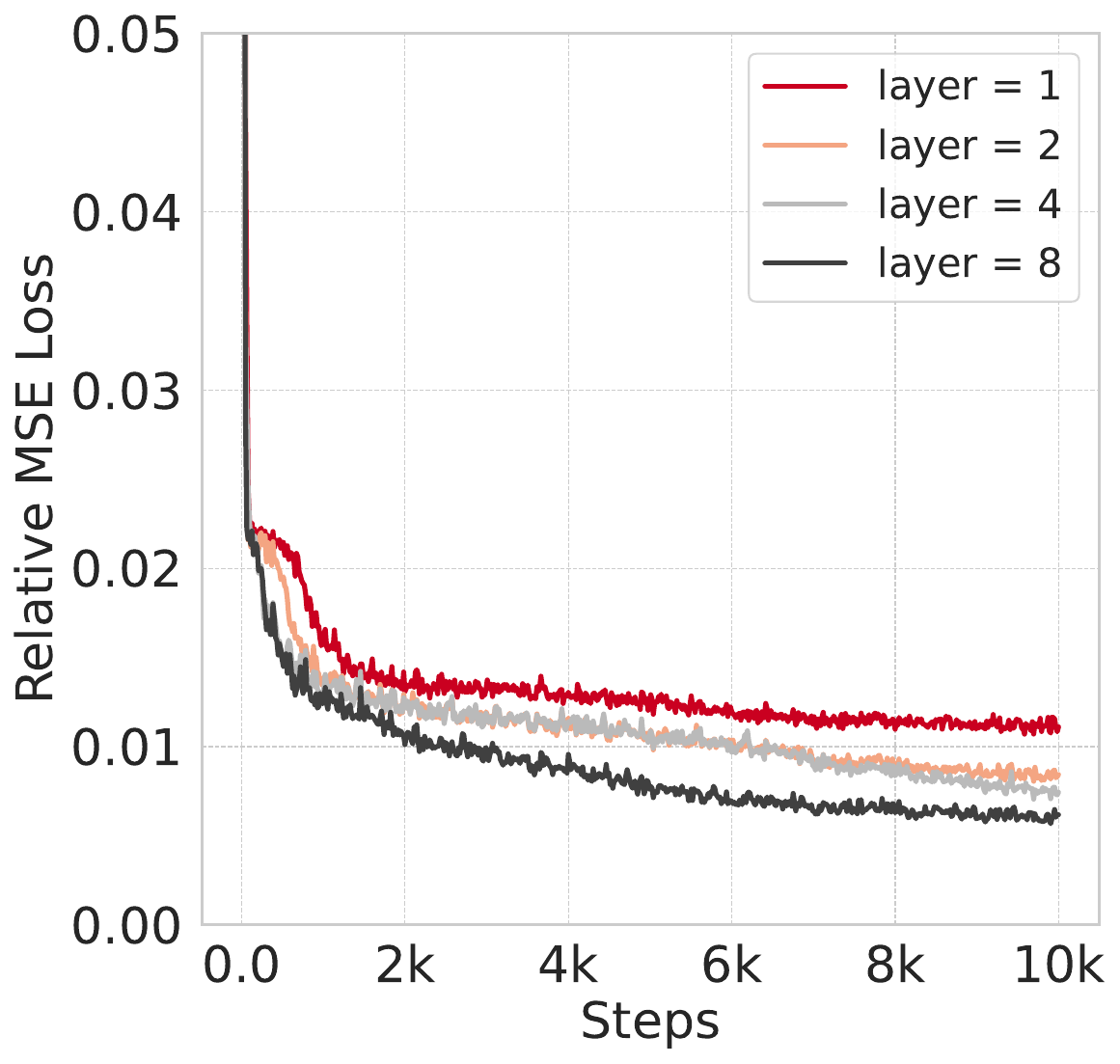}
    \endminipage\hfill
    \minipage{0.3\textwidth}
        \includegraphics[width=\linewidth]{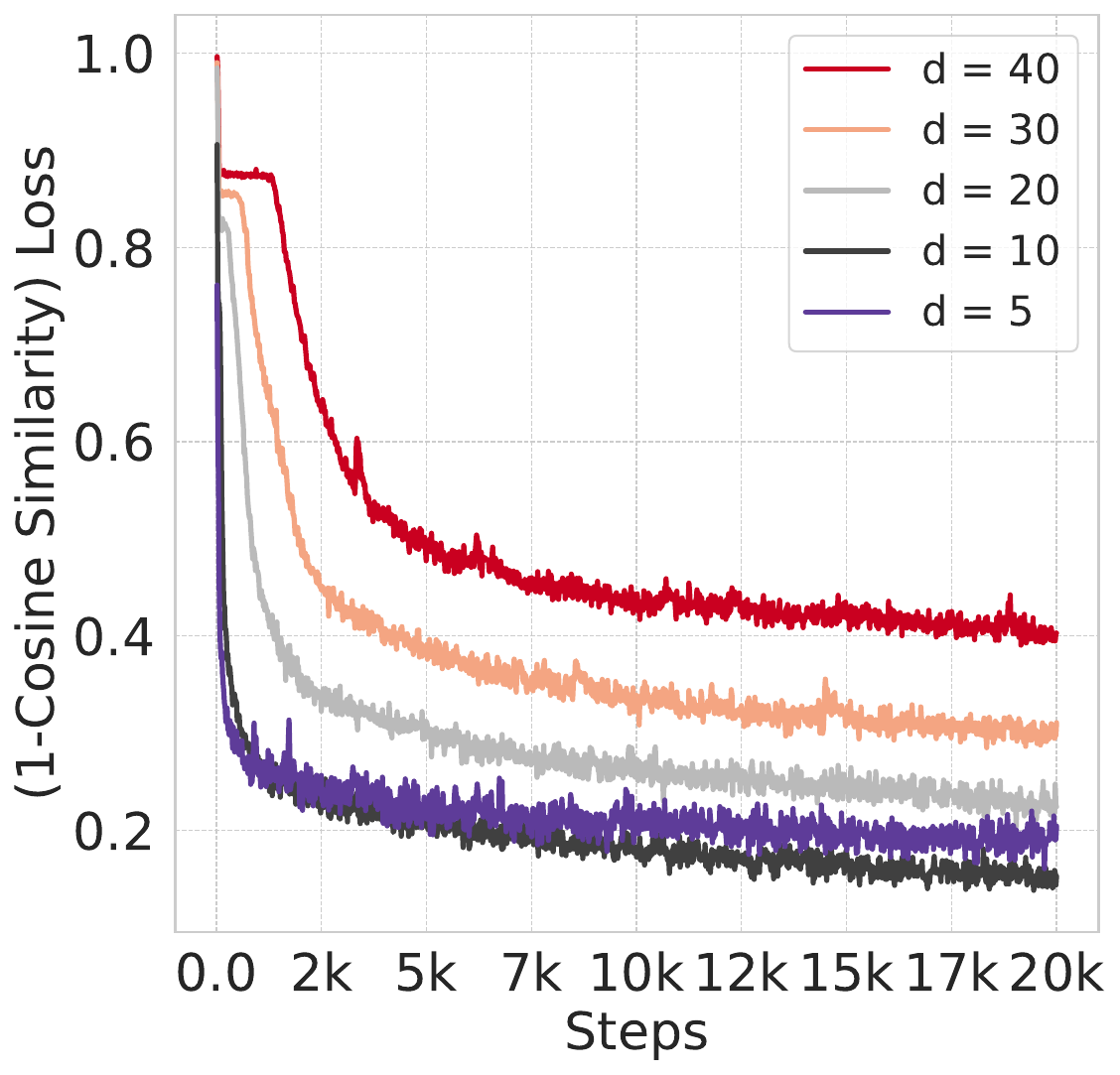}
    \endminipage\vspace{1em}
    \minipage{0.45\textwidth}
        \includegraphics[width=\linewidth]{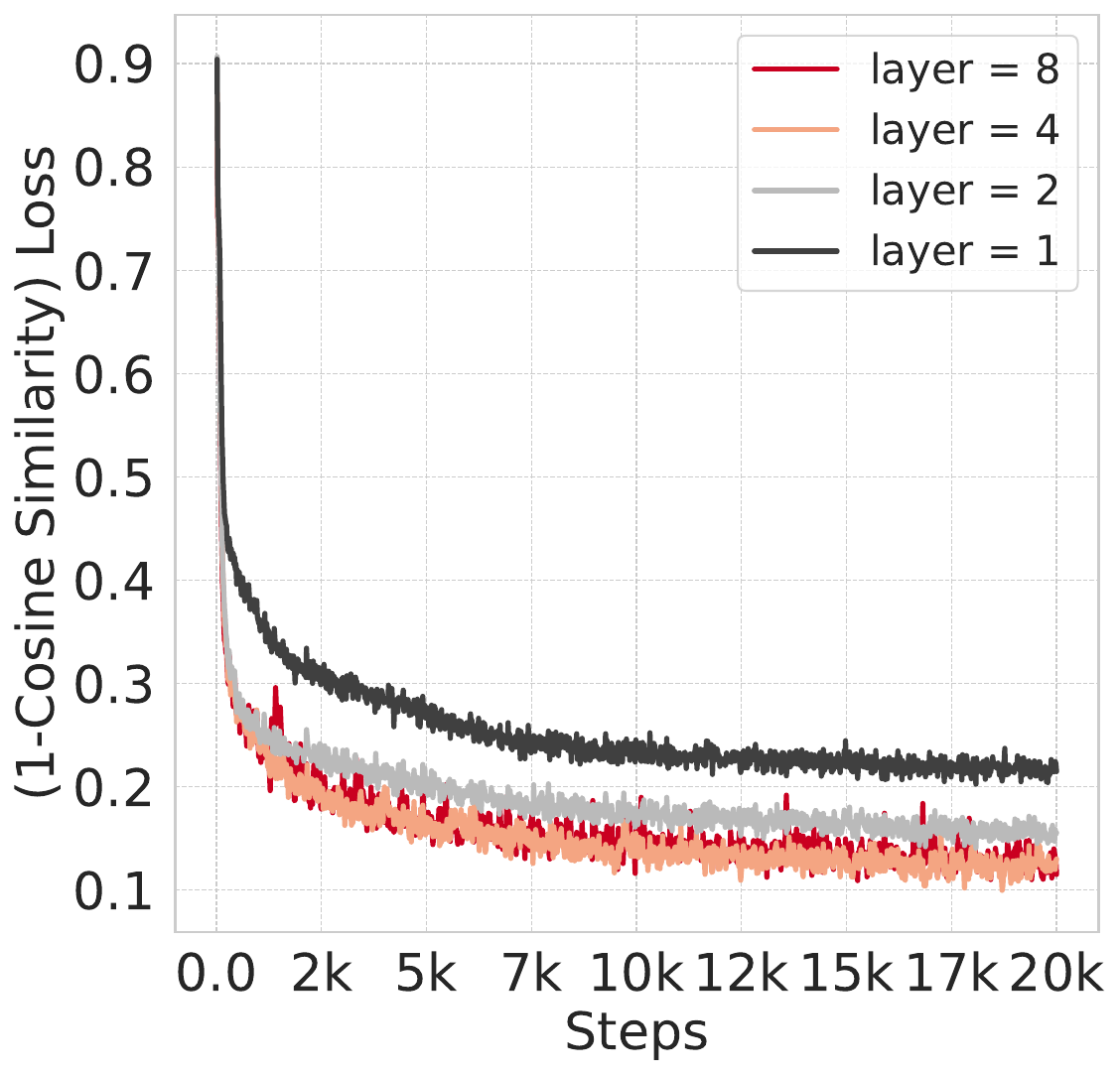}
    \endminipage\hfill
    \minipage{0.45\textwidth}
        \includegraphics[width=\linewidth]{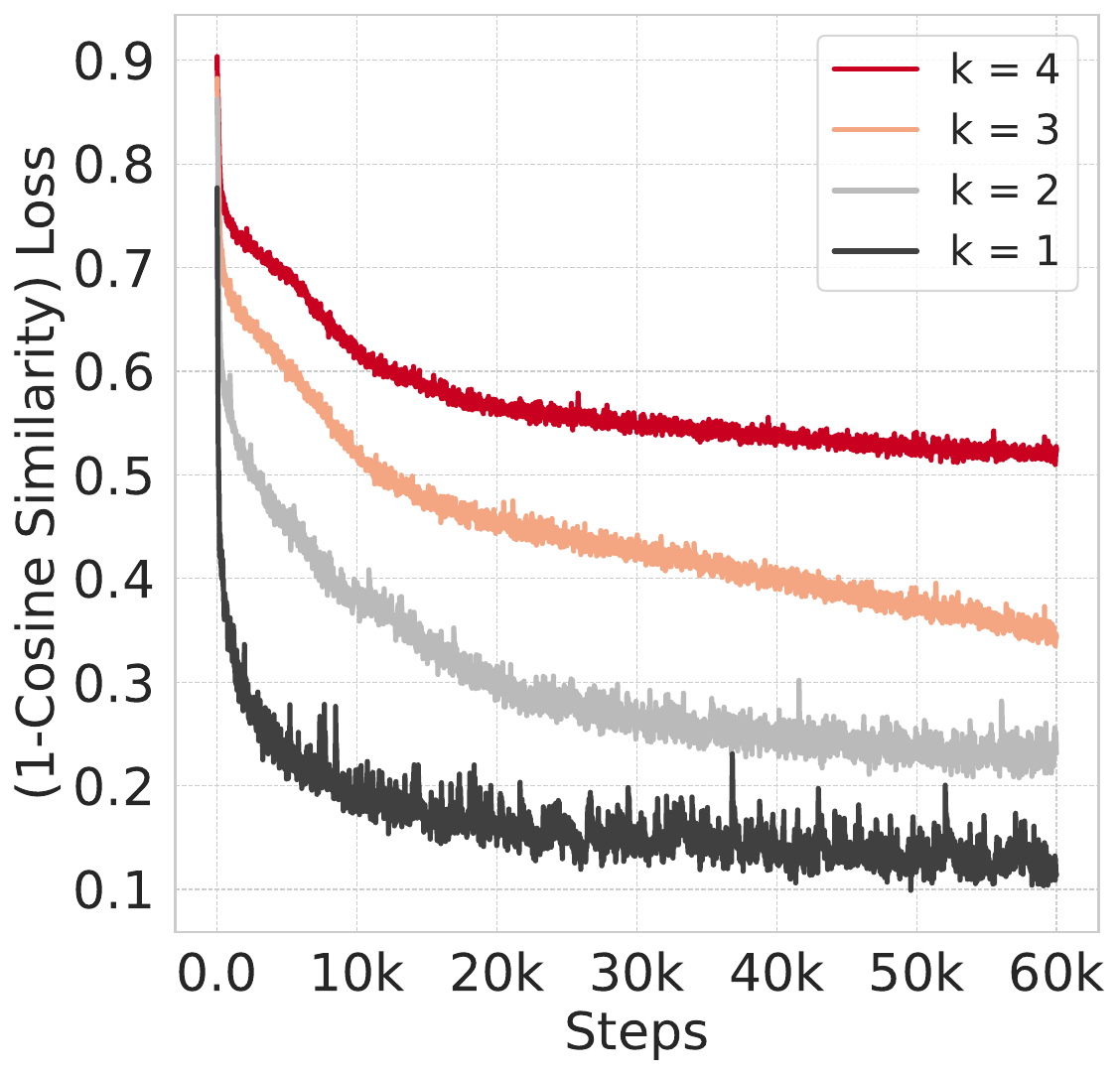}
    \endminipage
    \caption{\textbf{Convergence Results on Eigenvalue, Eigenvector Prediction with Different Parameters.}
    \emph{(1) Top left: Loss curve on eigenvalue prediction with different size of $d$}
    \emph{(2) Top middle: Loss curve on eigenvalue prediction with different number of layers}
    \emph{(3) Top right: Loss curve on eigenvector prediction with different size of $d$}
    \emph{(4) Bottom left: Top right: Loss curve on eigenvector prediction with different number of layers}
    \emph{(5) Bottom left: Loss curve on eigenvector prediction with different number of $k_{\text{train}}$}
    For (1), we observe that smaller $d$ is easiser for transformers as they present lower loss.
    For (2), we see that with more layers, transformers are also capable of predicting eigenvalues more accurately.
    For (3), transformers also predict eigenvectors better when $d$ is small.
    For (4), similar to (2), transformers with more layers shows improved performance.
    For (5), we want to highlight that the loss value is mainly affected by the fact that predicting 3rd or 4th eigenvectors are significantly harder, which contributes to higher loss value.
    }
    \label{fig:loss}
\end{figure}

We provide two additional results in this section. 
Table \ref{tab:mul_k_vector_encoder} is the detailed cosine similarity of individual principal components prediction of the right subplot in Figure \ref{fig:encoder_eigenvector}.
Figure \ref{fig:loss} illustrates the training loss from the experiments discussed in Section \ref{sect401}. These experiments explore the effects of (1) varying data dimension, (2) adjusting the number of layers for predicting eigenvalues and principal components, and (3) varying the number of principal components models are trained to predict.

\clearpage

\bibliographystyle{plain}

\bibliography{iclr2025_conference}
\end{document}